\documentclass[10pt, journal, compsoc]{IEEEtran}
\usepackage{cite}

\ifCLASSINFOpdf
  \usepackage[pdftex]{graphicx}
  \DeclareGraphicsExtensions{.pdf,.jpeg,.png}
\else
  \usepackage[dvips]{graphicx}
  \DeclareGraphicsExtensions{.eps}
\fi
\usepackage{xcolor}
\usepackage{comment}
\usepackage{kotex}
\usepackage{booktabs}
\usepackage{multirow}
\usepackage{makecell}
\usepackage{amsmath}
\usepackage{amsfonts}
\usepackage{amssymb}
\usepackage{amsthm}
\usepackage{nicefrac}
\usepackage{hyperref}
\usepackage{enumerate}
\usepackage{enumitem}

\usepackage{algorithmic}
\usepackage{array}

\begin{document}
%
% paper title
% Titles are generally capitalized except for words such as a, an, and, as,
% at, but, by, for, in, nor, of, on, or, the, to and up, which are usually
% not capitalized unless they are the first or last word of the title.
% Linebreaks \\ can be used within to get better formatting as desired.
% Do not put math or special symbols in the title.
\title{A Review of Online Diffusion Policy RL Algorithms for Scalable Robotic Control}
% Systems}
%
%
% author names and IEEE memberships
% note positions of commas and nonbreaking spaces ( ~ ) LaTeX will not break
% a structure at a ~ so this keeps an author's name from being broken across
% two lines.
% use \thanks{} to gain access to the first footnote area
% a separate \thanks must be used for each paragraph as LaTeX2e's \thanks
% was not built to handle multiple paragraphs
%

\author{Wonhyeok~Choi, Shutong~Ding, Minwoo~Choi, Jungwan Woo, Kyumin Hwang, Jaeyeul Kim,\\Ye~Shi$^\dagger$, and~Sunghoon~Im$^\dagger$,~\IEEEmembership{Member,~IEEE}% <-this % stops a space
\thanks{W. Choi, M. Choi, J. Woo, K. Hwang, J. Kim, S. Im with the Daegu Gyeongbuk Institute of Science and Technology (DGIST), Daegu 42988, South Korea.}%
\thanks{S. Ding, Y. Shi with the ShanghaiTech University, Shanghai 201210, China.}%
\thanks{E-mail: \{smu06117, subminu, friendship1, kyumin, jykim94, sunghoonim\}@dgist.ac.kr, \{dingsht, shiye\}@shanghaitech.edu.cn}% <-this % stops a space
% \thanks{$\dagger$: S. Im are the corresponding author.}}
\thanks{$\dagger$: S. Im and Y. Shi are the co-corresponding authors.}}

\maketitle

% As a general rule, do not put math, special symbols or citations
% in the abstract or keywords.
\begin{abstract}
Diffusion policies have emerged as a powerful approach for robotic control, demonstrating superior expressiveness in modeling multimodal action distributions compared to conventional policy networks.
However, their integration with online reinforcement learning remains challenging due to fundamental incompatibilities between diffusion model training objectives and standard RL policy improvement mechanisms.
This paper presents the first comprehensive review and empirical analysis of current Online Diffusion Policy Reinforcement Learning (Online DPRL) algorithms for scalable robotic control systems.
We propose a novel taxonomy that categorizes existing approaches into four distinct families—Action-Gradient, Q-Weighting, Proximity-Based, and Backpropagation Through Time (BPTT) methods—based on their policy improvement mechanisms.
Through extensive experiments on a unified NVIDIA Isaac Lab benchmark encompassing 12 diverse robotic tasks, we systematically evaluate representative algorithms across five critical dimensions: task diversity, parallelization capability, diffusion step scalability, cross-embodiment generalization, and environmental robustness.
Our analysis identifies key findings regarding the fundamental trade-offs inherent in each algorithmic family, particularly concerning sample efficiency and scalability. Furthermore, we reveal critical computational and algorithmic bottlenecks that currently limit the practical deployment of online DPRL.
Based on these findings, we provide concrete guidelines for algorithm selection tailored to specific operational constraints and outline promising future research directions to advance the field toward more general and scalable robotic learning systems.
\end{abstract}

% Note that keywords are not normally used for peerreview papers.
% \begin{abstract}
% The abstract goes here.
% \end{abstract}

% Note that keywords are not normally used for peerreview papers.
% \begin{IEEEkeywords}
% Robot Learning, Diffusion Policy, Online Reinforcement Learning, non-imitation learning, Multi-environment simulation
% \end{IEEEkeywords}

% For peer review papers, you can put extra information on the cover
% page as needed:
\ifCLASSOPTIONpeerreview
\begin{center} \bfseries EDICS Category: 3-BBND \end{center}
\fi
%
% For peerreview papers, this IEEEtran command inserts a page break and
% creates the second title. It will be ignored for other modes.
\IEEEpeerreviewmaketitle

\section{Introduction}
\label{sec:introduction}

\begin{figure*}[t]
  \centering
  \includegraphics[width=0.99\linewidth]{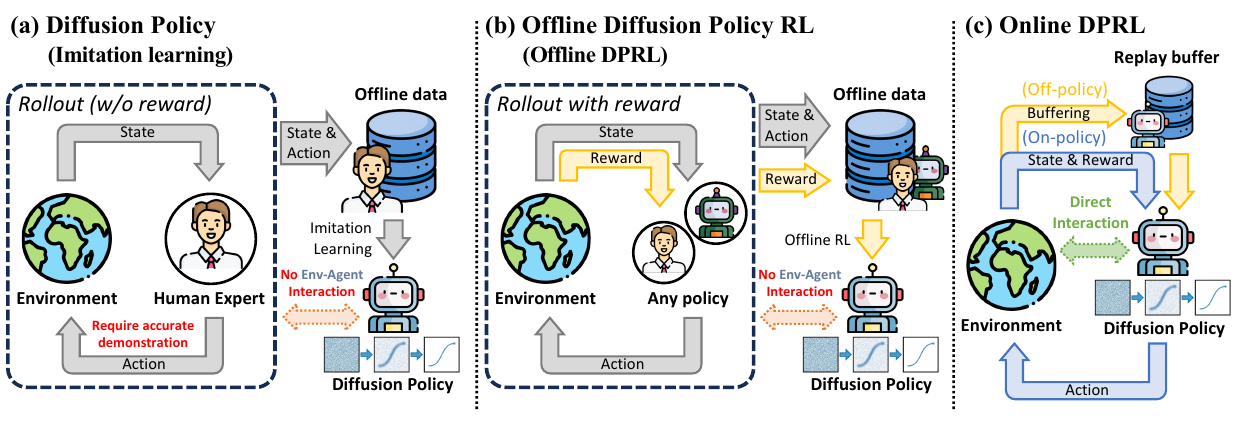}
  \caption{Comparison of training methodologies for Diffusion Policies. (a) Imitation Learning (IL): Most conventional diffusion policies are trained in a supervised manner using expert demonstrations, focusing on behavior cloning. (b) Offline Diffusion Policy Reinforcement Learning (Offline DPRL): Recent approaches leverage additional reward labeling to learn reward signals and optimize policies within a static offline dataset. (c) Online Diffusion Policy RL (Online DPRL): Unlike previous methods, online DPRL enables real-time interaction with the environment, facilitating a more explorative learning process through direct feedback and continuous policy refinement.}
  \label{fig:dprl_paradigm_comparison}
\end{figure*}

\IEEEPARstart{T}{he} advent of diffusion policies~\cite{diffusionpolicy, mpd, diffuser} has marked a breakthrough in Physical AI, demonstrating superior performance over conventional policy networks~\cite{agostini2010reinforcement, sql} in complex robotic manipulation tasks.
% diffusion policies excel over conventional policy networks due to their intrinsic ability to model multimodal action distributions and their stable training dynamics.
This remarkable advancement stems from their inherent capacity to model multimodal action distributions~\cite{diffusionpolicy}, which provides the expressive power necessary to capture the diverse and geometrically constrained action spaces required for real-world tasks.
While diffusion policies excel in modeling complex and diverse robot behaviors, such as navigation~\cite{nomad, navdp}, manipulation~\cite{diffusionpolicy, hdp}, and locomotion tasks~\cite{diffuseloco, ze2025generalizable}, their foundation in imitation learning inherently limits their potential.
Specifically, policies derived via imitation are primarily bound by the optimality of the provided demonstration data~\cite{dppo}, creating a performance ceiling that prevents the agent from autonomously discovering actions superior to expert demonstrations.

To address this optimality gap, recent research has explored combining offline Reinforcement Learning (offline RL) principles with diffusion policies~\cite{diffusionql, consistencyac}.
%, aiming to enhance optimality beyond direct imitation.
This mechanism achieves superior performance and learning efficiency by guiding the policy to prioritize high-reward actions extracted from potentially sub-optimal demonstrations.
% However, Offline RL methods, by exclusively learning from the high-reward actions within a fixed offline dataset, are still incapable of discovering optimal actions outside the dataset or generalizing to out-of-distribution (OOD) scenarios.
Nevertheless, offline RL methods still remain constrained by the fixed nature of static datasets, rendering them incapable of discovering novel optimal actions or generalizing to out-of-distribution (OOD) scenarios.
This inability to adapt to unseen states presents a significant challenge for robotic tasks, which require robustness and strict safety guarantees when deployed in dynamic, real-world environments.

To overcome the lack of autonomous exploration in offline settings, there is a growing necessity for Online Diffusion Policy RL (Online DPRL)~\cite{dipoerror, qvpoerror, dacer, genpo}, which integrates diffusion policies with online RL~\cite{sac, ppo, dsac} to learn through direct interaction with the environment.
Recent advancements in high-fidelity simulation platforms~\cite{mujoco, brax, isaaclab}, which leverage photorealistic rendering and high physical accuracy to reduce the domain gap to the real world significantly, have fueled the feasibility of online learning for diffusion models.
Moreover, the multi-environment parallelization capabilities of these simulators mitigate the multi-step sampling inherent to diffusion models by accelerating data collection. 
% Crucially, the massive parallelization capabilities of these simulators compensate for the sampling latency inherent to diffusion models by accelerating data collection.
This convergence of advanced simulation and continuous policy refinement creates a compelling case for the Online DPRL paradigm.
The distinct advantages of online RL over its offline and imitation learning counterparts include:
\begin{itemize}[leftmargin=9pt, rightmargin=4pt]
    \item \textit{Autonomous Exploration}: Unlike offline learning, online RL enables the discovery of new behaviors through real-time interaction, allowing the agent to explore the action space beyond the scope of initial demonstrations via trial and error.
    \item \textit{Optimal Policy Discovery}: While imitation learning is capped by expert performance, online RL allows the agent to discover near-globally optimal policies, potentially identifying efficient solutions that experts may not have conceived.
    \item \textit{Data Acquisition}: Simulation-based online RL offers a decisive advantage in scenarios where real-world data collection is hazardous or where high-quality expert demonstrations are scarce or expensive.
\end{itemize}

Despite these potential advantages, the integration of diffusion policies with online RL remains relatively underexplored due to fundamental structural conflicts.
The primary challenge lies in the incompatibility between modern RL mechanisms, which often require backpropagation through the policy network, and the surrogate objectives of diffusion models~\cite{ddpm, sgm, fmn}.
First, propagating gradients through the entire reverse diffusion chain---similar to standard policy gradient methods~\cite{policygradient}---is computationally prohibitive and suffers from gradient instability such as vanishing or exploding gradients~\cite{bengio1994learning}.
Second, unlike tractable Gaussian policies~\cite{ppo, agostini2010reinforcement} commonly used in modern RL, diffusion policies lack an analytical expression for likelihood or entropy, making the calculation of tractable objectives difficult.
To tackle these problems, a few nascent online DPRL approaches employ various strategies such as decoupling policy improvement~\cite{dipoerror, DDiffPG} or utilizing surrogate objective reweighting~\cite{qvpoerror, dpmdsdac}.

\begin{figure*}[t]
  \centering
  \includegraphics[width=0.99\linewidth]{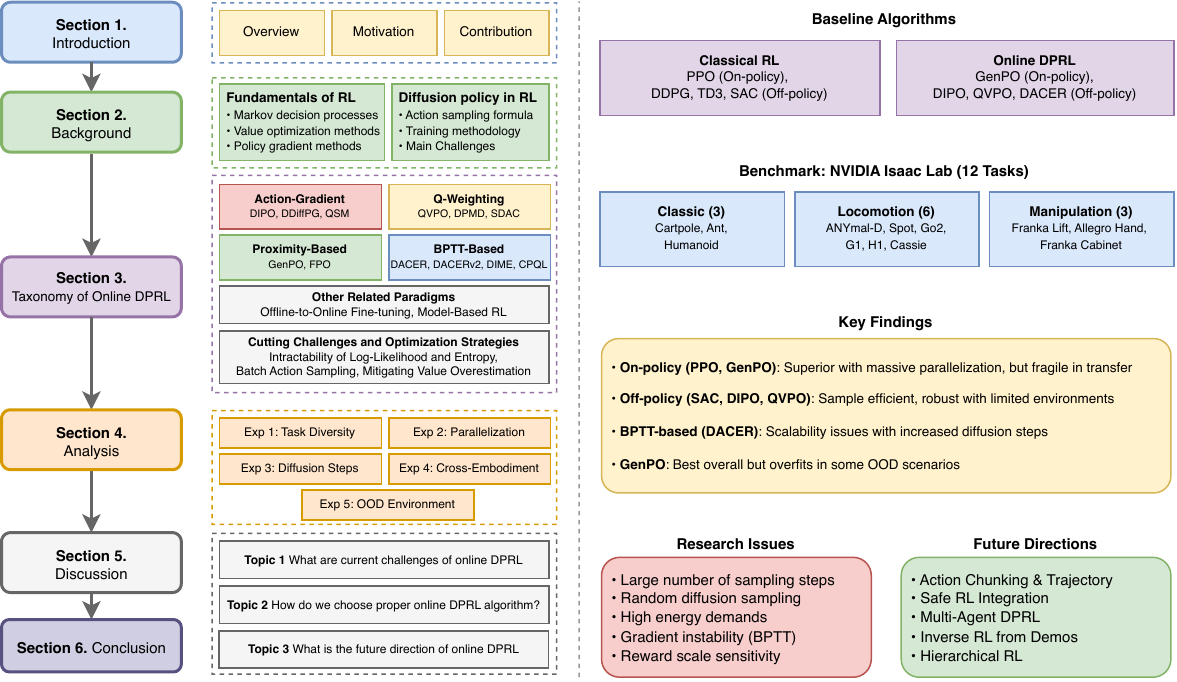}
  \caption{Overview of the paper’s organization and core themes. This review offers a holistic perspective on the intersection of diffusion models and online RL paradigms. By synthesizing empirical analyses and theoretical frameworks, we classify the field's emerging progress into a structured taxonomy, bridging the gap between current technical challenges and their respective solutions.}
  % The diagram illustrates our systematic analysis across six sections, covering the evolution from fundamental progress (Section 2) to current challenges and solutions (Sections 3--5), and future research opportunities (Section 6).}
  \label{fig:overview}
\end{figure*}
% Fig. 3. The taxonomy of this survey. We provide a comprehensive overview of the current research landscape
% at the intersection of DMs and RL by systematically analyzing this emerging field’s progress, challenges,
% solutions, and opportunities.

This paper specifically focuses on \textit{online model-free reinforcement learning} approaches that employ diffusion-based policy representations.
These online model-free methods offer greater generality across domains without requiring learned dynamics models, making them more suitable for deployment in diverse robotic scenarios.
We establish the theoretical foundations of RL and diffusion policies in Sec.~\ref{sec:background}, then propose a comprehensive taxonomy in Sec.~\ref{sec:literature_review} that categorizes existing Online DPRL methods into four distinct groups---Action-gradient, Q-weighting, Proximity-based, and BPTT-based methods---critically examining how each approach circumvents the intractability of policy improvement.
Sec.~\ref{sec:analysis} presents an empirical analysis within a unified NVIDIA Isaac Lab benchmark~\cite{isaaclab}, which is a GPU-accelerated robotics simulation framework, integrating both classical RL algorithms~\cite{sac, ppo} and representative Online DPRL methods~\cite{dipoerror, qvpoerror, genpo, dacer} to investigate generalization, efficiency, and robustness across diverse robotic scenarios.
% Finally, Section~\ref{sec:discussion} synthesizes our findings and provides a practical rationale for algorithm selection based on specific operational constraints, while identifying current computational and algorithmic bottlenecks.
Finally, Sec~\ref{sec:discussion} synthesizes our findings to provide a practical rationale for algorithm selection under specific operational constraints, while identifying current computational bottlenecks and outlining promising future research directions to overcome these limitations.
A schematic overview of the paper’s structure and its central pillars is presented in Fig.~\ref{fig:overview}.
% The contributions of our paper are as follows:
% \begin{itemize}
%     \item We propose a novel taxonomy of Online DPRL methods based on their specific policy improvement mechanisms, providing a structured overview of the current research landscape.
%     \item We present a unified, open-source benchmark built on a robotic simulation benchmark to rigorously evaluate the generalizability, efficiency, and robustness of online DPRL across diverse robotic scenarios.
%     \item We identify key computational bottlenecks in current online DPRL, providing concrete guidelines for algorithm selection in practical robot learning scenarios.
% \end{itemize}
% \input{scripts/2_background}
\begin{table}[!t]
    \caption{General notations.}
    \centering
    % \resizebox{1.00\linewidth}{!}{
    \label{tab:notation}
    \begin{tabular}{l|l}
         \toprule
         % \multicolumn{2}{c}{Notation}  \\
         % \midrule
         % dataset, set (index, task, etc) state, in/output
         % normal font
         \multicolumn{2}{c}{Diffusion Network} \\
         \midrule
         Gaussian Noise & $\epsilon~(\text{or}~a^K)$ \\
         Gaussian distribution & $\mathcal{N}(\mu, \sigma^2)$ \\
         Discrete Uniform distribution & $\mathcal{U}(x_1, x_2)$ \\
         Continuous Uniform distribution & $U(x_1, x_2)$ \\
         noise estimator & $\boldsymbol{\epsilon}_\theta(\cdot)$ \\
         score estimator & $\mathbf{s}_\theta(\cdot)$ \\
         velocity estimator & $\mathbf{v}_\theta(\cdot)$ \\
         timestep & $k$ \\
         % Image & $\mathbf{I}~:~\mathbb{R}^{3 \times H \times W}$ \\
         % $i$-th image & $\mathbf{I}^i~:~\mathbb{R}^{3 \times H \times W}$ \\
         % Image set in mini-batch & $\mathbf{I}_{b}~:~\mathbb{R}^{B \times 3 \times H \times W}$ \\
         % Camera intrinsic matrix & $\mathbf{K}$ \\
         % Coarse projected 3D center & $(u,v)$ \\
         % Object depth & $z$ \\
         % Center offset & $(\delta u, \delta v)$ \\
         % 3D size dimensions & $(h,w,l)$ \\
         % Heading direction & $\gamma$ \\
         % Number of training images & $M$ \\
         % Number of objects in $\mathbf{I}^i$ & $N^i$ \\
         % Number of objects in entire images & $L$ \\
         \midrule
         \toprule
         \multicolumn{2}{c}{Reinforcement Learning} \\
         \midrule
         State, Next state & $s$, $s' \in \mathcal{S}$ \\
         Action, Next action & $a$ (or $a^0$), $a' \in \mathcal{A}$ \\
         Action under the policy $\pi$ & $\hat{a}$ \\
         Reward & $r \in \mathcal{R}$ \\
         Discount factor & $\gamma$ \\
         Trajectories & $\tau = \{s_0, a_0, r_0, s_1, \cdots\}$ \\
         Return & $G$ \\
         Replay buffer (or offline dataset) & $\mathcal{D}$ \\
         Policy & $\pi(a|s)$ \\
         Parameterized policy (Actor) & $\pi_\theta(a|s)$ \\
         % Policy Network (Actor) & $\pi_\theta(\cdot)$ \\
         % behavior policy & $\pi^b_\theta(\cdot)$ \\ % optional 빼도됨
         State value & $V(s)$ \\
         Q-function (Action value) & $Q(s, a)$ \\
         Paramaeterized Q-function (Critic) & $Q_\phi(s, a)$ \\
         Action value under the policy $\pi$ & $Q^\pi(s, a)$ \\
         % Task set & $\mathbf{T}=\{t_{c}, t_{u}, t_{v}, t_{\delta u}, t_{\delta v}, t_{z}, t_{h}, t_{w}, t_{l}, t_{\gamma}\} \ni t$ \\
         % Image index set & $I = \{1, 2, \dots, M\}$ \\
         % object index set in image $\mathbf{I}^i$ & $J = \{1, 2, \dots, N^i\}$ \\
         % Feature maps (hidden state) & $\mathbf{h}~:~\mathbb{R}^{C \times H' \times W'}$\\
         % Per-pixel output maps & $\Tilde{\mathbf{o}}_{\mathbf{T}}~:~\mathbb{R}^{|\mathbf{T}| \times H' \times W'}$ \\
         % Object-wise output (prediction) & $\mathbf{o}_{\mathbf{T}}~:~\mathbb{R}^{|\mathbf{T}| \times N^i}$ \\
         % Object descriptor & $\rho~:~\mathbb{R}^{C \times 1 \times 1}$ \\
         % Distance metric & $d(\cdot, \cdot)$ \\
         % Object descriptor metric space & $(\mathbf{P}, d) \equiv \mathbf{P}\text{-space}$ \\
         % Object depth metric space & $(\mathbf{Z}, d) \equiv \mathbf{Z}\text{-space}$ \\
         % \midrule
         % \toprule
         % \multicolumn{2}{c}{Network components, Function} \\
         % \midrule
         % Feature extractor & $\mathcal{F}_{\theta}(\cdot)~:~\mathbb{R}^{3 \times H \times W} \rightarrow \mathbb{R}^{C \times H' \times W'}$ \\
         % Task-specific lightweight head & $\mathcal{G}_{\phi^{t}}(\cdot)~:~\mathbb{R}^{C \times H' \times W'} \rightarrow \mathbb{R}^{H' \times W'}$ \\
         % Extract function & $\mathcal{H}(\cdot)~:~\mathbb{R}^{|\mathbf{T}| \times H' \times W'} \rightarrow \mathbb{R}^{|\mathbf{T}| \times N^i}$ \\
         % % Quasi-isometry & $\mathcal{Q}(\cdot)~:~\mathbb{R}^{L} \rightarrow \mathbb{R}^{L}$ \\
         % Quasi-isometry & $\mathcal{Q}(\cdot)~:~\mathbb{R} \rightarrow \mathbb{R}^{C}$ \\
         \bottomrule
    \end{tabular}%}
\end{table}

\section{Background}
\label{sec:background}

In this section, we provide a brief introduction to fundamental reinforcement learning concepts and standard model-free online RL algorithms to establish the foundation for discussing online diffusion policy RL literature (Sec~\ref{subsec:rl_fundamentals}).
We also present a mathematical formulation of diffusion policies in RL environments (Sec~\ref{subsec:diffusion_rl}).
For clarity, we unify the mathematical notation as presented in Tab.~\ref{tab:notation}.

\subsection{Reinforcement Learning Fundamentals}
\label{subsec:rl_fundamentals}

Reinforcement learning (RL) formulates the problem of learning optimal sequential decisions through interaction with an environment. This is formally defined as a Markov Decision Process (MDP) tuple $\mathcal{M} = (\mathcal{S}, \mathcal{A}, \mathcal{R}, \gamma)$, where $\mathcal{S}$ and $\mathcal{A}$ denote the state and action spaces, $\mathcal{R}: \mathcal{S} \times \mathcal{A} \rightarrow \mathbb{R}$ is the reward function, and $\gamma \in [0,1)$ is the discount factor.
The core objective of an RL agent is to learn an optimal policy $\pi(a|s)$, which is a probability distribution over actions conditioned on a state, that maximizes the expected cumulative discounted return $G = \sum_{t=0}^{\infty} \gamma^t r_t$.

\subsubsection{Value Optimization in RL}
The action-value function (Q-function) $Q^\pi(s, a)$ estimates the expected return of taking action $a$ in state $s$ and following policy $\pi$ thereafter.
It satisfies the Bellman equation, which recursively relates the current value to future values:
\begin{equation}
    Q^\pi(s,a) = \mathbb{E}_{s',r} \left[ r + \gamma \mathbb{E}_{a' \sim \pi(\cdot|s')} [Q^\pi(s',a')] \right].
\end{equation}
For the optimal policy, the Bellman optimality equation holds: $Q^*(s, a) = \mathbb{E}_{s',r} [r + \gamma \max_{a'} Q^*(s', a')]$.
Given the optimal Q-function $Q^*(s, a)$, the optimal policy $\pi^*(a|s)$ can be derived by greedily selecting the action with the highest value:
\begin{equation}
    \label{eq:optimal_policy}
    \pi^*(s) = \underset{a}{\arg\max} \ Q^*(s,a).
\end{equation}

Despite its solid mathematical foundations, reinforcement learning struggled with approximating Q-functions in high-dimensional state spaces.
Deep Q-Networks (DQN)~\cite{dql} addressed this challenge by employing neural networks as Q-function approximators, thereby enabling effective learning in high-dimensional state spaces with discrete action spaces.

\subsubsection{Policy Gradient Methods}
\label{subsec:policy_gradient}
However, in tasks requiring continuous action spaces, directly solving Eq.~\ref{eq:optimal_policy} is computationally intractable because finding the global maximum over a continuous domain at every step is prohibitively expensive.
This necessitates the use of parameterized policy networks $\pi_\theta$ and policy gradient methods~\cite{suttonbible}.
Policy gradient methods directly optimize the expected return $J(\theta) = \mathbb{E}_{\tau \sim \pi_\theta}[G]$ via gradient ascent on the policy parameters $\theta$.
The foundational Policy Gradient Theorem~\cite{policygradient, gae} enables gradient estimation using sampled trajectories $\tau = \{s_0, a_0, r_0, s_1, a_1, \cdots\}$ as follows:
\begin{equation}
\label{eq:policy_gradient}
    \nabla_\theta J(\theta) = \mathbb{E}_{\tau \sim \pi_\theta} \left[ \sum_{t=0}^{\infty} \nabla_\theta \log \pi_\theta(a_t|s_t) \cdot Q^{\pi_\theta}(s_t,a_t) \right].
\end{equation}

\subsubsection{Taxonomy of Deep Reinforcement Learning}
Most standard Deep Reinforcement Learning (DRL) algorithms utilize Actor-Critic architectures, leveraging both a parameterized policy $\pi_\theta$ (actor) and a parameterized value function $Q_\phi$ (critic).
These DRL algorithms employ policy gradients for policy optimization and are categorized by their value optimization methods: (a) Standard RL, (b) Maximum Entropy RL, and (c) Distributional RL.

\paragraph{Standard RL}
Standard RL represents the most prevalent approach adopted by many foundational methods~\cite{dql, ddpg, ppo}.
To maximize the expected return $G$, this approach leverages the Bellman equation to learn the Q-function by minimizing the Temporal Difference (TD) error between the predicted action-value $Q_\phi(s, a)$ and the target action-value $\hat{Q}(s, a)$:
\begin{equation}
\begin{gathered}
\label{eq:standard_rl_loss}
    \mathcal{L}_\phi^\text{Standard} = \mathbb{E}_{s,a,r,s'} \left[ \|Q_\phi(s, a) - \hat{Q}(s, a) \|^2 \right], \\
    \hat{Q}(s, a) = r + \gamma Q_{\phi'}(s', a'), \quad a' \sim \pi_\theta(\cdot|s'),
\end{gathered}
\end{equation}
where the current state-action pair $(s, a)$ and next state-action pair $(s', a')$ can be sampled either from the current policy $\pi$ (on-policy) or from a replay buffer $\mathcal{D}$ (off-policy).
For training stability, a slowly-updated target network $Q_{\phi'}$ using exponential moving average (EMA) is commonly employed to compute the target action-value.

\paragraph{Maximum Entropy RL (MaxEnt RL)}
Maximum Entropy RL~\cite{sac, sql} augments the standard objective by incorporating a weighted policy entropy term $\beta \log \pi(a|s)$ to encourage exploration and prevent premature convergence to suboptimal policies.
The action-value function is learned by minimizing the residual of the Soft Bellman Eq.~\cite{sql}:
\begin{equation}
\begin{gathered}
    \mathcal{L}_\phi^\text{MaxEnt} = \mathbb{E}_{s,a,r,s'} \left[ \| Q_\phi(s, a) - \hat{Q}(s, a) \|^2 \right], \\
    \hat{Q}(s, a) = r + \gamma(Q_{\phi'}(s', a') - \beta \log \pi_\theta(a'|s')), ~ a' \sim \pi_\theta(\cdot|s'),
\end{gathered}
\end{equation}
where $\beta > 0$ is the temperature parameter controlling the trade-off between reward maximization and entropy regularization.

\paragraph{Distributional RL (Dist-RL)}
In contrast to standard approaches, Distributional RL~\cite{dsac, c51} shifts the focus from estimating the expected return $Q(s, a)$ to modeling the complete probability distribution of returns $Z(s, a)$, which satisfies $Q(s, a) = \mathbb{E}_{Z}[Z(s, a)]$.
This enables the agent to reason about risk, uncertainty, and potential extreme outcomes.
The return distribution $Z_\phi(s, a)$ is learned by minimizing the distance between the predicted distribution and the target distribution $\hat{Z}(s, a)$:
\begin{equation}
\begin{gathered}
    \mathcal{L}_\phi^\text{Dist} = \mathbb{E}_{s,a,r,s'} \left[ d_\text{Dist}(Z_\phi(s, a), \hat{Z}(s, a)) \right], \\
    \hat{Z}(s, a) = r + \gamma Z_{\phi'}(s', a'), \quad a' \sim \pi_\theta(\cdot|s'),
\end{gathered}
\end{equation}
where $d_\text{Dist}$ is a distributional distance metric, such as the Kullback-Leibler divergence~\cite{kldivergence} or the Wasserstein distance~\cite{wasersteindistance}.

\subsection{Diffusion Policies in Reinforcement Learning}
\label{subsec:diffusion_rl}

Standard RL implementations model policies as unimodal Gaussians or deterministic mappings~\cite{ppo, ddpg, sac}, fundamentally limiting their ability to represent multimodal behavior distributions required for complex robotic tasks.
While Gaussian Mixture Models~\cite{agostini2010reinforcement} and energy-based policies~\cite{sql, florence2022implicit} attempted to address this, they suffer from training instability or high inference costs.
Diffusion Policies~\cite{diffusionpolicy}, formulating policies as conditional denoising processes, have emerged as a powerful alternative offering high expressiveness and training stability.
They have demonstrated remarkable success in visuomotor control~\cite{hdp, diffuseloco, ze2025generalizable}, effectively handling high-dimensional observations and actions.

% Standard implementations of RL algorithms typically model the policy as a unimodal Gaussian distribution or a deterministic mapping. 
% However, this assumption fundamentally limits their ability to represent complex, multimodal behavior distributions inherent in many real-world tasks, such as robotic manipulation with multiple valid solution strategies.
% To address this limitation, previous studies have attempted to approximate multimodal distributions using Gaussian Mixture Models (GMMs)~\cite{agostini2010reinforcement} or implicit energy-based policies~\cite{sql, florence2022implicit}. 
% However, these approaches often suffer from training instability, mode collapse, or prohibitively high inference costs.

% Recently, diffusion Policies~\cite{diffusionpolicy}, which formulate the policy as a conditional denoising diffusion process, have emerged as a powerful alternative.
% Diffusion policies offer distinct advantages, including high expressiveness for multimodal distributions and training stability due to their regression-based objective.
% In the domain of robot learning, diffusion policies have demonstrated remarkable success in visuomotor control tasks~\cite{hdp, diffuseloco, ze2025generalizable}, effectively handling high-dimensional observation spaces and complex manipulation behaviors.

\subsubsection{Action Generation via Diffusion Sampling}
\label{subsubsec:policy_sampling}

A diffusion policy represents the parameterized policy $\pi_\theta(a|s)$ as a conditional diffusion model that synthesizes actions by iteratively denoising a sample from a Gaussian distribution, conditioned on the state $s$.
During online interaction, the agent generates an action by executing the reverse diffusion process.
Starting from pure noise $a^K \sim \mathcal{N}(\mathbf{0},\mathbf{I})$, the model progressively refines the action over $K$ timesteps using the learned noise prediction network $\boldsymbol{\epsilon}_\theta$.
The generation process iteratively applies the following update rule:
\begin{equation}
\label{eq:diffusion_policy_sampling}
    a^{k-1} \leftarrow \frac{1}{\sqrt{\alpha_k}} \left( a^k - \frac{1-\alpha_k}{\sqrt{1-\bar{\alpha}_k}} \boldsymbol{\epsilon}_{\theta}(a^k, s, k) \right) + \sigma_k \epsilon_i,
\end{equation}
where $k$ denotes the diffusion timestep decreasing from $K$ to $0$, $\epsilon_i \sim \mathcal{N}(\mathbf{0}, \mathbf{I})$ represents Gaussian noise scaled by $\sigma_k \in \mathbb{R}$, and $\alpha_k, \bar{\alpha}_k \in \mathbb{R}$ are coefficients determined by a predefined noise schedule~\cite{song2020denoising}.
This iterative denoising process allows the agent to construct precise actions conditioned on the current environmental state $s$.

\subsubsection{Training Diffusion Policies via Imitation Learning}
Diffusion policies have been predominantly trained via Imitation Learning (IL) using offline demonstration datasets.
Given a dataset $\mathcal{D} = \{(s, a^*)\}$ of state-action pairs from expert demonstrations, the noise prediction network $\boldsymbol{\epsilon}_\theta$ is optimized to predict the Gaussian noise $\epsilon$ added to clean actions $a^*$ at diffusion timestep $k$ by minimizing the diffusion loss:
\begin{equation}
\label{eq:diffusion_policy_bc}
    \mathcal{L}(\theta) = \underset{\substack{(s,a^*) \sim \mathcal{D}, \\k \sim \mathcal{U}(1,K)}}{\mathbb{E}} \left[ \left\| \epsilon - \boldsymbol{\epsilon}_\theta(\sqrt{\bar{\alpha}_k} a^* + \sqrt{1-\bar{\alpha}_k} \epsilon, s, k) \right\|^2 \right].
\end{equation}
Through this objective, the diffusion policy learns the expert's behavior distribution in a supervised regression manner, avoiding the complexities of reinforcement learning.

\subsubsection{Fundamental Challenges of Integration between Diffusion Policy and Online RL}
\label{subsubsec:major_challenges}
While diffusion policies excel in imitation learning scenarios where ground-truth expert actions $a^*$ are directly provided, their application to online reinforcement learning introduces fundamental algorithmic challenges.
Unlike supervised imitation learning, online RL requires the policy to improve through environmental feedback (rewards) without access to optimal target actions, necessitating gradient-based policy improvement.
% While diffusion policies excel in imitation learning by leveraging ground-truth expert actions $a^*$, their integration into online RL introduces significant algorithmic hurdles.
% Unlike supervised paradigms, online RL relies on environmental rewards rather than optimal targets, necessitating efficient gradient-based policy improvement.

The core difficulty arises when attempting to apply standard policy gradient methods (Eq.~\ref{eq:policy_gradient}) to diffusion policies.
To compute the policy gradient $\nabla_\theta J(\theta)$, one must differentiate through the action generation process described in Eq.~\ref{eq:diffusion_policy_sampling}.
However, since the diffusion policy generates actions via an iterative chain of $K$ denoising steps (where $K$ is typically large, e.g., 50-100), this requires backpropagating gradients through the entire reverse process chain.
This approach is problematic for two main reasons:
% The primary challenge lies in differentiating through the iterative action generation process (Eq.~\ref{eq:diffusion_policy_sampling}) to compute the policy gradient $\nabla_\theta J(\theta)$.
% Backpropagating through a $K$-step denoising chain—where $K$ typically ranges from 50 to 100—poses two critical issues:
\begin{enumerate}
    \item \textbf{Computational Inefficiency:} Constructing the computation graph for the entire $K$-step chain at every policy update is computationally prohibitive and memory-intensive, especially when $K$ is large.
    \item \textbf{Optimization Instability:} Backpropagating through a long chain of operations often leads to vanishing or exploding gradient problems~\cite{bengio1994learning}, degrading the learning signal for early denoising steps and destabilizing policy optimization.
\end{enumerate}
Consequently, naive integration of diffusion models into standard online actor-critic frameworks is impractical, necessitating novel and efficient gradient estimation techniques tailored for diffusion policies.
\section{Taxonomy of Online DPRL}
\label{sec:literature_review}

\begin{table*}[!ht]
    \caption{Taxonomy of online Diffusion Policy Reinforcement Learning (DPRL). For comparative clarity, non-diffusion RL algorithms are denoted in gray italics. Representative online DPRL methods evaluated in our experiments are highlighted with an asterisk (*).}
    \centering
    \resizebox{0.99\linewidth}{!}{
    \label{tab:taxonomy}
    \begin{tabular}{llllll}
         \toprule
         Algorithms & RL objective & On/off-policy & Performance evaluation & Compared RL baselines \\
         \midrule
         \textcolor{gray}{\textit{PPO}}~\cite{ppo} & Standard & on-policy & MuJoCo~\cite{mujoco} & PG~\cite{policygradient}, TRPO~\cite{trpo}, CEM~\cite{cem}, A2C~\cite{a2c}, ACER~\cite{acer} \\
         \textcolor{gray}{\textit{DDPG}}~\cite{ddpg} & Standard & off-policy & MuJoCo, Torcs~\cite{torcs} & DPG~\cite{dpg}, iLQG~\cite{ilqg} \\
         \textcolor{gray}{\textit{TD3}}~\cite{td3} & Standard & off-policy & MuJoCo & PPO, DDPG, TRPO, ACKTR~\cite{acktr}, SAC \\
         \textcolor{gray}{\textit{SAC}}~\cite{sac} & MaxEnt & off-policy & MuJoCo, RLLab~\cite{rllab} & PPO, DDPG, SQL~\cite{sql}, TD3 \\
         \midrule
         *DIPO~\cite{dipoerror} & Standard & off-policy & MuJoCo & PPO, TD3, SAC \\
         DDiffPG~\cite{DDiffPG} & Standard & off-policy & D4RL~\cite{d4rl}, Panda-Gym~\cite{pandagym} & TD3, SAC, Diffusion-QL~\cite{diffusionql}, Consistency-AC~\cite{consistencyac}, RPG~\cite{rpg}, DIPO \\
         QSM~\cite{qsm} & Standard & off-policy & DMC~\cite{dmcontrol} & TD3, SAC \\
         \midrule
         *QVPO~\cite{qvpoerror} & Standard & off-policy & MuJoCo & PPO, TD3, SAC, DIPO, QSM, HD~\cite{hd} \\
         DPMD~\cite{dpmdsdac} & Standard & off-policy & MuJoCo & PPO, TD3, SAC, QSM, DIPO, DACER, QVPO, DPPO~\cite{dppo} \\
         SDAC~\cite{dpmdsdac} & MaxEnt & off-policy & MuJoCo & PPO, TD3, SAC, QSM, DIPO, DACER, QVPO, DPPO \\
         \midrule
         *GenPO~\cite{genpo} & Standard & on-policy & Isaac Lab~\cite{isaaclab} & PPO, DDPG, TD3, SAC, DACER, QVPO \\
         FPO~\cite{fpo} & Standard & on-policy & DMC & PPO, DPPO \\
         \midrule
         *DACER~\cite{dacer} & Standard+MaxEnt & off-policy & MuJoCo & PPO, TRPO~\cite{trpo}, DDPG, TD3, SAC, DSAC~\cite{dsac} \\
         DACERv2~\cite{dacerv2} & Standard+MaxEnt & off-policy & MuJoCo & PPO, SAC, DSAC, DIPO, DIME, DACER, QVPO \\
         CPQL~\cite{cpql} & Standard & off-policy & MuJoCo, DMC & PPO, MPO~\cite{mpo}, DMPO~\cite{dmpo}, D4PG~\cite{d4pg}, TD3, SAC, DIPO, DreamerV3~\cite{dreamerv3} \\
         DIME~\cite{dime} & MaxEnt & off-policy & MuJoCo, DMC, MyoSuite~\cite{myosuite} & CrossQ~\cite{crossq}, BRO~\cite{bro}, QSM, DIPO, QVPO, DACER\\
         DSAC-D~\cite{dsacd} & Distributional & off-policy & MuJoCo, Vehicle Meeting~\cite{dsacd} & PPO, TRPO, DDPG, TD3, SAC, DSAC-T~\cite{dsact}, DACER \\
         \bottomrule
    \end{tabular}}
\end{table*}
\begin{figure}[t]
  \centering
  \includegraphics[width=0.99\columnwidth]{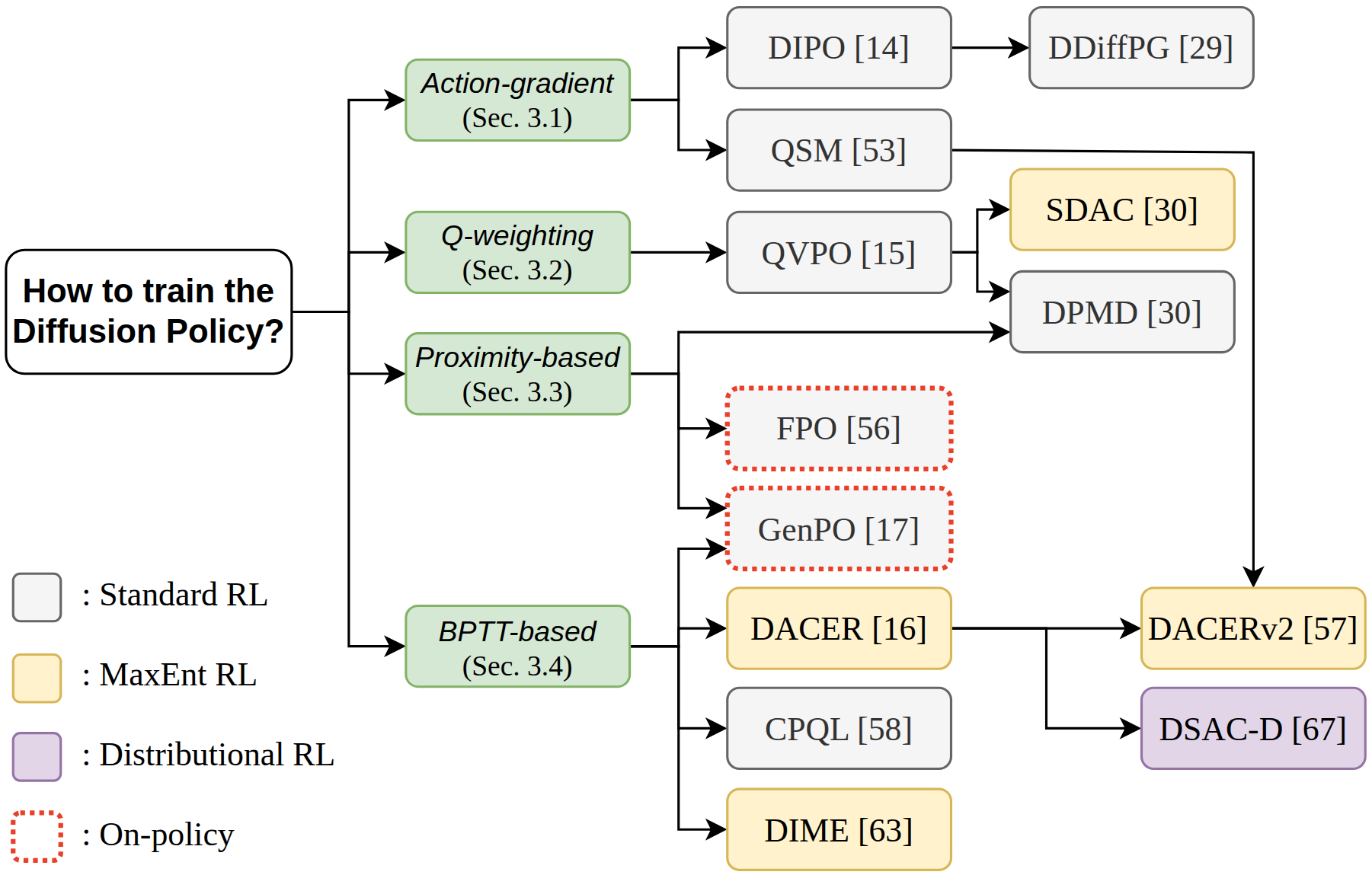}
  \caption{Taxonomy of online Diffusion Policy Reinforcement Learning Paradigm. We categorize existing online DPRL approaches into four distinct families based on their policy training mechanisms: Action-Gradient, $Q$-weighting, Proximity-based, and BPTT-based methods.}
  \label{fig:taxonomy}
\end{figure}

% \subsection{Taxonomy of Online model-free DPRL}

Existing online model-free DPRL methods tackle the challenges outlined in Sec.~\ref{subsubsec:major_challenges} through diverse algorithmic strategies.
We classify these approaches into four distinct families based on their core mechanism for policy improvement: \textit{(1) Action-Gradient-Based Methods} (Sec.~\ref{subsec:action_gradient}), \textit{(2) Q-Weighting Methods} (Sec.~\ref{subsec:q-weighting}), \textit{(3) Proximity-Based Methods} (Sec.~\ref{subsec:proximity}), and \textit{(4) Backpropagation Through Time (BPTT)} (Sec.~\ref{subsec:bptt}).
Each family employs fundamentally different techniques to circumvent the computational and optimization difficulties inherent in training diffusion policies with online RL.
In the following sections, we provide a comprehensive review organized by this taxonomy, with a particular focus on the mathematical formulations of policy improvement.
Furthermore, we summarize these concepts in the following sections: Sec.~\ref{subsec:related} categorizes the diverse Diffusion-based RL paradigms by analyzing how diffusion models are integrated into online policy optimization.
Sec.~\ref{subsec:crosscutting} then identifies the cross-cutting challenges of online DPRL and discusses key optimization strategies to improve computational efficiency and training stability.
A consolidated summary of online model-free DPRL approaches is presented in Fig.~\ref{fig:taxonomy} and Tab.~\ref{tab:taxonomy}.

%------------------------------------------------------------------------------

\subsection{Action-Gradient-Based Methods}
\label{subsec:action_gradient}
To apply diffusion policy to model-free online RL tasks, several pioneering approaches~\cite{dipoerror, qsm} leverage the \textit{action gradient}, which provides a novel approach for policy improvement as follows:
\begin{equation}
    \text{(action gradient)} \triangleq \nabla_a Q(s, a).
\end{equation}
The action gradient indicates the direction in which an action should be improved according to the state-action value function $Q$, which evaluates the quality of actions. The methods in this section directly utilize action gradients to perform policy improvement.

\subsubsection{Model-free Online RL with \textbf{DI}ffusion \textbf{PO}licy (\textbf{DIPO})}

DIPO~\cite{dipoerror} is the pioneering work in online Diffusion Policy RL, introducing the action gradient method for policy optimization.
By first improving actions through action gradients and then using these improved actions as targets for diffusion model training, DIPO avoids the computational burden of backpropagating through the entire diffusion chain.

Given sampled state-action pairs $(s, a) \in \mathcal{D}$, DIPO computes an improved action $\hat{a}$ using action gradient ascent as follows:
\begin{align}
\begin{split}
\label{eq:action_improve}
    \hat{a} = a + \eta_a\nabla_a Q(s, a),
\end{split}
\end{align}
where $\eta_a > 0$ is the learning rate of action-gradient.
These improved state-action pairs $(s, \hat{a})$ are stored in the replay buffer $\mathcal{D}$ and used to train the diffusion policy in an off-policy manner as Eq.~\ref{eq:dipo} as follows:
\begin{equation}
\label{eq:dipo}
    \mathcal{L}_\theta^{\text{DIPO}} = \underset{\substack{(s,\hat{a}) \sim \mathcal{D}, \\k \sim \mathcal{U}(1,K)}}{\mathbb{E}} \left[ \left\| \epsilon - \boldsymbol{\epsilon}_\theta(\sqrt{\bar{\alpha}_k} \hat{a} + \sqrt{1-\bar{\alpha}_k} \epsilon, s, k) \right\|^2 \right].
\end{equation}
By separating policy improvement in the replay buffer from policy model training, the model can learn actions with higher state-action values.
However, replacing $\hat{a}$ with the original action $a$ from the buffer causes misalignment with the current MDP dynamics and reward function, resulting in training the Q-function being problematic.
% However, this two-stage approach introduces additional training complexity, requiring careful balancing between action gradient updates and diffusion model training.
% Furthermore, the method's performance heavily depends on the accuracy of the Q-function estimates, and the additional action gradient computation increases overall training time compared to end-to-end approaches.

\subsubsection{\textbf{D}eep \textbf{Diff}usion \textbf{P}olicy \textbf{G}radient (\textbf{DDiffPG})}
DDiffPG~\cite{DDiffPG} extends the methodology of DIPO to actively promote multimodal behaviors.
Specifically, this approach collects multiple trajectories through rollouts and extracts $M$ representative modes using hierarchical clustering~\cite{hierarchicalclustering}.
Each clustered trajectory set, denoted as $\tau_m$, then undergoes a process of action improvement similar to Eq.~\ref{eq:action_improve}, utilizing a mode-specific parameterized Q-function $Q_{\phi_m}(s, a)$ as follows:
% DDiffPG~\cite{} 는 DIPO의 방법론을 개량하여 multimodal behaviors를 장려하기 위한 방법론이다.
% 이 방법은 여러개의 trajectories를 rollout을 통해 수집한 뒤 hierachical clustering~\cite{}을 이용해 $M$개의 representative modes를 추출한다.
% 그리고 이렇게 clustering된 각 trajectories set $\tau_m$는 각각 mode-specific Q-function $Q_{\phi_m}(s, a)$을 활용하여 equation~\ref{eq:action_improve}와 비슷하게 action improvement과정을 거친다 as follows:
\begin{equation}
\begin{gathered}
    \hat{a}_m = a_m + \eta_a\nabla_a Q_{\phi_m}(s, a_m), \\
    \forall~1 \leq m \leq M,~ a_m \in \tau_m.
\end{gathered}
\end{equation}
% DIPO의 objective와 비슷하게 이렇게 만들어진 $\hat{a}_m$은 diffusion policy update에 사용된다 as follows:
Similar to DIPO, the multi-modal actions $\{\hat{a}_m\}_{m=1}^M$ are used for updating the diffusion policy as follows:
\begin{equation}
\label{eq:ddiffpg}
    \mathcal{L}_\theta^{\text{DDiffPG}} = \underset{\substack{s \sim \mathcal{D}, \hat{a}_m \\k \sim \mathcal{U}(1,K)}}{\mathbb{E}} \left[ \left\| \epsilon - \boldsymbol{\epsilon}_\theta(\sqrt{\bar{\alpha}_k} \hat{a}_m + \sqrt{1-\bar{\alpha}_k} \epsilon, s, k) \right\|^2 \right].
\end{equation}

% By employing these learned multi-mode actions $\{\hat{a}_m\}_{m=1}^{M}$ in a manner consistent with behavior cloning, DDiffPG conducts multimodal training, thereby empowering the agent to exhibit diverse, multimodal action distributions.
By incorporating these learned multi-modal actions $\{\hat{a}_m\}_{m=1}^M$ into a unified training batch, DDiffPG employs a behavior cloning objective to capture complex action distributions.
This batch-wise optimization effectively empowers the agent to model and exhibit highly diverse, multimodal behaviors.
% This optimization method allows the agent to effectively capture and exhibit complex, multimodal action distributions.
% 이렇게 학습된 각 multi-mode action $\{\hat{a}_m\}_{m=1}^{M}$을 이용하여 behavior cloning과 같은 manner로 multimodal training을 진행함으로써 DDiffPG는 agent의 multimodal action을 장려한다.

\subsubsection{\textbf{Q}-\textbf{S}core \textbf{M}atching (\textbf{QSM})}
Compared to DIPO and DDiffPG, QSM~\cite{qsm} proposes a more direct approach to training diffusion policies using action gradients through Q-score matching.
QSM theoretically proves that aligning the score function of score-based generative models with action gradients leads to global improvement of the policy's action-value function. Given score estimator $\mathbf{s}_\theta(\cdot, \cdot)$ and action gradient $\nabla_a Q(s, a)$, the empirical loss is:
\begin{equation}
    \mathcal{L}_\theta^\text{QSM} = \underset{(s, a) \sim \mathcal{D}}{\mathbb{E}}\|\mathbf{s}_\theta(s, a) - \alpha \nabla_a Q(s, a)\|^2.
\end{equation}
By directly matching the score function to action gradients, QSM enables the diffusion policy to implicitly learn the optimal Boltzmann distribution without explicit imitation learning.
However, since this method directly matches the score function with action gradients, it is highly sensitive to the accuracy of the Q-function.
As approximating accurate Q-functions in most robotic tasks is extremely challenging, QSM poses a potential risk of yielding suboptimal performance.

%------------------------------------------------------------------------------

\subsection{$Q$-weighting Methods}
\label{subsec:q-weighting}

While action-gradient-based methods directly utilize Q-gradients for policy training, $Q$-weighting methods take a different approach by modulating the diffusion loss according to action quality, thereby implicitly guiding the policy toward high-reward regions.
Each method uses its own unique weighting function $\mathbf{w}(s, a)$---which is related to action value $Q$---and demonstrates through theoretical analysis that policy improvement is possible when optimizing the surrogate diffusion loss with the weighting function.
Given a weighting function $\mathbf{w}(s, a)$, the general form of the empirical loss function for the diffusion model is:
\begin{equation}
    % \mathcal{L}_\theta^\text{Q-weight} = \underset{\substack{s \sim \text{env} \\ a \sim \pi_\theta}}{\mathbb{E}}
    \mathcal{L}_\theta^\text{$Q$-weight} = \underset{s,a}{\mathbb{E}}
    \left[
    \mathbf{w}(s, a) \cdot \text{(surrogate objective)}
    \right].
    % \|\epsilon - \epsilon_\theta(\sqrt{\bar{\alpha}_k}\hat{a} + \sqrt{1 - \bar{\alpha}_k}\epsilon, s, k)\|_2^2
\end{equation}

\subsubsection{\textbf{Q}-weighted \textbf{V}ariational \textbf{P}olicy \textbf{O}ptimization (\textbf{QVPO})}
QVPO~\cite{qvpoerror} is the pioneering work in $Q$-weighting methods, theoretically proving that when all state-action values $Q(s, a)$ are positive for any state-action pair $(s, a)$, the Q-weighted variational bound objective of the diffusion model~\cite{ddpm} provides a tight lower bound of the policy objective in online RL.
However, since $Q(s, a)$ is not always positive in general settings, QVPO proposes a weight transformation function utilizing the advantage term:
\begin{equation}
\begin{gathered}
    \mathcal{L}_\theta^\text{QVPO} = \underset{\substack{s \sim \mathcal{D} \\ a \sim \pi_\theta}}{\mathbb{E}}
    \left[
    \mathbf{w}(s, a) \| \epsilon  - \boldsymbol{\epsilon}_\theta(\sqrt{\bar{\alpha}_k}a + \sqrt{1 - \bar{\alpha}_k}\epsilon, s, k)\|^2
    \right],
    \\
    \mathbf{w}(s, a) = 
    \begin{cases}
        A(s, a), & A(s,a) \geq 0, \\
        0 & A(s,a) < 0,
    \end{cases}
\end{gathered}
\end{equation}
where advantage $A(s, a) = Q(s, a) - V(s)$ indicates how much better taking action $a$ in state $s$ is compared to the average value of that state.
Through this formulation, QVPO performs policy improvement by selectively learning only when advantages are positive or high, effectively filtering out suboptimal actions while preserving the diffusion model's capacity for multimodal action representation.

\subsubsection{\textbf{D}iffusion \textbf{P}olicy \textbf{M}irror \textbf{D}escent \& \textbf{S}oft \textbf{D}iffusion \textbf{A}ctor-\textbf{C}ritic (\textbf{DPMD \& SDAC})}
Similar to QVPO, Ma et al.~\cite{dpmdsdac} introduce reweighted score matching methods, which constitute a general loss family as follows:
\begin{equation}
    \mathcal{L}_{\theta}^\text{RSM} = \underset{\substack{s \sim \mathcal{D} \\ a \sim \pi_\theta}}{\mathbb{E}}
    \left[
    \mathbf{w}(s, a) \|\mathbf{s}_\theta(a^k, s, k) - \nabla_{a^k} \log p_k(a^k|s)\|^2
    \right],
\end{equation}
where $p_k(a^k|s)$ is the noise-perturbed policy with a Gaussian kernel.
Note that we can compute $p_k(a^k|s)$ in a tractable space (e.g., Gaussian space) through several mathematical foundations such as Tweedie's identity~\cite{tweedie}.

Specifically, they introduce two different $Q$-weighting approaches named DPMD and SDAC, based on mirror descent policy~\cite{md} and soft actor-critic~\cite{sac}, respectively.
Both algorithms employ the following exponential reweighting function:
\begin{equation}
    \mathbf{w}(s, a) = \exp \left(\frac{Q(s, a)}{\lambda}\right),
\end{equation}
where $\lambda$ is the regularization coefficient.
% This exponential weighting naturally arises from entropic regularization and allows for smooth differentiation between high-value and low-value actions, avoiding the hard thresholding employed by QVPO while maintaining stable gradient flow throughout the action space.

\subsection{Proximity-Based Methods}
\label{subsec:proximity}

Several proximity-based methods (e.g., TRPO~\cite{trpo}, PPO~\cite{ppo}) require direct computation of the log-likelihood of the policy $\pi(a|s)$ as follows:
\begin{equation}
\begin{gathered}
\label{eq:ppo}
    \mathcal{L}_\theta^{\text{PPO}} \triangleq \underset{(s, a) \sim \pi_{\text{old}}}{\mathbb{E}}
    \left[
    \min(r_\theta \hat{A}, \text{clip}(r_\theta, 1 - \delta, 1+\delta)\hat{A})
    \right], \\
    \text{where}~~r_\theta = \frac{\pi_\theta(a|s)}{\pi_{\text{old}}(a|s)},
\end{gathered}
\end{equation}
% \todo{A에 대한 설명 추가, e.g., GAE로 구한 advantage term}
where $\hat{A}$ is an estimator of the advantage function and $\delta$ is small constant, respectively.
These methods generally employ stochastic policy networks (e.g., Gaussian policies) to facilitate straightforward computation of action log-likelihoods. However, when combining diffusion policies with proximity-based methods, there is no direct way to compute the diffusion network's log-likelihood $\pi_\theta(a|s)$. The methods in this section address this challenge through various approaches.

\subsubsection{\textbf{Gen}erative \textbf{P}olicy \textbf{O}ptimization (\textbf{GenPO})}
GenPO~\cite{genpo} computes $\pi_\theta(a|s)$ through exact diffusion inversion~\cite{edict} combined with a doubled dummy action mechanism. Given an original action $a \in \mathbb{R}^d$, GenPO defines a dummy action $\tilde{a} = (x, y) \in \mathbb{R}^{2d}$ where the final action is computed as $a = \frac{x + y}{2}$. By alternately updating the two components of the dummy action, GenPO constructs an invertible flow process. This enables exact probability density computation via the change of variables theorem from normalizing flows~\cite{nice, realnvp}:
\begin{equation}
    \pi_\theta(a|s) = \mathcal{N}(\epsilon | 0, \mathbf{I}) \left| \det \frac{\partial f_\theta}{\partial \epsilon} \right|^{-1},
\end{equation}
where $f_\theta$ represents the entire forward diffusion process.
Note that the Jacobian determinant of $f_\theta$ is efficiently computed using normalizing flow techniques.
With tractable log-likelihoods, GenPO directly applies PPO's clipped surrogate objective in Eq.~\ref{eq:ppo}.
However, to learn GenPO's surrogate objective, all iterative gradients of the inversion process for computing $\pi_\theta(a|s)$ must be calculated, resulting in scalability issues (Accordingly, GenPO can be categorized as a BPTT-based method, as illustrated in Fig.~\ref{fig:taxonomy}).
Consequently, GenPO employs a short denoising timestep (i.e., NFE=5) to maintain computational feasibility.

\subsubsection{\textbf{F}low \textbf{P}olicy \textbf{O}ptimization (\textbf{FPO})}
The key innovation of FPO~\cite{fpo} is that it reformulates the policy optimization objective by replacing the intractable log-likelihood ratio $r_\theta$ with conditional flow matching losses $\mathcal{L}_\text{CFM}$. Specifically, FPO treats the flow matching objective as a surrogate for the log-likelihood in policy gradient methods. Given the conditional flow matching loss $\mathcal{L}_\text{CFM}$ and advantage $A(s, a)$, FPO optimizes:
\begin{equation}
\begin{gathered}
    r_\theta^{\text{FPO}} = \exp(\mathcal{L}_{\text{old}}^\text{CFM}(s, a) - \mathcal{L}_{\theta}^\text{CFM}(s, a)),\\
    \mathcal{L}_\text{CFM}(s, a) = \mathbb{E}_{k,\epsilon}\| \mathbf{v}_\theta(a_k, s, k) - (a - \epsilon)\|_2^2.
\end{gathered}
\end{equation}
This formulation enables FPO to be integrated seamlessly into PPO-style frameworks while maintaining stable on-policy updates.
By optimizing the Evidence Lower Bound (ELBO) as a proxy for log-likelihood, FPO circumvents expensive likelihood computations while preserving the expressive power of flow-based policies for multimodal action distributions.

\subsection{BackPropagation Through Time (BPTT)-based Methods}
\label{subsec:bptt}

These methods do not rely on the surrogate objectives typically used in diffusion models, but instead train through the entire diffusion process in a manner similar to BPTT of recurrent neural network training.
To mitigate the high computational cost of BPTT, they employ strategies such as using a small number of timesteps~\cite{dime, dacer, dacerv2}, or leveraging approximations through few-step generation techniques like consistency models~\cite{cpql}.
However, these approaches face potential scalability issues: when extending the number of sampling steps to improve performance, they either incur increased computational costs or suffer from degraded action quality due to approximation errors.

\subsubsection{\textbf{D}iffusion \textbf{A}ctor-\textbf{C}ritic with \textbf{E}ntropy \textbf{R}egulator (\textbf{DACER})}
\label{subsubsec:dacer}

DACER directly optimizes the diffusion policy by maximizing the expected Q-value through end-to-end backpropagation through the entire reverse diffusion chain as follows:
\begin{equation}
    % \max_\theta \mathbb{E}_{s \sim \mathcal{D}}[Q_\phi^\pi(s, a_0)]
    \mathcal{L}_\theta^\text{DACER} = \underset{\substack{s \sim \mathcal{D}\\a \sim \pi_\theta}}{\mathbb{E}}[-Q_\phi^\pi(s, a)].
\end{equation}

Similar to MaxEnt RL frameworks~\cite{sql, sac}, DACER introduces a mechanism to regulate the exploitation-exploration trade-off by incorporating an estimated entropy $\hat{\mathcal{H}} = \mathbb{E}_{a \sim \pi}[-\beta \log \pi(a|s)]$.
To circumvent the intractability of calculating entropy directly from diffusion models, the authors leverage a tractable Gaussian Mixture Model (GMM) as a proxy.
This approximated entropy $\hat{\mathcal{H}}$ is then used to update the exploration parameter $\alpha$ according to the following formulation:
% Similar to Maximum Entropy Reinforcement Learning (MaxEnt RL)~\cite{sql, sac}, DACER 는 estimated entropy $\hat{\mathcal{H}} = \underset{a \sim \pi}{\mathbb{E}}[-\beta \log \pi(a|s)]$ 를 regulator로 이용하여, training에서의 exploitation-exploration balance를 조정하는 방법론을 제안한다.
% 하지만 diffusion policy의 entropy는 intractable이기 때문에 이 방법론은 tractable Gaussian mixture model (GMM)을 활용하여 근사된 entropy $\hat{\mathcal{H}}$를 계산한 후, exploration parameter $\alpha$ 를 다음과 같이 update한다:
\begin{equation}
    \alpha \leftarrow \alpha - \lambda_\mathcal{H}(\hat{\mathcal{H}} - \bar{\mathcal{H}}),
\end{equation}
where $\bar{\mathcal{H}}$ is target entropy and $\lambda_\alpha$ is the hyperparameter.
Lastly, the parameter $\alpha$ is then used to modulate the variance of Gaussian noise added to the diffusion policy output: $a \leftarrow a + \lambda\alpha \cdot \mathcal{N}(\mathbf{0}, \mathbf{I})$.

\subsubsection{Enhanced DACER (\textbf{DACERv2})}
% DACERv2~\cite{dacerv2}는 DACER의 BPTT manner로 인한 확장성 문제를 해결하기 위해서 QSM~\cite{qsm} 와 비슷한 방식의 action gradient guidance를 auxiliary objective로 활용한 방법론이다 (so it can be regarded as action-gradient families).
% 이들은 action-gradient $\nabla_a Q(s, a)$를 Q-gradient field $\nabla_{a^k} Q(s, a^k)$로 확장하여 모든 diffusion step k에 대해서 action gradient를 활용할 수 있게 함.
% 이렇게 제안된 Q-gradient field를 이용해 QSM~\cite{qsm}와 비슷한 방식으로 time dependent한 score model과 더욱 정교한 score matching을 다음과 같이 수행함으로써 성능 증가를 도모한다:
DACERv2~\cite{dacerv2} is specifically designed to overcome the scalability bottlenecks associated with the BPTT-based approach of DACER, aiming to maintain high performance even with a significantly reduced number of diffusion steps.
To achieve this, it incorporates action gradient guidance---similar to the QSM~\cite{qsm} framework---as an auxiliary objective, thereby placing DACERv2 within the action-gradient family of algorithms.
% DACERv2~\cite{dacerv2} is a methodology designed to address the scalability limitations inherent in DACER's BPTT-based approach by utilizing action gradient guidance---similar to the framework of QSM~\cite{qsm}---as an auxiliary objective; as such, it can be categorized within the action-gradient family of algorithms.
Specifically, they extend the standard action gradient $\nabla_a Q(s, a)$ into a Q-gradient field $\nabla_{a^k} Q(s, a^k)$, enabling the utilization of action gradients across all diffusion steps $k$.
By leveraging this proposed Q-gradient field, the method facilitates a more sophisticated score-matching process with a time-dependent score model, analogous to QSM~\cite{qsm}, thereby enhancing overall performance as follows:
% \begin{equation}
% \begin{gathered}
\begin{align}
\begin{split}
    \mathcal{L}_\theta^\text{DACERv2} &= \mathcal{L}_\theta^\text{DACER} \\
    &+ \eta \underset{\substack{s \sim \mathcal{D}}}{\mathbb{E}}
    \left[
    \|\mathbf{s}_\theta(a^k, s, k) - w(k) \nabla_{a^k} Q(s, a^k)\|_2^2
    \right],
\end{split}
\end{align}
% \end{gathered}
% \end{equation}
where $\eta$ is a hyperparameter, and $w(k)$ denotes an exponential decay function for precise control over the denoising process.
Note that DACERv2 normalizes the action gradient field (i.e., $\nicefrac{\nabla_{a^k} Q(s, a^k)}{(\|\nabla_{a^k} Q(s, a^k)\| + \delta})$, where $\delta > 0$ is a small constant) to improve the stability of the training process.
While DACERv2 demonstrates superior multimodal capabilities compared to its predecessor, it still faces scalability challenges due to its reliance on the $\mathcal{L}_\theta^\text{DACER}$, which necessitates BPTT for policy improvement.

\subsubsection{\textbf{D}iffusion-\textbf{B}ased \textbf{M}aximum \textbf{E}ntropy \textbf{RL} (\textbf{DIME})}
% DIME~\cite{} 은 Entropy term을 regulartor로 이용하던 DACER, DACERv2와 다르게 좀더 직접적으로 Maximum Entropy Reinforcement Learning을 이용해서 diffusion policy를 학습할 수 있는 방법을 제안한다.
% 이방법론은 diffusion model의 intractable entropy를 직접 계산하는대신 근사 추론 기법을 활용하여 MaxEnt 목적함수에 대한 tractable lower bound를 유도함으로써, diffusion model을 학습가능하도록 한다.
% policy improvement를 위해서 DIME은 다음과 같은 approximate inference problem을 푼다:
Unlike DACER and DACERv2, which employ entropy terms primarily as regularizers, DIME~\cite{dime} proposes a more direct framework for training diffusion policies within the MaxEnt RL paradigm~\cite{sql}.
To overcome the intractability of calculating entropy in diffusion models, this method derives a tractable lower bound for the MaxEnt objective by leveraging approximate inference techniques~\cite{berner2022optimal}.
For policy improvement, DIME formulates and solves the following approximate inference problem:
\begin{equation}
    \mathcal{L}_\theta^\text{DIME} = D_\text{KL}(\pi_\theta^{0:K}(a^{0:K} | s)~|~\bar{\pi}^{0:K}(a^{0:K} | s)),
\end{equation}
where $D_\text{KL}(\cdot | \cdot)$ is the KL divergence metric, and $\pi_\theta^{0:K}(a^{0:K} | s)$ denotes the joint probability distribution of the parameterized reverse diffusion trajectories.
Here, $\bar{\pi}^{0:K}(a^{0:K} | s)$ represents the target joint distribution defined by the forward diffusion process, where the marginal distribution at the clean action step $a^0$ follows the energy-based formulation $\bar{\pi}^0(a^0|s) \propto \exp(Q(s, a^0))$.
% 전체 reverse process와 forward process의 trajectory distribution을 align해주는 과정을 통해 정책이 Critic (Q-function)이 선호하는 행동 분포를 학습하도록 유도할 수 있다.
By aligning the entire reverse trajectory distribution with the forward target, the policy is effectively guided to capture the action distributions prioritized by the Critic (Q-function).

\subsubsection{\textbf{C}onsistency \textbf{P}olicy with \textbf{Q}-\textbf{L}earning (\textbf{CPQL})}
CPQL~\cite{cpql} addresses the dual challenges of inference latency and inaccurate guidance in diffusion-based RL by leveraging the consistency model~\cite{cm} for one-step action generation.
CPQL formulates the policy as a parameterized consistency function $f_\theta(\cdot)$ that directly maps noise to actions in a single step as follows:
\begin{equation}
    \pi_\theta(a|s) \triangleq f_\theta(a^k, s, k)= c_{skip}(k)a^k + c_{out}(k)F_\theta(a^k, s, k),
\end{equation}
where $F_\theta(\cdot)$ is the trainable network, and $c_{skip}(k)$ and $c_{out}(k)$ are the coefficient functions.
The consistency policy establishes a direct mapping from any point on the Probability Flow ODE trajectory to the target action, effectively eliminating the need for iterative denoising.
To ensure stable training, CPQL facilitates policy improvement by combining a simplified reconstruction loss $\mathcal{L}_{\theta}^\text{RC}$—derived from the consistency training objective—with Q-value maximization as follows:
% where $F_\theta(\cdot)$ is the trainable network, $c_{skip}(k)$ and $c_{out}(k)$ are the coefficient functions.
% The consistency policy establishes a direct mapping from any point on the probability flow ODE trajectory to the desired action, eliminating the need for iterative denoising.
% CPQL은 stable training을 위해 consistency model training loss를 simplify한 reconstruction loss $\mathcal{L}_{\theta}^\text{RC}$와 Q-value maximization을 조합하여 다음과 같이 policy improvement를 진행한다:
\begin{equation}
\begin{gathered}
    % \mathcal{L}_\theta^\text{CPQL} = \alpha\mathcal{L}_\theta^\text{RC} - \frac{\eta}{\mathbb{E}_{(s,a) \sim \mathcal{D}}[Q(s,a)]}\mathbb{E}_{s \sim \mathcal{D}}[Q_\phi^\pi(s, \hat{a})]\\
    \mathcal{L}_\theta^\text{CPQL} = \alpha\mathcal{L}_\theta^\text{RC} - \frac{\eta}{\underset{(s,a) \sim \mathcal{D}}{\mathbb{E}}[Q(s, a)]}\underset{\substack{s \sim \mathcal{D}\\ a \sim \pi_\theta}}{\mathbb{E}}[Q_\phi^\pi(s, a)],\\
    \text{where}~~\mathcal{L}_{\theta}^\text{RC} = \mathbb{E}[d(f_\theta(a^k, s,  k), a)],
\end{gathered}
\end{equation}
% where $\alpha$ and $\eta$ are the trainable balancing parameters, and $d(\cdot, \cdot)$ is the distance metric, respectively.
% CPQL's one-step generation paradigm, which leverages the consistency model, yields dramatic efficiency improvements.
where $\alpha$ and $\eta$ are balancing parameters, and $d(\cdot, \cdot)$ denotes a distance metric.
By adopting this one-step generation paradigm, CPQL achieves dramatic improvements in inference efficiency while maintaining the expressive advantages of diffusion-based modeling.

\subsubsection{\textbf{D}istributional \textbf{S}oft \textbf{A}ctor-\textbf{C}ritic with \textbf{D}iffusion Policy (\textbf{DSAC-D})}
% 이전의 방법들이 standard RL이나 MaxEnt RL에 기반한 방법을 제안한 것과 다르게, DSAC-D~\cite{dsacd}는 처음으로 $Q(s, a)$ 대신 probability distributions of returns $Z(s, a)$를 modeling하는 Distributional RL~\cite{c51, iql}에 diffusion network를 적용한 첫 논문이다.
% Distributional RL에서의 $Z(s, a)$를 더욱 정교하게 근사하기 위하여, DSAC-D는 diffusion policy 뿐만 아니라 multi-peaked value distribution estimation이 가능한 Diffusion Value Network (DVN)를 제안한다.
Departing from prior works rooted in standard or MaxEntRL, DSAC-D~\cite{dsacd} represents the first attempt to integrate diffusion networks with Distributional RL~\cite{c51, dsac, dsact}, which models the probability distribution of returns $Z(s, a)$ rather than a scalar $Q(s, a)$.
To achieve a more precise approximation of the return distribution in this framework, DSAC-D introduces the Diffusion Value Network (DVN), which can model multi-peaked value distributions.
% Unlike conventional critics, DVN enables the estimation of complex, multi-peaked value distributions, thereby synergizing the expressive power of diffusion models with the robustness of distributional reinforcement learning.
Similar to Eq.~\ref{eq:diffusion_policy_sampling}, the DVN generates return samples through the reverse diffusion process:
\begin{equation}
    z^{k-1} = \frac{1}{\sqrt{\alpha_k}}\left(z^k - \frac{1 - \alpha_k}{\sqrt{1-\bar{\alpha}_k}}\boldsymbol{\epsilon}_\theta(a^k, s, k)\right) + \sqrt{1- \alpha_k}\epsilon,
\end{equation}
where $z^K \sim \mathcal{N}(0, \mathbf{I})$ and $z^0$ represents the generated return sample.
This approach enables accurate characterization of multi-peaked return distributions, significantly suppressing value estimation bias.

For policy improvement, DSAC-D incorporates the entropy term $\beta \log \pi(a|s)$ from MaxEnt RL into a revised distributional Bellman equation~\cite{c51} to facilitate exploration. 
% DSAC-D는 diffusion policy의 exploration을 위해 MaxEnt RL의 entropy term $\beta \log \pi(a|s)$ 를 additionally incorporate한 revised distributional Bellman equation을 활용한다.
The loss function for the diffusion policy with distributional return under the policy $Z^\pi(s, a)$ is formulated as follows:
% DSAC-D의 loss term of diffusion policy as follows:
\begin{equation}
    \mathcal{L}_\theta^\text{DSAC-D} = \underset{\substack{s \sim \mathcal{D}\\a \sim \pi_\theta}}{\mathbb{E}}\left[-\mathbb{E}[Z^\pi(s,a)] + \beta \log \pi(a|s)\right].
\end{equation}
Note that to estimate the entropy term $\log \pi(a|s)$, which balances exploitation and exploration, DSAC-D adopts the entropy estimation technique from DACER~\cite{dacer}, utilizing Gaussian Mixture Models (GMMs) to approximate policy entropy.

\subsection{Other Related Diffusion-Based RL Paradigms}
\label{subsec:related}

Our primary focus is on online model-free methods that utilize diffusion policy representations.
However, the landscape of diffusion models in reinforcement learning and control is broad, encompassing several adjacent yet distinct paradigms that warrant comparison to clearly situate our work.
% \paragraph{Offline-to-Online Diffusion Policy Fine-Tuning} 

A closely aligned and frequently explored research direction is the Offline-to-Online Fine-tuning RL approach~\cite{dppo, parl, policydecorator, resfit, resip}.
These methods typically leverage the expressive power of a diffusion policy initialized via imitation learning on a large demonstration dataset, and then subject it to policy optimization in the online RL setting.
Similar to the aforementioned model-free DPRL paradigms, these approaches either propose specialized online RL frameworks tailored for diffusion policies~\cite{dppo} or utilize model-agnostic training strategies—such as decoupled policy improvement—to fine-tune large-scale pre-trained models like Vision-Language-Action (VLA) and high-capacity diffusion policies~\cite{parl}.
% 이들은 주로 이전에 소개했던 model-free DPRL방법론과 비슷하게, diffusion policy에 맞는 online RL 방법론을 제안하거나~\cite{dppo}, DIPO와 같이 policy improvement를 분리한 model-agnostic training방법론을 이용하여 VLA, diffusion policy와 같은 pretrained model을 학습한 방법론들이 있다~\cite{parl}.
Furthermore, recognizing the non-trivial nature of applying full RL updates to the complex diffusion network, specialized methods utilize Residual Policy Learning~\cite{rpl}.
These techniques freeze the bulk of the pre-trained base diffusion policy and only learn a residual action term on top of it~\cite{policydecorator, resfit, resip}.

Another branch involves Model-Based Reinforcement Learning (MBRL), where diffusion models are used to learn a world model or system dynamics rather than the policy itself.
These approaches focus on training a generative model to directly estimate future states, rewards, and other dynamics given a state-action pair~\cite{dwm, world4rl} to achieve ``online-like" learning through synthetic rollouts.
Other works use diffusion models for synthesizing limited modalities~\cite{pgd, polygrad}, such as motion planning trajectories~\cite{mpd, diffuser}.
While these MBRL methods achieve ``online-like" learning through synthetic rollouts, their reliance on accurately modeling the environment's transition function fundamentally distinguishes them from the model-free policy optimization focus of this paper.

% ----------------

\subsection{Cross-Cutting Challenges and Optimization Strategies in Online DPRL}
\label{subsec:crosscutting}

Beyond the core policy improvement mechanisms, the integration of diffusion models with online RL introduces several structural and computational hurdles.
This section discusses how existing literature addresses the intractability of analytical expressions, exploration efficiency, and value estimation stability.

% \subsubsection{Addressing the Intractability of Log-Likelihood and Entropy}
% The absence of an analytical form for stochastic policy $\log \pi_\theta(a|s)$ remains a primary barrier to applying standard and MaxEnt RL to diffusion policies.
% To circumvent this, GenPO and FPO employ latent mapping or lower-bound approximations for importance sampling, while DACER and DSAC-D utilize Gaussian Mixture Models (GMM) to estimate intractable entropy.
% Alternatively, SDAC adopts a noise-based heuristic surrogate, whereas DIME derives a theoretically rigorous MaxEnt lower bound tailored for diffusion objectives.
% These diverse strategies enable effective exploration-exploitation balancing despite the structural complexity of diffusion-based action distributions.

\subsubsection{Addressing the Intractability of Log-Likelihood and Entropy}
The most significant barrier in applying standard RL to diffusion policies is the absence of an analytical form for the stochastic policy $\log \pi_\theta(a|s)$.
As previously discussed in the context of proximity-based methods, this intractability complicates the calculation of importance sampling ratios---a challenge tackled by GenPO through latent space mapping and FPO via approximate lower bounds.
This same limitation extends to Maximum Entropy (MaxEnt) RL, where the entropy term $\mathcal{H}(\pi(\cdot|s)) = \mathbb{E}_{a \sim \pi_\theta}[-\log \pi(a|s)]$ must be estimated to balance exploitation and exploration.

To circumvent this, various approximation strategies have been proposed.
DACER and DACERv2 utilize Gaussian Mixture Models (GMM) to fit the action distribution and approximate entropy, while SDAC adopts a more heuristic approach by using the log-probability of the additive exploration noise as a surrogate for the policy's log-likelihood.
In contrast, DIME provides a rigorous theoretical treatment by deriving a maximum-entropy lower bound specifically tailored for diffusion-based objectives, ensuring that the exploration incentive remains consistent with the diffusion training process.

\subsubsection{Enhancing Exploitation via Batch Action Sampling}
Due to the high stochasticity of the diffusion process, a single sampled action may not reliably represent the policy's optimal intent, leading to inefficient exploitation.
To mitigate this, many Online DPRL frameworks adopt Batch Action Sampling, a technique popularized by Diffusion-QL~\cite{diffusionql} in the offline setting.
Methods such as QVPO, SDAC, and DPMD sample multiple action candidates for a given state and utilize them to reduce variance in gradient estimation or to select the most promising action based on the critic's evaluation, thereby stabilizing the online learning signal.

\subsubsection{Mitigating Value Overestimation Bias}
As with conventional DRL, Online DPRL is susceptible to value overestimation bias, which can be exacerbated by the high-dimensional and multimodal nature of diffusion action spaces.
To ensure stable value landscapes, many algorithms (e.g., DIPO, QVPO, CPQL, and DACERv2) incorporate Double Q-learning architectures~\cite{doubleql}.
Furthermore, some recent approaches seek to capture the full aleatoric uncertainty of the environment by integrating Distributional Value Approximators.
For instance, DSAC-D and DACERv2 leverage distributional RL principles to provide a more robust critic, which in turn leads to more reliable policy updates for the diffusion actor.
\begin{table*}[t]
\centering
\caption{Summary of the 12 Isaac Lab environments utilized in the empirical analysis. Images are sourced from the official Isaac Lab documentation (available at \url{https://isaac-sim.github.io/IsaacLab/main/source/overview/environments.html}).}
\label{tab:isaac_lab_tasks_integrated}
\resizebox{0.95\linewidth}{!}{
\renewcommand{\arraystretch}{1.6}
\scriptsize
\begin{tabular}{l c m{4.2cm} | l c m{4.2cm}}
\toprule
 & \textbf{Task} & \textbf{Description} & & \textbf{Task} & \textbf{Description} \\
\midrule

% --- Row 1 ---
& \makecell{\includegraphics[width=2.0cm]{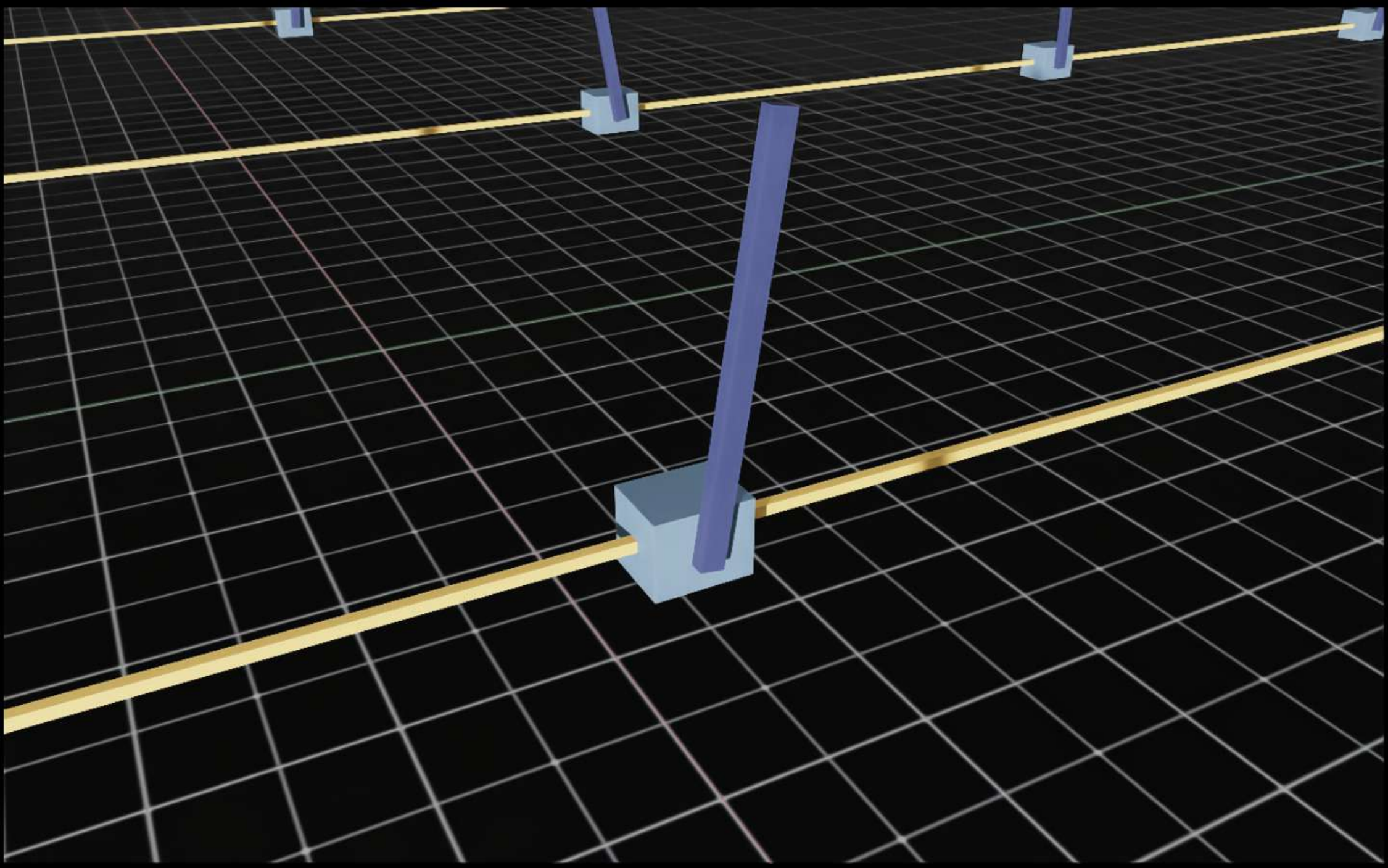} \\ \textbf{Cartpole}} & Move the cart to keep the pole upwards in the classic cartpole control. The state and action spaces defined as $s \in \mathbb{R}^{4}, a \in \mathbb{R}^{1}$ & 
& \makecell{\includegraphics[width=2.0cm]{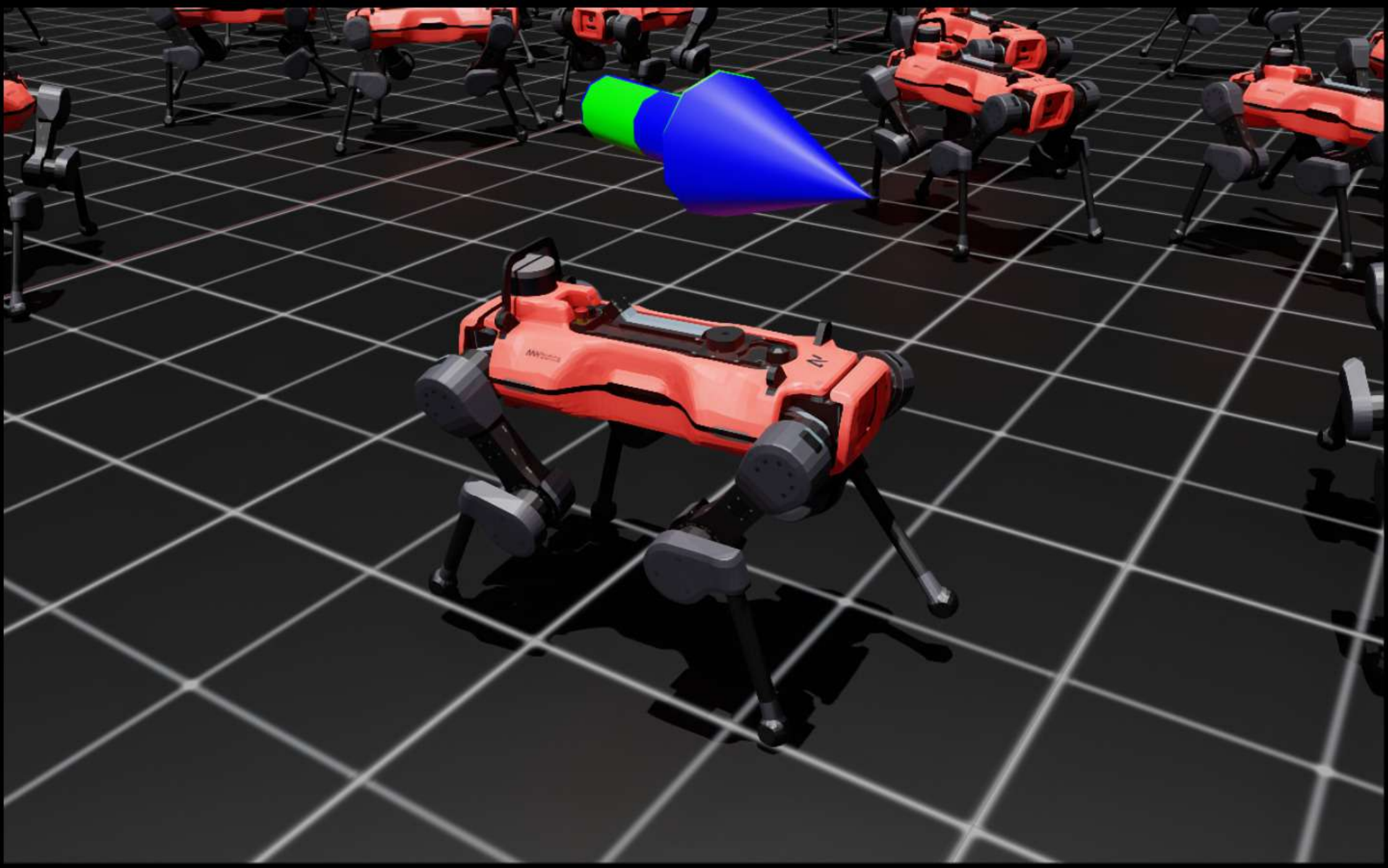} \\ \textbf{ANYmal-D}} & 	
Track a velocity command on flat terrain with the quadrupedal ANYbotics Anymal-D robot. The state and action spaces defined as $s \in \mathbb{R}^{48}, a \in \mathbb{R}^{12}$ \\
\cmidrule{2-3} \cmidrule{5-6}

% --- Row 2 ---
\rotatebox[origin=c]{90}{\textbf{Classic}} 
& \makecell{\includegraphics[width=2.0cm]{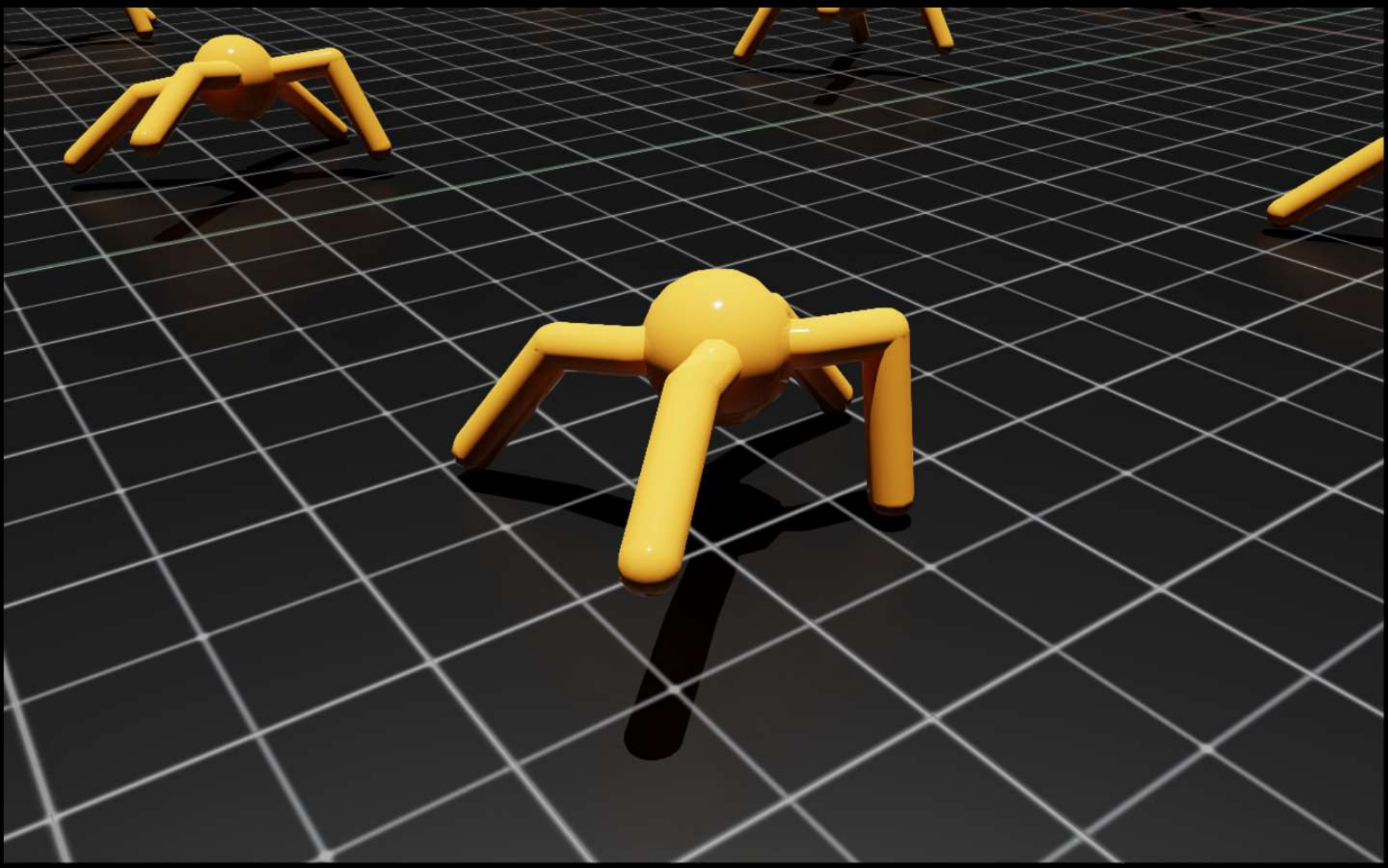} \\ \textbf{Ant}} & Move towards a direction with the MuJoCo ant robot. The state and action spaces defined as $s \in \mathbb{R}^{60}, a \in \mathbb{R}^{8}$ & 
& \makecell{\includegraphics[width=2.0cm]{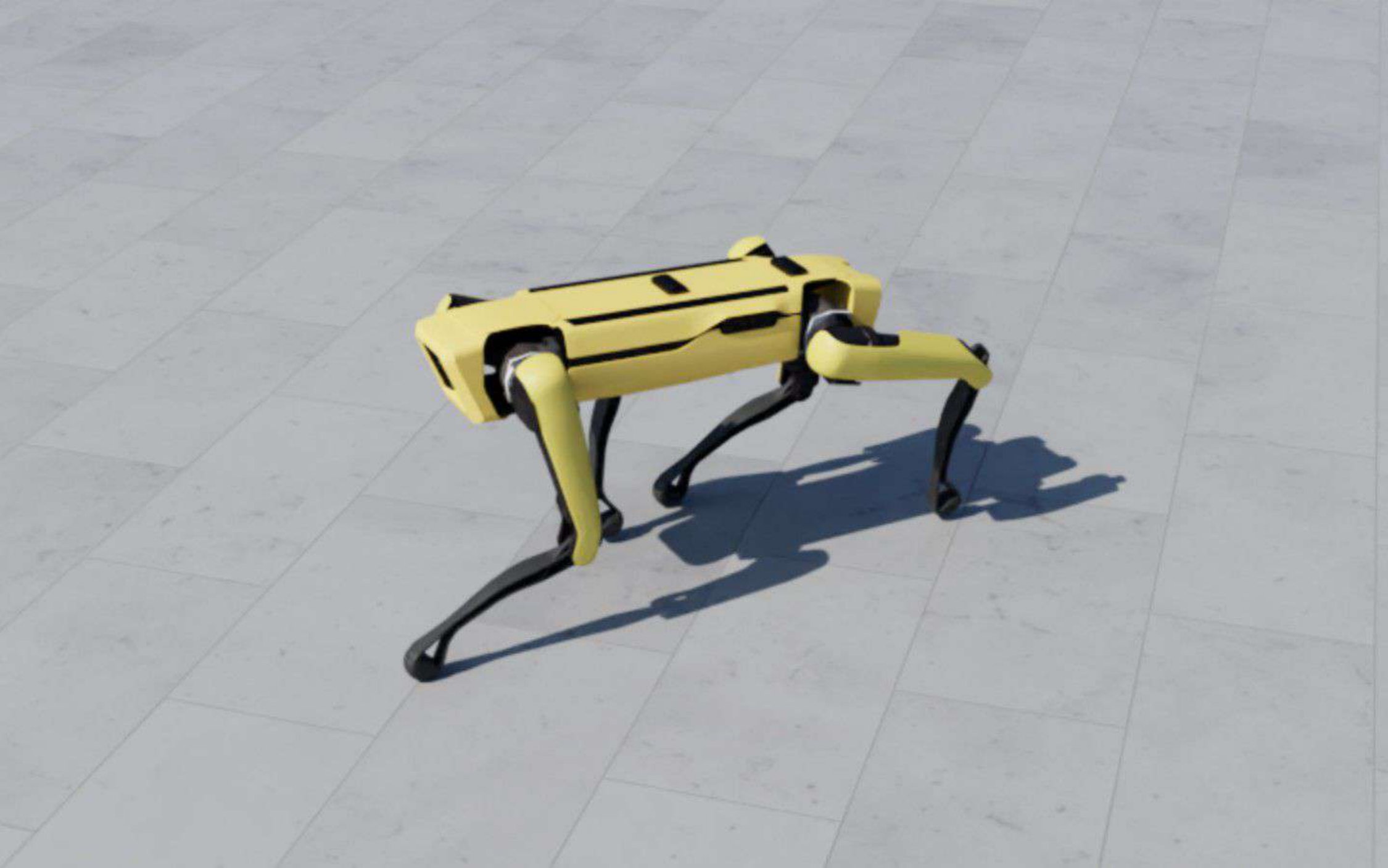} \\ \textbf{B.D. Spot}} & Track a velocity command on flat terrain with the quadrupedal Boston Dynamics Spot robot. The state and action spaces defined as $s \in \mathbb{R}^{48}, a \in \mathbb{R}^{12}$ \\
\cmidrule{2-3} \cmidrule{5-6}

% --- Row 3 ---
& \makecell{\includegraphics[width=2.0cm]{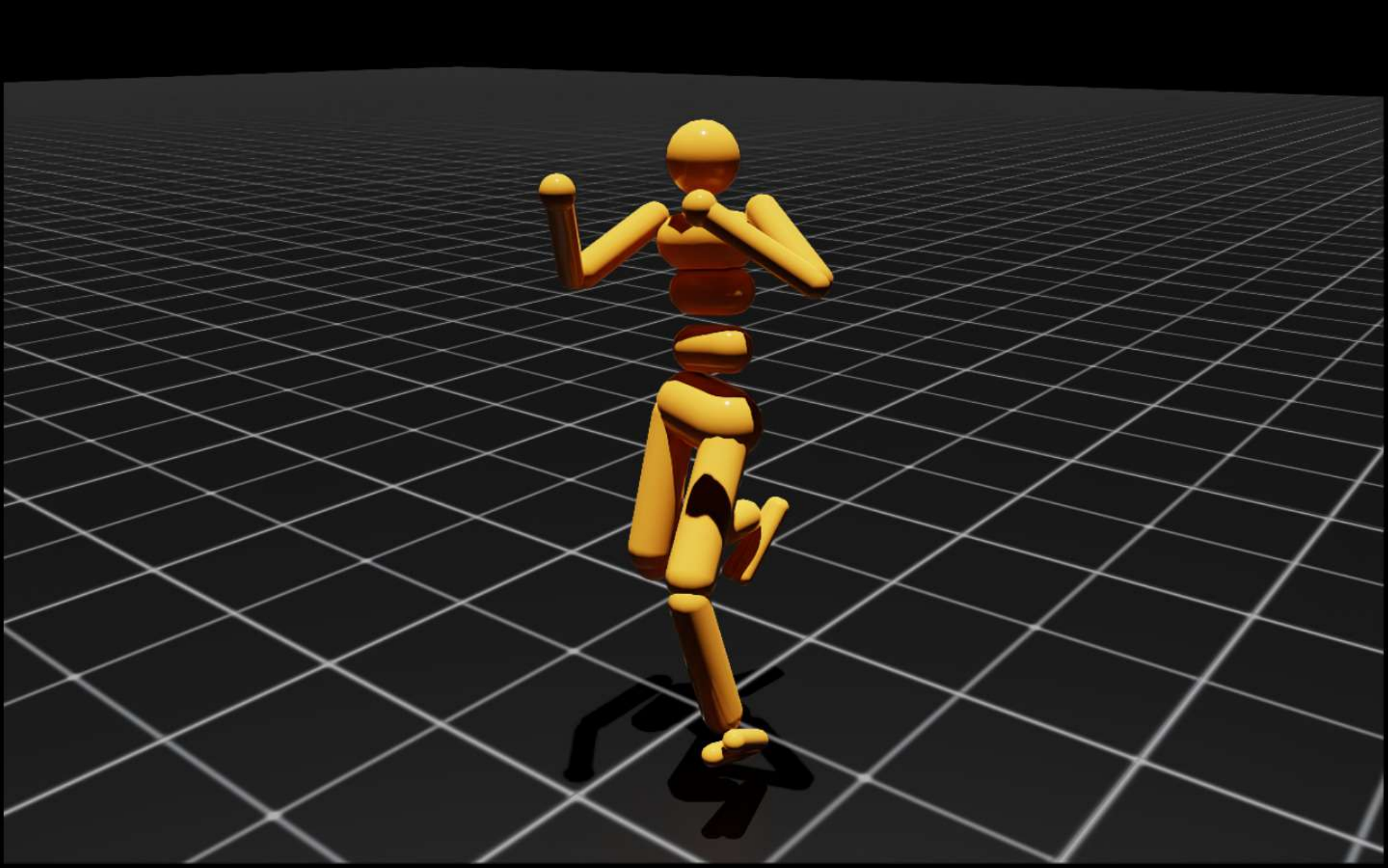} \\ \textbf{Humanoid}} & Move towards a direction with the MuJoCo humanoid robot. The state and action spaces defined as $s \in \mathbb{R}^{87}, a \in \mathbb{R}^{21}$ & 
\multirow{2}{*}{\rotatebox[origin=c]{90}{\textbf{Locomotion~~~~~~}}} 
& \makecell{\includegraphics[width=2.0cm]{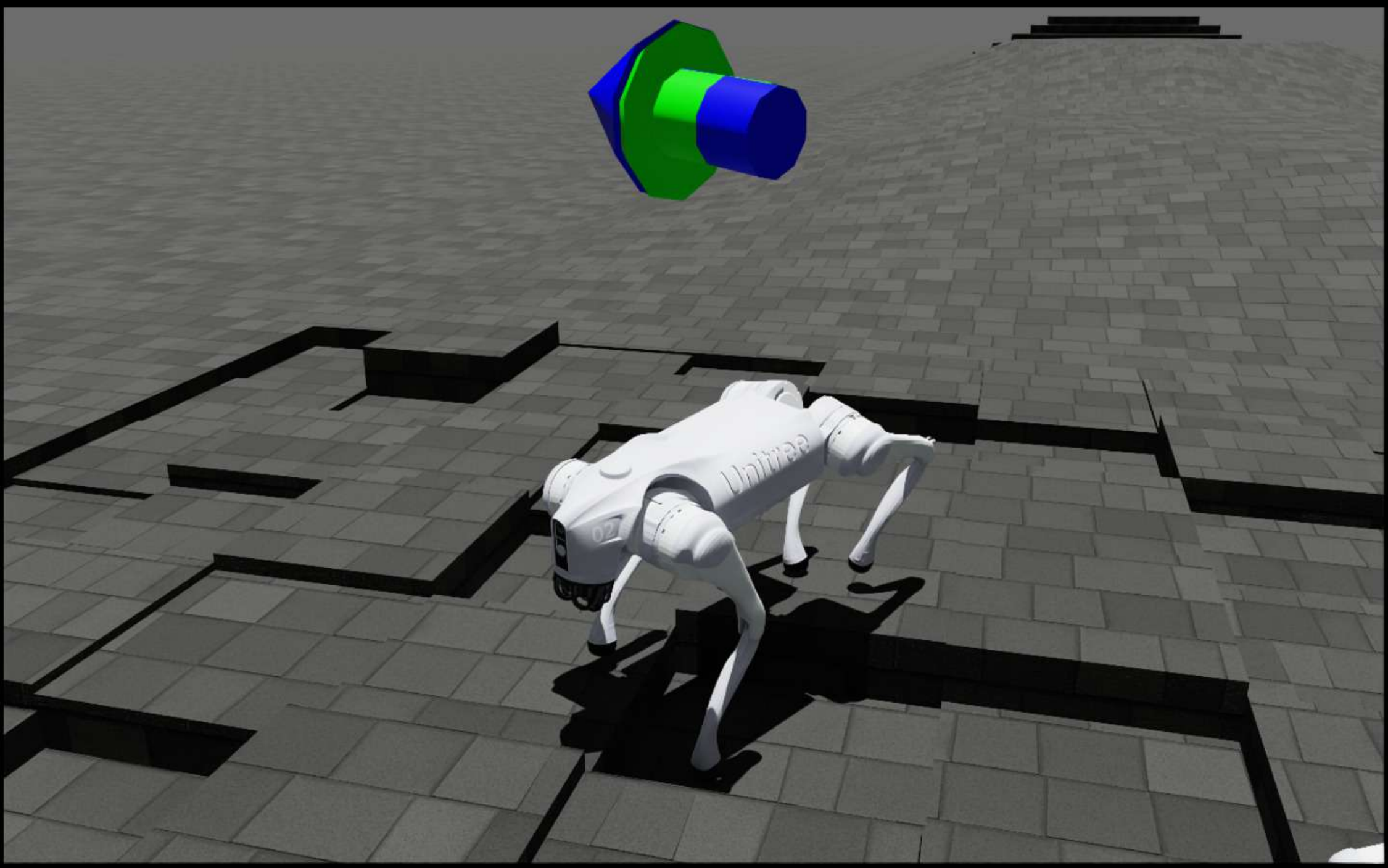} \\ \textbf{Unitree Go2}} & Track a velocity command on rough terrain with the quadrupedal Unitree Go2 via LiDAR exteroception. The state and action spaces defined as $s \in \mathbb{R}^{235}, a \in \mathbb{R}^{12}$ \\ \cmidrule{1-3} \cmidrule{5-6}

% --- Row 4 ---
% \rotatebox[origin=c]{90}{\textbf{Manipulation}}
\multirow{3}{*}{\rotatebox[origin=c]{90}{\textbf{Manipulation~~~~~~~~~~~~~~~}}} 
& \makecell{\includegraphics[width=2.0cm]{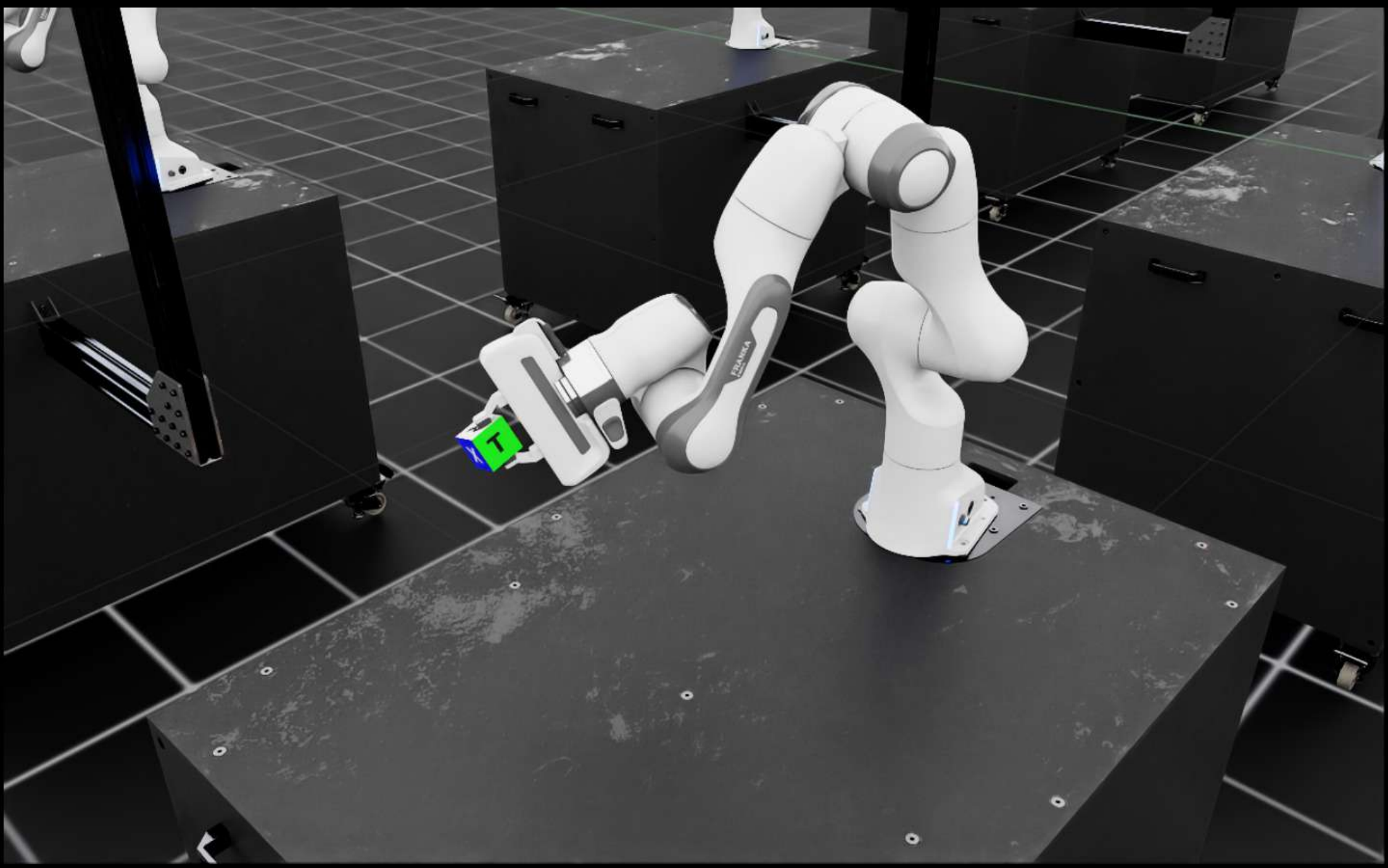} \\ \textbf{Franka Lift}} & Pick a cube and bring it to a sampled target position with the Franka robot. The state and action spaces defined as $s \in \mathbb{R}^{36}, a \in \mathbb{R}^{8}$ & 
& \makecell{\includegraphics[width=2.0cm]{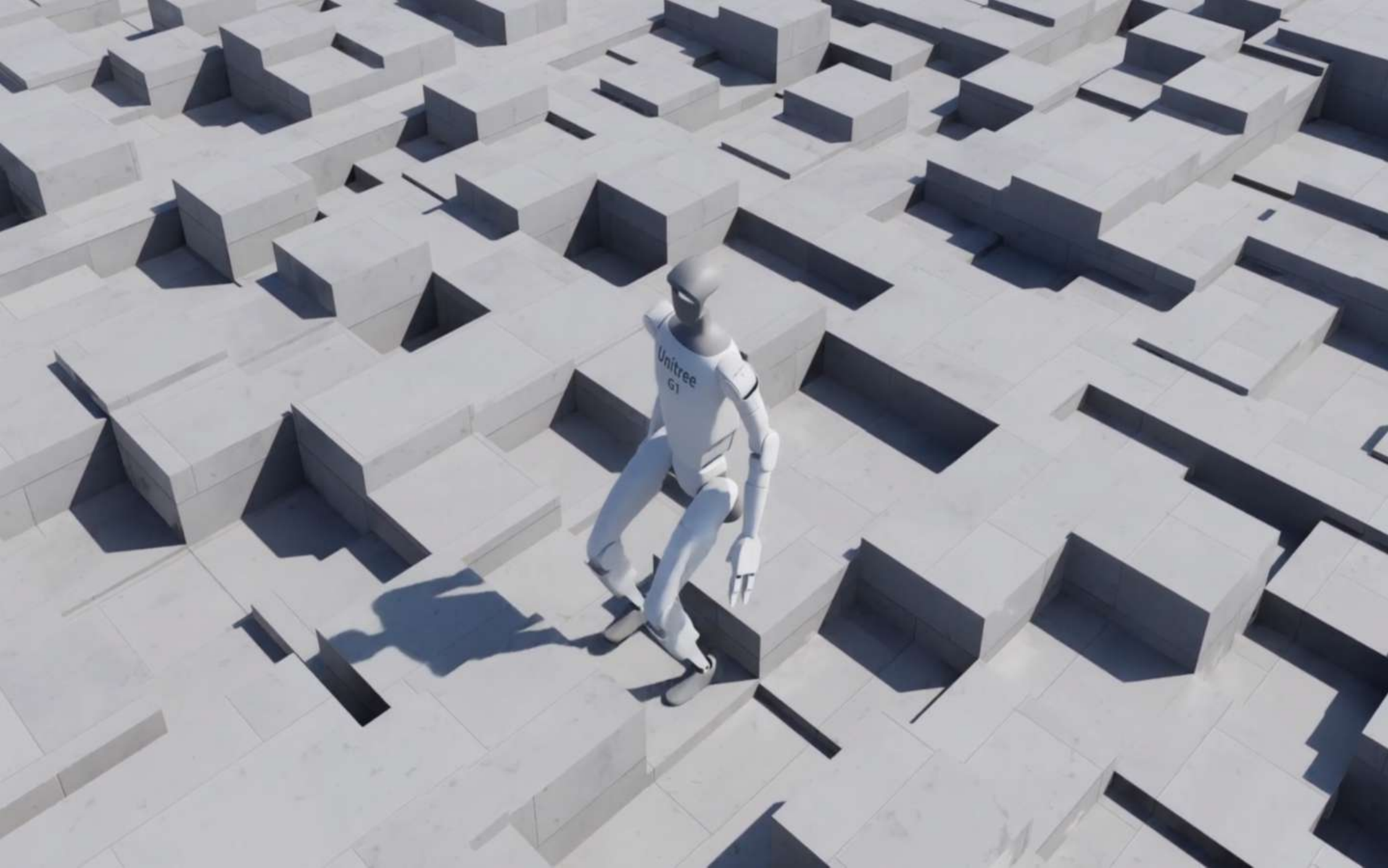} \\ \textbf{Unitree G1}} & Track a velocity command on rough terrain with the humanoid Unitree G1 via LiDAR exteroception. The state and action spaces defined as $s \in \mathbb{R}^{310}, a \in \mathbb{R}^{37}$ \\
\cmidrule{2-3} \cmidrule{5-6}

% --- Row 5 ---
& \makecell{\includegraphics[width=2.0cm]{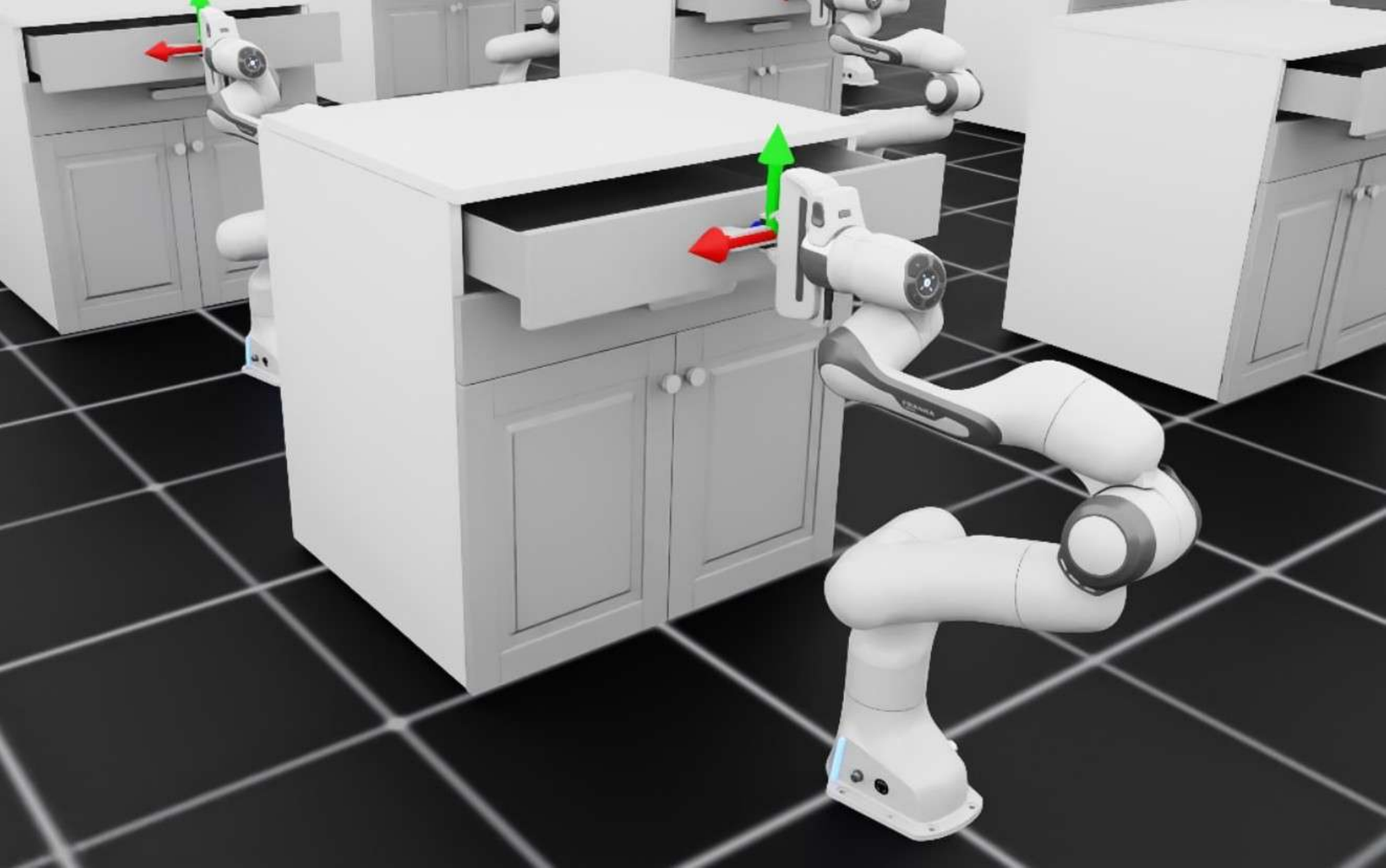} \\ \textbf{Franka Open Drawer}} & Grasp the handle of a cabinet’s drawer and open it with the Franka robot. The state and action spaces defined as $s \in \mathbb{R}^{31}, a \in \mathbb{R}^{8}$ & 
& \makecell{\includegraphics[width=2.0cm]{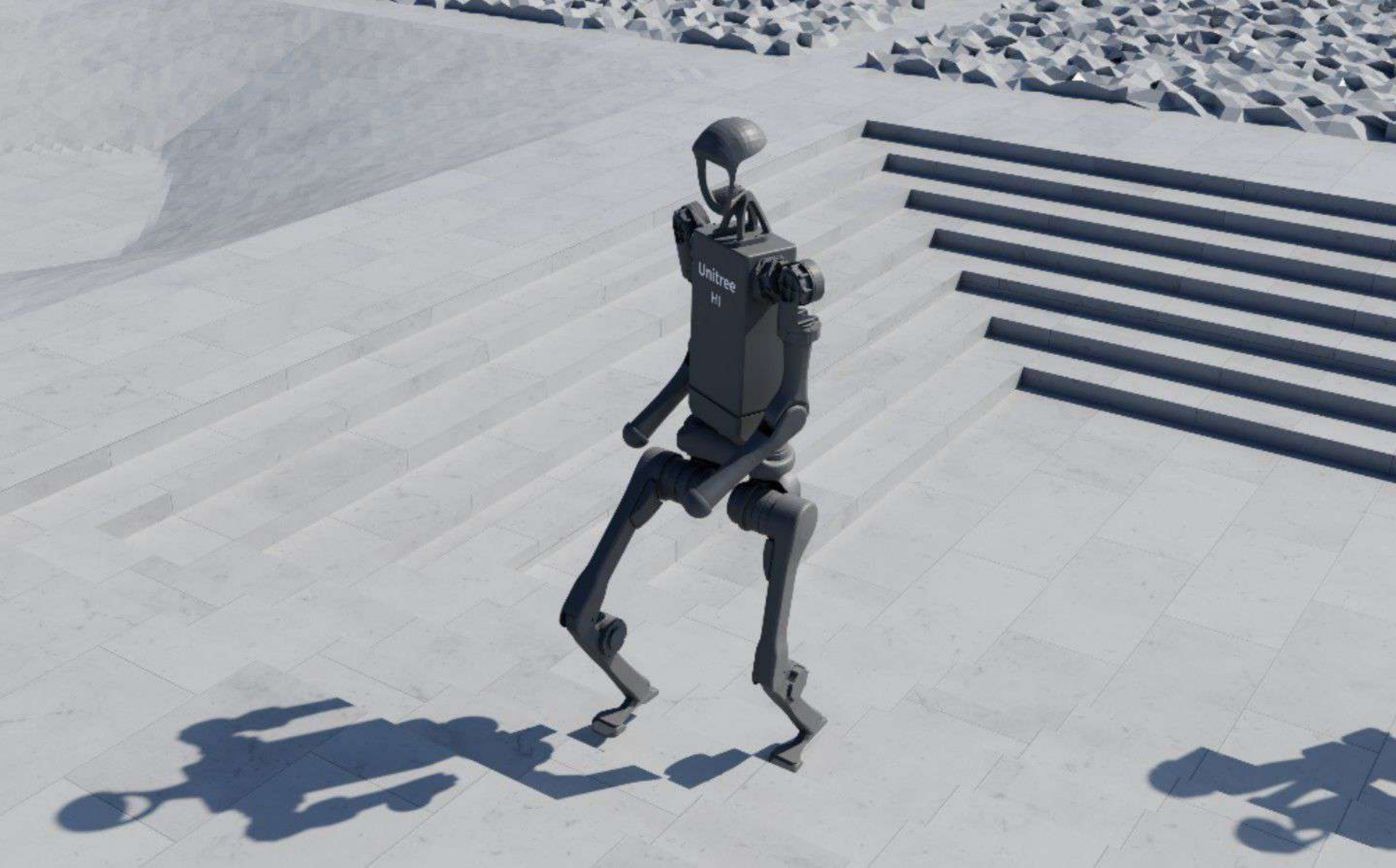} \\ \textbf{Unitree H1}} & Track a velocity command on rough terrain with the Humanoid Unitree H1 via LiDAR exteroception. The state and action spaces defined as $s \in \mathbb{R}^{256}, a \in \mathbb{R}^{19}$ \\
\cmidrule{2-3} \cmidrule{5-6}

% --- Row 6 ---
& \makecell{\includegraphics[width=2.0cm]{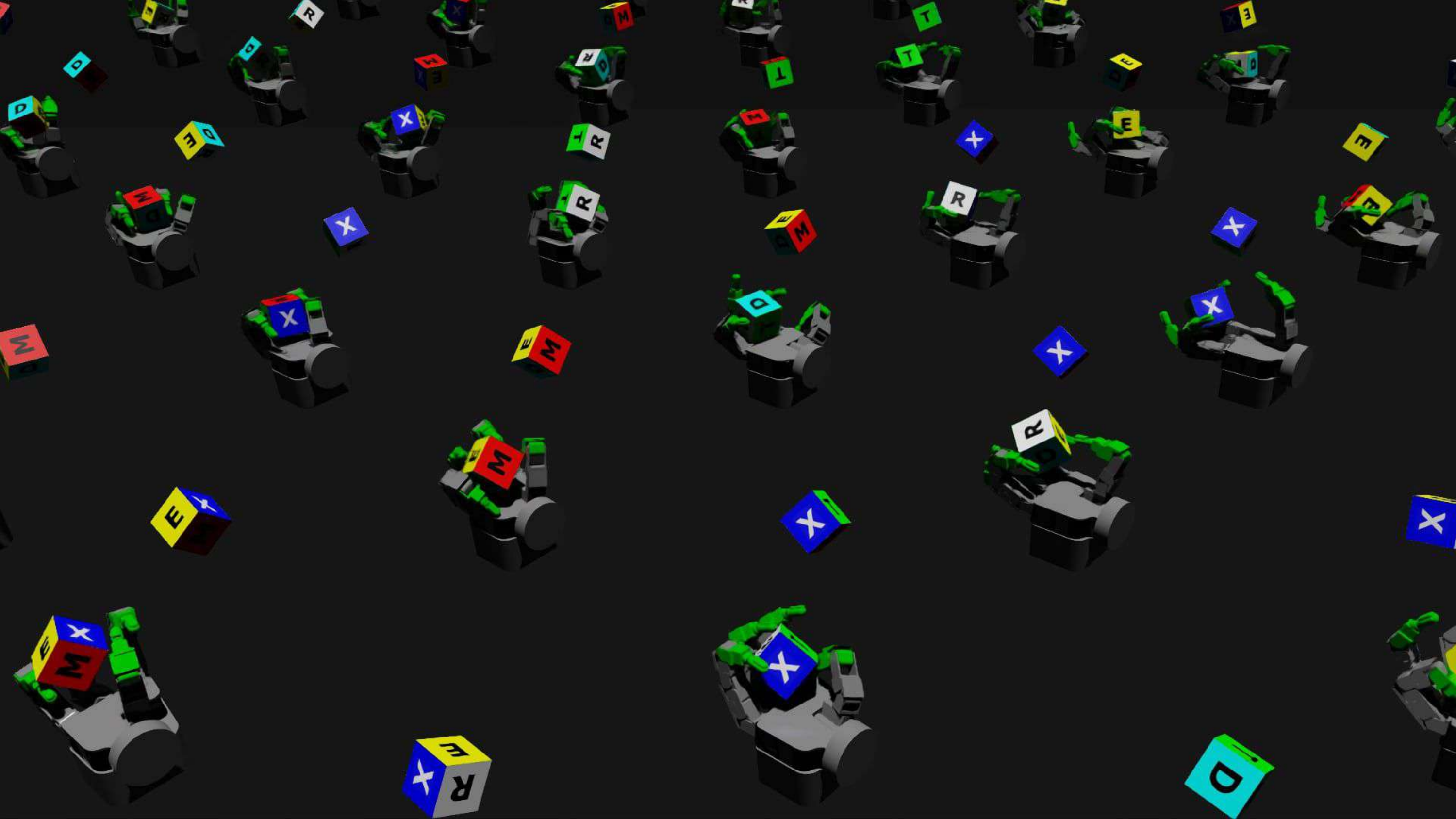} \\ \textbf{Allegro Cube}} & In-hand reorientation of a cube using Allegro hand. The state and action spaces defined as $s \in \mathbb{R}^{157}, a \in \mathbb{R}^{20}$ & 
& \makecell{\includegraphics[width=2.0cm]{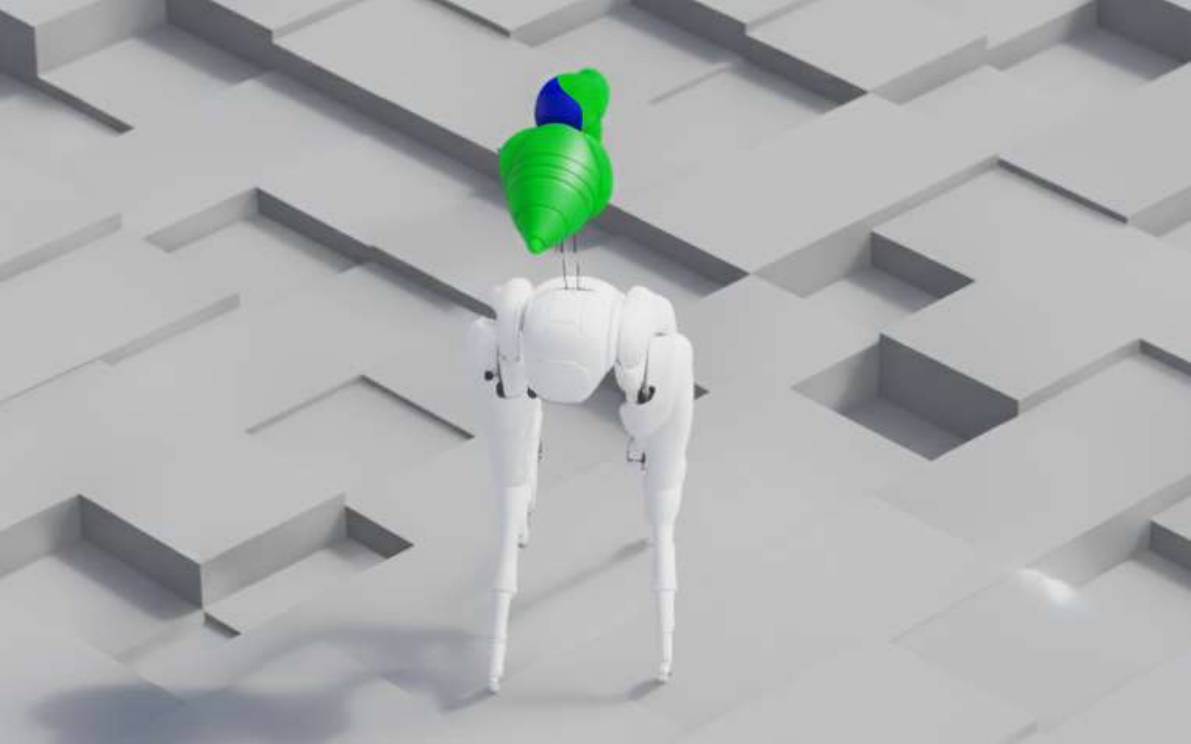} \\ \textbf{Cassie}} & Track a velocity command on rough terrain with the bipedal Cassie via LiDAR exteroception. The state and action spaces defined as $s \in \mathbb{R}^{235}, a \in \mathbb{R}^{12}$ \\

\bottomrule
\end{tabular}}
\end{table*}

\section{Analysis}
\label{sec:analysis}

In this section, we present a comprehensive simulation-based analysis of representative Online DPRL methods.
Our evaluation focuses on five key components critical for real-world robotic deployment: Task Diversity, Environment Parallelization Capability, Computational Complexity, Cross-Embodiment Generalization, and Environmental Robustness.
The primary objective of these experiments is to systematically identify the inherent trade-offs between different algorithmic architectures.
By evaluating each method within a unified benchmark, we uncover the specific strengths and critical bottlenecks that characterize current Online DPRL approaches.
This empirical study provides a rigorous performance profile for each methodology, offering a data-driven foundation for algorithm selection and future architectural refinements in the field of diffusion-based reinforcement learning.

\subsection{Experimental Setup}
\label{subsec:setup}

\subsubsection{Simulation Environment and Task Selection}
% \noindent\textbf{Simulation Environment and Task Selection}
We utilize the NVIDIA Isaac Lab benchmark~\cite{isaaclab}, which is specifically designed as a large-scale robot learning framework.
Isaac Lab's environments consist of various tasks, primarily involving locomotion and manipulation.
From these, we select 12 distinct Isaac Lab tasks spanning diverse robot specifications and task types, including 3 classic tasks, 6 locomotion tasks, and 3 manipulation tasks.
For the locomotion tasks, the observation configuration consists of 2 tasks configured with only proprioception (internal state, such as joint angles, velocities), and 4 tasks configured with both proprioception and LiDAR exteroception (external state).
We summarize the details related to task selection and state-action space in Tab.~\ref{tab:isaac_lab_tasks_integrated}.

\begin{figure*}[t]
  \centering
  \includegraphics[width=0.99\linewidth]{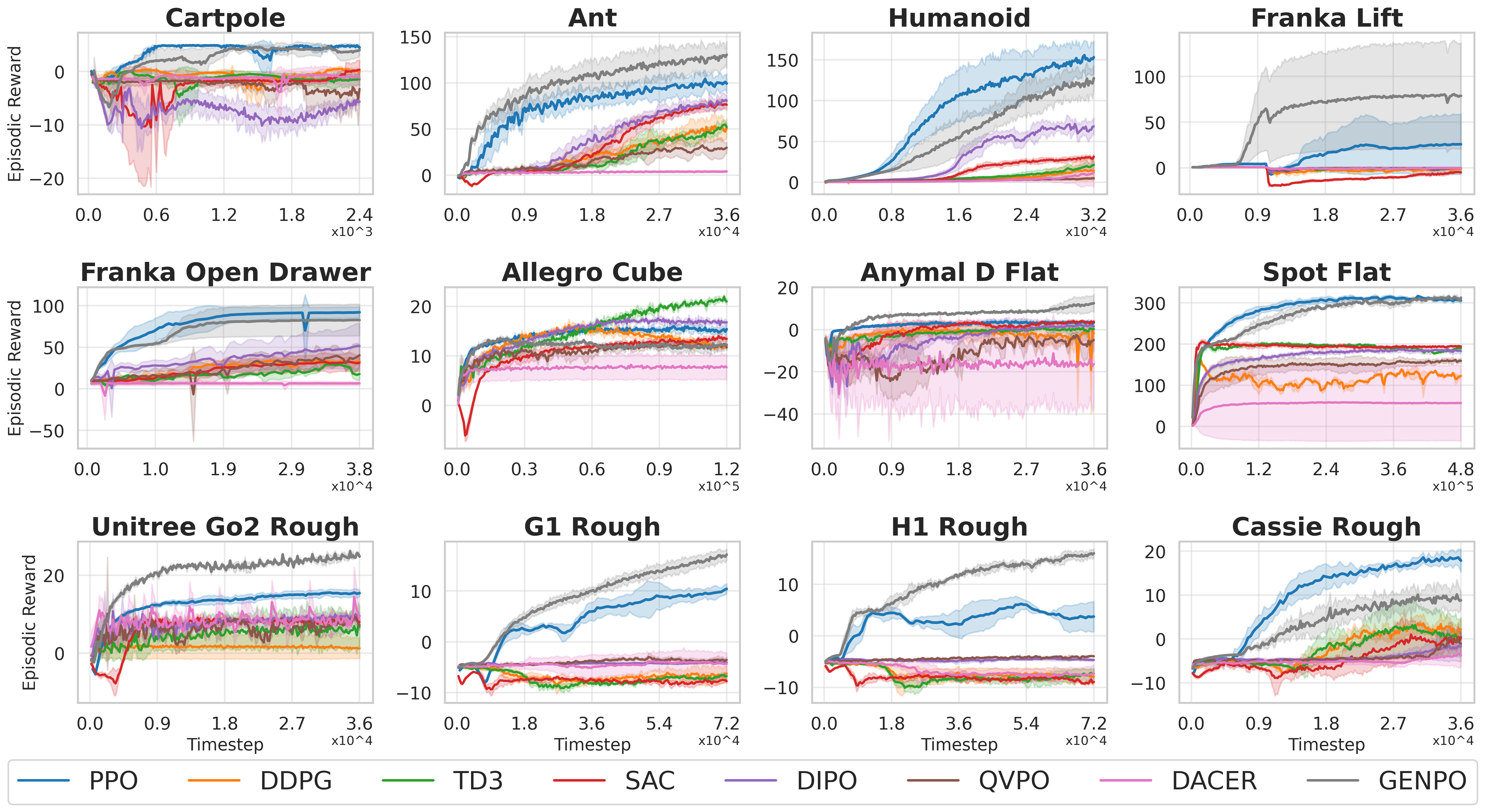}
  \vspace{-5pt}
  \caption{Training curves of online RL and DPRL methods across various Isaac Lab tasks. The solid lines represent the mean episodic reward averaged over five independent runs, with the shaded regions indicating the variance. All experiments were conducted using 1,024 parallelized environments to ensure high-throughput data collection.}
  \label{fig:reward_curve}
\end{figure*}
\begin{table*}[t]
\centering
\caption{Comparison of the average final rewards of online DPRL baselines with some prevalent RL methods in IsaacLab Benchmarks. The mean and standard deviation of the five runs are summarized. \textbf{Best} and \underline{second-best} results are highlighted.}
\vspace{-5pt}
\resizebox{0.99\linewidth}{!}{
\label{tab:task_diversity}
\small
\begin{tabular}{l|rrrr|rrrr}

\toprule
\multirow{2}{*}{\textbf{Environment}} & \multicolumn{4}{c|}{\textbf{online RL (conventional)}} & \multicolumn{4}{c}{\textbf{online DPRL}} \\
\cmidrule(lr){2-5} \cmidrule(lr){6-9}
& \multicolumn{1}{c}{PPO~\cite{ppo}} & \multicolumn{1}{c}{DDPG~\cite{ddpg}} & \multicolumn{1}{c}{TD3~\cite{td3}} & \multicolumn{1}{c|}{SAC~\cite{sac}} & \multicolumn{1}{c}{DIPO~\cite{dipoerror}} & \multicolumn{1}{c}{QVPO~\cite{qvpoerror}} & \multicolumn{1}{c}{DACER~\cite{dacer}} & \multicolumn{1}{c}{GenPO~\cite{genpo}} \\
\midrule
Cartpole           & \textbf{4.72 $\pm$ 0.06}   & 0.14 $\pm$ 0.30  & -1.55 $\pm$ 1.38 & 0.13 $\pm$ 2.02  & -5.88 $\pm$ 0.55 & -3.91 $\pm$ 4.06 & -1.02 $\pm$ 1.87 & \underline{3.79 $\pm$ 1.45} \\
Ant                & \underline{99.25 $\pm$ 11.50} & 50.71 $\pm$ 12.19 & 51.98 $\pm$ 2.90 & 76.33 $\pm$ 6.37 & 80.60 $\pm$ 6.16 & 29.72 $\pm$ 6.39 & 3.97 $\pm$ 1.40  & \textbf{130.40 $\pm$ 15.90} \\
% Humanoid           & \textbf{153.09 $\pm$ 22.16} & 14.24 $\pm$ 14.61 & 21.25 $\pm$ 5.45 & 31.32 $\pm$ 2.95 & \underline{68.35 $\pm$ 11.71} & 4.91 $\pm$ 1.24  & 11.72 $\pm$ 19.01 & 23.36 $\pm$ 15.24 \\
Humanoid           & \textbf{153.09 $\pm$ 22.16} & 14.24 $\pm$ 14.61 & 21.25 $\pm$ 5.45 & 31.32 $\pm$ 2.95 & 68.35 $\pm$ 11.71 & 4.91 $\pm$ 1.24  & 11.72 $\pm$ 19.01 & \underline{127.33 $\pm$ 18.33} \\
Franka Lift        & \underline{25.88 $\pm$ 40.40} & -1.38 $\pm$ 0.97  & -0.16 $\pm$ 0.24 & -4.80 $\pm$ 1.82 & -0.06 $\pm$ 0.12 & -0.23 $\pm$ 0.60 & -0.17 $\pm$ 0.07 & \textbf{78.82 $\pm$ 71.01} \\
Franka Open Drawer & \textbf{92.03 $\pm$ 7.22}   & 32.83 $\pm$ 4.15  & 17.79 $\pm$ 10.00 & 31.37 $\pm$ 11.21 & 51.46 $\pm$ 30.94 & 40.36 $\pm$ 14.62 & 6.51 $\pm$ 2.67  & \underline{82.64 $\pm$ 23.22} \\
Allegro Cube       & 15.37 $\pm$ 1.31  & 11.92 $\pm$ 0.21  & \textbf{20.98 $\pm$ 0.66} & 13.45 $\pm$ 0.69 & \underline{16.77 $\pm$ 0.36} & 11.69 $\pm$ 0.45 & 7.67 $\pm$ 2.82  & 12.23 $\pm$ 1.66  \\
Anymal D Flat      & 3.41 $\pm$ 1.46   & -1.83 $\pm$ 5.32  & 0.29 $\pm$ 1.79  & \underline{3.80 $\pm$ 0.23}  & 1.55 $\pm$ 1.42  & -4.92 $\pm$ 6.13 & -16.41 $\pm$ 22.85 & \textbf{12.55 $\pm$ 5.07} \\
Spot Flat          & \underline{305.88 $\pm$ 6.86} & 122.74 $\pm$ 0.51 & 189.75 $\pm$ 9.82 & 194.22 $\pm$ 2.10 & 182.83 $\pm$ 6.66 & 159.32 $\pm$ 7.17 & 56.83 $\pm$ 105.77 & \textbf{312.54 $\pm$ 2.34} \\
Unitree Go2 Rough  & \underline{15.40 $\pm$ 1.00}  & 1.24 $\pm$ 3.20   & 6.77 $\pm$ 6.50  & 8.03 $\pm$ 2.06  & 9.53 $\pm$ 0.69  & 7.22 $\pm$ 1.62  & 8.58 $\pm$ 2.33  & \textbf{24.77 $\pm$ 0.12} \\
% G1 Rough           & \textbf{10.41 $\pm$ 1.20}  & -6.24 $\pm$ 1.65  & -6.76 $\pm$ 0.58 & -7.76 $\pm$ 0.43 & -4.09 $\pm$ 0.26 & \underline{-3.57 $\pm$ 0.66} & -4.38 $\pm$ 2.44 & -5.94 $\pm$ 1.76 \\
G1 Rough           & \underline{10.41 $\pm$ 1.20}  & -6.24 $\pm$ 1.65  & -6.76 $\pm$ 0.58 & -7.76 $\pm$ 0.43 & -4.09 $\pm$ 0.26 & -3.57 $\pm$ 0.66 & -4.38 $\pm$ 2.44 & \textbf{17.15 $\pm$ 1.24} \\
% H1 Rough           & \underline{3.73 $\pm$ 3.60}   & -7.90 $\pm$ 1.49  & -7.32 $\pm$ 1.21 & -8.96 $\pm$ 0.46 & -4.69 $\pm$ 0.27 & -3.98 $\pm$ 0.01 & -7.30 $\pm$ 1.86 & \textbf{4.17 $\pm$ 9.49}  \\
H1 Rough           & \underline{3.73 $\pm$ 3.60}   & -7.90 $\pm$ 1.49  & -7.32 $\pm$ 1.21 & -8.96 $\pm$ 0.46 & -4.69 $\pm$ 0.27 & -3.98 $\pm$ 0.01 & -7.30 $\pm$ 1.86 & \textbf{16.02 $\pm$ 1.05}  \\
Cassie Rough       & \textbf{17.87 $\pm$ 3.14}  & 2.23 $\pm$ 2.84   & 0.46 $\pm$ 5.45  & 0.32 $\pm$ 2.12  & -1.69 $\pm$ 4.20 & -1.10 $\pm$ 3.15 & -3.77 $\pm$ 2.82 & \underline{8.82 $\pm$ 2.45} \\
\midrule
\textbf{Mean Rank}
% & \textbf{1.92} & 6.58 & 5.50 & 4.33 & 3.58 & 5.33 & 6.67 & \underline{2.08} \\
& \underline{1.83} & 5.83 & 5.08 & 5.08 & 4.08 & 5.83 & 6.58 & \textbf{1.67} \\

\bottomrule
\end{tabular}}
\end{table*}

\subsubsection{Baseline Algorithms}

Our benchmarking suite consists of six representative algorithms, including both classical online RL and online DPRL approaches.
% For classical RL baselines, we select four representative methods: Proximal Policy Optimization (PPO)~\cite{ppo}---a proximity-based trust region on-policy method---and Soft Actor-Critic (SAC)~\cite{sac}---an off-policy algorithm based on maximum entropy RL.
We select four representative methods for our evaluation: Proximal Policy Optimization (PPO)~\cite{ppo}, a proximity-based trust region on-policy method; Deep Deterministic Policy Gradient (DDPG)~\cite{ddpg}, a fundamental off-policy algorithm for continuous action spaces; Twin Delayed Deep Deterministic Policy Gradient (TD3)~\cite{td3}, which improves upon DDPG by addressing overestimation bias; and Soft Actor-Critic (SAC)~\cite{sac}, an off-policy framework based on the maximum entropy RL principle.
For online DPRL approaches, we select one representative method from each of the four categories in our taxonomy: DIPO~\cite{dipoerror} for action-gradient methods, QVPO~\cite{qvpoerror} for Q-weighting approaches, GenPO~\cite{genpo} for proximity-based methods (also incorporates BPTT), and DACER~\cite{dacer} for BPTT-based approaches.

Since manually tuning hyperparameters for the proposed method may introduce implicit bias, we unify general hyperparameters (e.g., model size, replay buffer size, discount factor) while preserving each method's unique or sensitive hyperparameters (e.g., number of diffusion steps, expected entropy) to the greatest extent possible.
Detailed hyperparameter settings can be found in the Appendix.

% REVIEWERS comment: I’d like to highlight a potential fairness issue. Manually tuning hyperparameters for the proposed method may introduce implicit bias, especially if equivalent tuning efforts were not made for the baselines. In general, I would recommend using an automated hyperparameter optimization procedure with an equal budget across all compared methods to ensure a fair comparison. Yet, it seems (from what I’ve seen looking at other papers) that many of the baselines were originally designed and evaluated on these benchmarks, which likely reduces this concern in practice.

\subsubsection{Experimental Design}
% \noindent\textbf{Experimental Design}
\label{subsec:exp_design}

To rigorously evaluate the performance and applicability of Online DPRL algorithms, we design our experiments based on the critical factors that influence the standard robot learning and deployment pipeline.
Our analysis is structured around five key components, each addressing a fundamental challenge in modern robotics as follows:

% \paragraph{Task Diversity and Scalability}
\noindent\textbf{Task Diversity and Scalability}
Robotic learning encompasses a wide array of scenarios, including locomotion, manipulation, and navigation, so an online model-free DPRL algorithm should demonstrate consistent performance across these diverse domains.
% We conduct experiments across multiple task categories to evaluate the versatility of each methodology in handling different reward structures and action dimensionalities.

% \paragraph{Environmental Parallelization Capability}
\noindent\textbf{Environmental Parallelization Capability}
The efficiency of online learning is often dictated by the degree of environment parallelization, which can be limited by the availability of computational resources or the complexity of the dynamics itself, such as large-scale diffusion architectures~\cite{rdt1b, diffusionvla, hybridvla, dexvla} or the complexity of exteroceptive observations (e.g., high-resolution RGB images).
% We investigate the performance trade-offs as a function of the number of parallel environments to understand how different DPRL approaches scale and whether they can maintain sample efficiency under constrained parallelization.

% \paragraph{Impact of Diffusion Sampling Steps on Latency}
\noindent\textbf{Impact of Diffusion Sampling Steps on Latency}
Robotic controllers often operate under strict latency requirements (e.g., $>50$ Hz) to ensure stability and safety.
Since the number of denoising steps in diffusion models directly dictates inference speed, it is crucial to analyze the sensitivity of the policy to reduced sampling budgets.
% We evaluate how performance fluctuates across varying diffusion steps to identify which algorithms remain robust or become vulnerable when pushed toward real-time operational constraints.

% \paragraph{Cross-Embodiment Generalizability}
\noindent\textbf{Cross-Embodiment Generalizability}
A significant gap often exists between simulation specifications and real-world robot configurations.
Furthermore, the emergence of generalist agents necessitates robustness to variations in robot morphology.
% We test each method in cross-embodiment scenarios to measure its ability to generalize across different robot specifications and hardware gaps, moving beyond simple within-domain evaluation.

\noindent\textbf{Out-of-Distribution (OOD) Environments Generalizability}
For robots incorporating exteroception, the ability to handle environmental variations (e.g., changes in backgrounds or object textures) is vital.
% We assess the OOD generalization performance of Online DPRL methods by introducing environmental perturbations and visual shifts, thereby evaluating the robustness of the learned representations against unforeseen real-world complexities.

\begin{figure*}[t]
  \centering
  \includegraphics[width=0.99\linewidth]{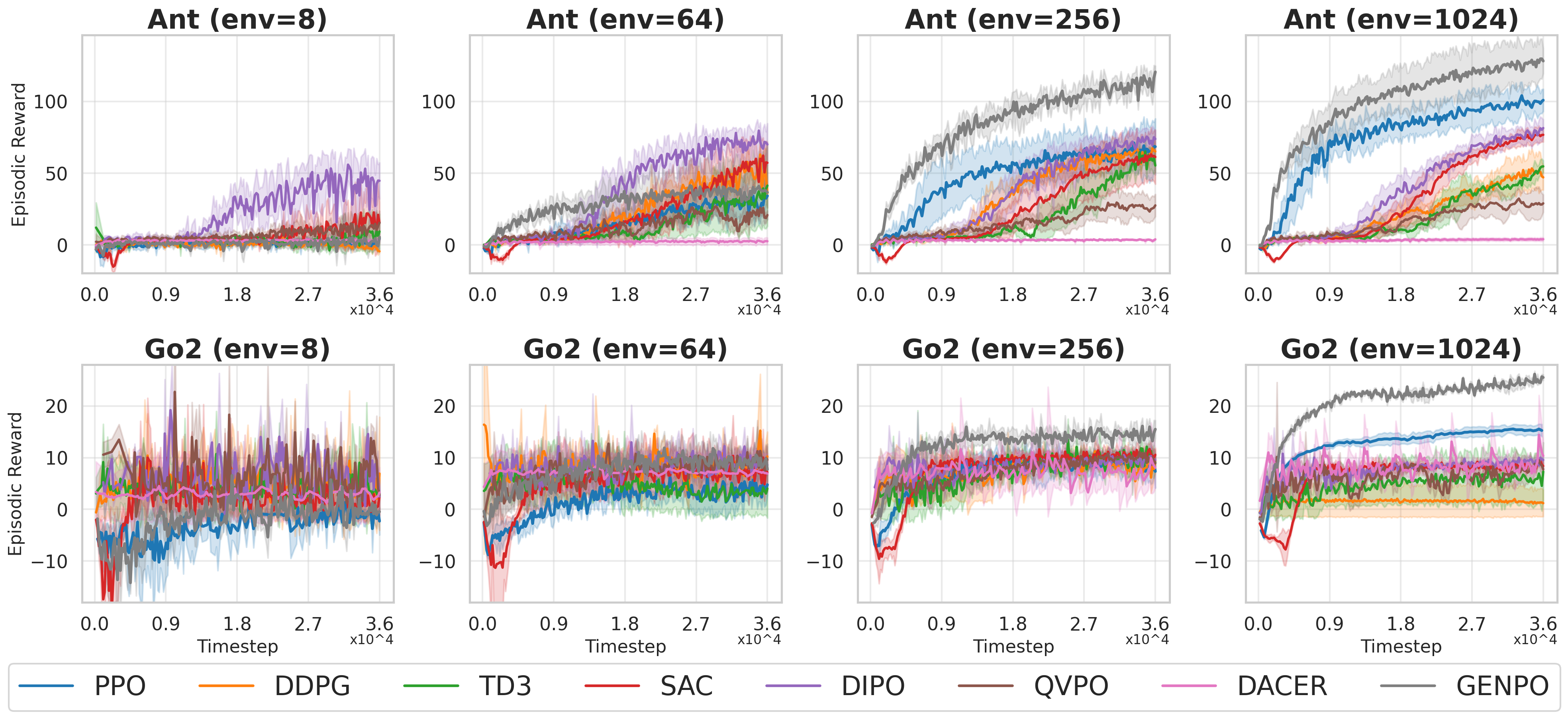}
  \vspace{-5pt}
  \caption{Training curves of online RL and DPRL methods across the number of simulation environments on the \textit{Ant} and \textit{Unitree Go2} environments. The solid lines represent the mean episodic reward averaged over five independent runs, with the shaded regions indicating the variance.}
  \label{fig:num_env}
\end{figure*}
\begin{table*}[t]
\centering
\caption{Comparison of the average final rewards of online DPRL baselines with some prevalent RL methods w.r.t. number of environments on \textit{Ant} and \textit{Unitree Go2} tasks. The mean and standard deviation of the five runs are summarized. \textbf{Best} and \underline{second-best} results are highlighted.}
\vspace{-5pt}
\resizebox{0.99\linewidth}{!}{
\label{tab:num_env}
\footnotesize
\begin{tabular}{c|l|rrrr|rrrr}
% \begin{table*}[h]
% \centering
% \caption{Comparison of Performance Across Different Numbers of Environments}
% \label{tab:env_num_comparison}
% \resizebox{\linewidth}{!}{
% \begin{tabular}{l c c c c | c c c c}
\toprule
& \multirow{2}{*}{\textbf{\# of Env.}} & \multicolumn{4}{c|}{\textbf{online RL (conventional)}} & \multicolumn{4}{c}{\textbf{online DPRL}} \\
% \cmidrule(lr){2-5} \cmidrule(lr){6-9}
\cmidrule{3-10}
& & \multicolumn{1}{c}{PPO~\cite{ppo}} & \multicolumn{1}{c}{DDPG~\cite{ddpg}} & \multicolumn{1}{c}{TD3~\cite{td3}} & \multicolumn{1}{c|}{SAC~\cite{sac}} & \multicolumn{1}{c}{DIPO~\cite{dipoerror}} & \multicolumn{1}{c}{QVPO~\cite{qvpoerror}} & \multicolumn{1}{c}{DACER~\cite{dacer}} & \multicolumn{1}{c}{GenPO~\cite{genpo}} \\
\midrule

\multirow{4}{*}{\rotatebox[origin=c]{90}{\textit{\textbf{Ant}}}}
& 8 Env
& 4.26 $\pm$ 1.63 & -4.53 $\pm$ 1.45 & 9.28 $\pm$ 14.86 & \underline{15.37 $\pm$ 16.13} & \textbf{44.68 $\pm$ 14.11} & \underline{15.38 $\pm$ 9.12} & 3.55 $\pm$ 1.73 & 5.39 $\pm$ 11.85 \\
& 64 Env
& 34.12 $\pm$ 8.99 & 36.13 $\pm$ 12.71 & 41.31 $\pm$ 34.64 & \underline{57.21 $\pm$ 16.12} & \textbf{70.04 $\pm$ 16.31} & 20.76 $\pm$ 3.64 & 2.49 $\pm$ 2.70 & 36.58 $\pm$ 21.05 \\
& 256 Env
& \underline{68.45 $\pm$ 21.87} & 65.17 $\pm$ 5.07 & 55.35 $\pm$ 5.72 & 61.28 $\pm$ 21.14 & 73.83 $\pm$ 6.48 & 27.49 $\pm$ 8.66 & 3.81 $\pm$ 0.40 & \textbf{120.58 $\pm$ 4.69} \\
& 1024 Env
& \underline{100.88 $\pm$ 8.64} & 47.05 $\pm$ 10.66 & 54.71 $\pm$ 5.03 & 76.73 $\pm$ 5.49 & 81.45 $\pm$ 6.64 & 28.90 $\pm$ 6.78 & 3.90 $\pm$ 1.31 & \textbf{128.22 $\pm$ 9.65} \\
% & 8 Env
% & 2.24 $\pm$ 1.97 & & & \underline{14.63 $\pm$ 20.20} & \textbf{49.18 $\pm$ 19.80} & 14.39 $\pm$ 7.68 & 3.44 $\pm$ 2.01 & 3.39 $\pm$ 9.35 \\

% & 64 Env
% & 30.41 $\pm$ 16.44 & & & \underline{52.80 $\pm$ 19.88} & \textbf{74.97 $\pm$ 5.67} & 18.22 $\pm$ 5.76 & 2.03 $\pm$ 2.65 & 42.14 $\pm$ 29.85 \\

% & 256 Env
% & 60.72 $\pm$ 13.60 & & & 55.65 $\pm$ 17.66 & \underline{69.14 $\pm$ 3.62} & 28.03 $\pm$ 7.63 & 3.49 $\pm$ 0.24 & \textbf{104.94 $\pm$ 8.27} \\

% & 1024 Env
% & \underline{98.73 $\pm$ 16.04} & & & 72.48 $\pm$ 5.31 & 75.96 $\pm$ 4.74 & 29.99 $\pm$ 8.84 & 4.04 $\pm$ 1.47 & \textbf{127.53 $\pm$ 15.28} \\

\midrule
\multirow{4}{*}{\rotatebox[origin=c]{90}{\textit{\textbf{Go2}}}}
& 8 Env
& -2.27 $\pm$ 3.26 & \textbf{6.92 $\pm$ 12.21} & \underline{3.08 $\pm$ 3.51} & 2.47 $\pm$ 4.45 & 2.51 $\pm$ 3.97 & 2.80 $\pm$ 4.17 & \underline{3.24 $\pm$ 5.98} & -0.45 $\pm$ 0.73 \\
& 64 Env
& 4.76 $\pm$ 2.22 & \textbf{10.23 $\pm$ 5.82} & 4.02 $\pm$ 6.33 & 7.11 $\pm$ 4.01 & 8.59 $\pm$ 1.74 & \underline{9.58 $\pm$ 3.52} & 6.93 $\pm$ 4.28 & 7.72 $\pm$ 4.89 \\
& 256 Env
& 7.34 $\pm$ 3.58 & 9.47 $\pm$ 2.90 & 9.07 $\pm$ 2.15 & \underline{10.54 $\pm$ 1.10} & 9.53 $\pm$ 1.43 & 8.89 $\pm$ 1.75 & 8.05 $\pm$ 4.36 & \textbf{15.49 $\pm$ 1.90} \\
& 1024 Env
& \underline{15.25 $\pm$ 1.44} & 1.20 $\pm$ 3.15 & 7.30 $\pm$ 6.88 & 8.37 $\pm$ 1.45 & 9.55 $\pm$ 0.75 & 7.74 $\pm$ 1.46 & 10.34 $\pm$ 3.22 & \textbf{25.49 $\pm$ 0.68} \\
% & 8 Env
% & & & & & & & & \\

% & 64 Env
% & & & & & & & & \\

% & 256 Env
% & & & & & & & & \\

% & 1024 Env
% & & & & & & & & \\
\bottomrule
\end{tabular}}
\end{table*}

\subsection{Experiment 1: General Performance w.r.t. Various Tasks}
\label{subsec:exp1}
First, we evaluate the versatility of each methodology across 12 diverse task categories.
Fig.~\ref{fig:reward_curve} depicts the mean episodic reward curves throughout the training phase, while Tab.~\ref{tab:task_diversity} presents the corresponding final KPIs.
These experimental results provide several key insights regarding the scalability and efficacy of online RL algorithms.

% 전반적으로 충분히 많은 수의 에이전트 병렬화가 된 시뮬레이션 환경에서는 proximity-based의 on-policy algorithm인 PPO와 GenPO가 좋은 성능을 보인다.
% While PPO serves as a strong baseline, GenPO demonstrates superior peak performance by securing the top rank in 6 out of 12 tasks, compared to PPO's 5 top-rank finishes.
% 특히 GenPO는 Humanoid, Allegro cube, G1 Rough 같은 몇몇 task에서는 PPO에 비해서 떨어지는 성능을 보였지만, Franka Lift, Anymal-D, Unitree Go2같은 몇몇 task에서 PPO에 비해 x~y\%의 significant performance improvement를 보임으로써 on-policy DPRL algorithm이 robotic task의 performance 고점을 뚫을 수 있다는 가능성을 보여준다.
In environments with extensive agent parallelization, proximity-based on-policy algorithms, specifically PPO and GenPO, demonstrate superior overall performance.
% While PPO serves as a robust and high-performance baseline with a Mean Rank of 1.83, GenPO exhibits a higher performance ceiling, securing the top rank in 6 out of 12 tasks (compared to 5 for PPO).
While PPO serves as a robust and high-performance baseline with a Mean Rank (\textbf{MR}) of 1.83, GenPO achieves the highest overall performance ceiling (\textbf{MR}=1.67), securing the top rank in 7 out of 12 tasks (compared to 4 for PPO).
% Although GenPO showed lower performance than PPO in specific tasks such as \textit{Humanoid}, \textit{Allegro Cube}, and \textit{G1 Rough}, it achieved significant performance gains---ranging from $1.51\times$ to $5.43\times$---in tasks like \textit{Franka Lift}, \textit{Anymal-D}, and \textit{Unitree Go2}.
Although PPO shows slightly better results in specific tasks like \textit{Cartpole}, \textit{Humanoid}, and \textit{Cassie}, GenPO demonstrates superior scalability and robustness in more complex scenarios.
Notably, GenPO achieves substantial performance gains in high-dimensional and challenging environments, particularly in complex humanoid locomotion tasks involving \textit{Unitree G1} and \textit{H1} robots on rough terrain.
These results underscore the potential of on-policy DPRL to transcend the representational limitations of conventional Gaussian policies in complex robotic tasks.

% On the other hand, off-policy algorithm은 on-policy algorithm에 비해 낮은 학습 수렴속도와, suboptimal performance를 보인다.
% 이는 off-policy algorithms이 struggle to track rapidly shifting data distributions within their replay buffers 이기 때문이다~\cite{genpo}.
% 또한 diffusion policy를 incorporating한 DIPO, QVPO의 경우, SAC가 significant하게 failure mode를 보인 complex humanoid locomotion tasks (\textit{G1 Rough} and \textit{H1 Rough}) 에서 suboptimal하지만 robust한 training reward curve를 보여주면서, complex task에 대한 robust policy representation에 대한 잠재적 가능성을 보인다.
Conversely, off-policy algorithms generally exhibit slower convergence and suboptimal asymptotic performance compared to their on-policy counterparts. This performance gap is primarily attributed to the challenge of tracking rapidly shifting data distributions within replay buffers under high-throughput parallelization~\cite{genpo}.
However, we observe a notable advantage of incorporating diffusion policies in complex locomotion tasks.
Among the off-policy DPRL methods, DIPO achieved a \textbf{MR} of 4.08, significantly outperforming traditional off-policy algorithms such as DDPG (\textbf{MR}=5.83), TD3 (\textbf{MR}=5.08), and SAC (\textbf{MR}=5.08), which is widely considered a strong classical baseline in off-policy RL.
% This performance gap demonstrates that integrating Diffusion models into the reinforcement learning framework yields substantial and meaningful performance gains.
% This suggests that the expressive power of Diffusion Models provides a more resilient policy representation for high-dimensional, complex robotic control.
These results demonstrate that the robust expressive capacity of Diffusion Models can effectively address the limitations of conventional off-policy RL, offering a more resilient framework for complex control tasks.

\begin{figure*}[t]
  \centering
  \includegraphics[width=0.99\linewidth]{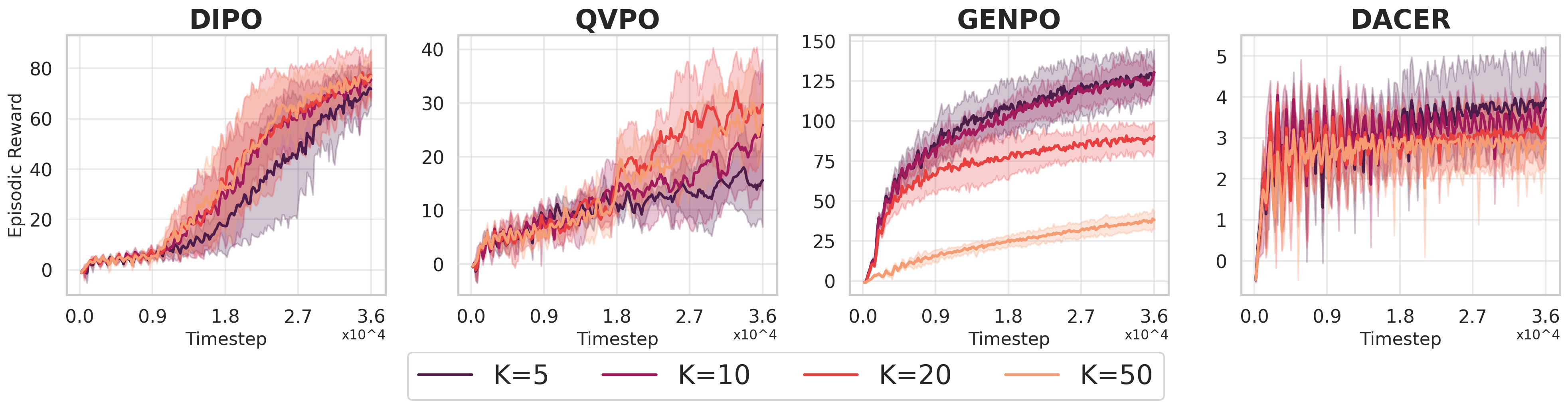}
  \vspace{-5pt}
  \caption{Training curves of online DPRL methods across the number of diffusion timesteps on the Ant environment. The solid lines represent the mean episodic reward averaged over five independent runs, with the shaded regions indicating the variance.}
  \label{fig:diffusion_step}
\end{figure*}

% 또한, Action-gradient method인 DIPO에 비해서 Q-weighting method와 BPTT-based method인 QVPO와 DACER는 off-policy DPRL중 가장 낮은 성능을 보여주면서 한계점을 보인다.
% QVPO의 경우 reward scale sensitivities에, as the $Q$-related reweighting function is directly coupled with the magnitude of manually designed rewards, 그렇기에 잠재적으로 suboptimal performance를 낸다고 사료된다.
% DACER는 대부분의 task에서 학습에 실패해 제일 낮은 성능을 보여주는데,
% 이는 policy gradient가 high variance를 가지고 있고~\cite{pgvar}, Q-value자체로 직접 BPTT로 diffusion을 학습하는 것이 매우 어렵기 때문에 학습에 실패한 것으로 보인다~\cite{bengio1994learning, gradclip, ddpograd}.
A closer examination of the four DPRL paradigms reveals distinct limitations:
DIPO consistently outperforms QVPO and DACER.
The suboptimal performance of QVPO (\textbf{MR}=5.83) is likely due to its sensitivity to reward scaling; since the $Q$-related reweighting function is directly coupled with the magnitude of manually designed rewards, it struggles with inconsistent reward landscapes across different tasks.
DACER recorded the lowest performance (\textbf{MR}=6.58), failing to converge in the majority of tasks.
This failure can be attributed to the inherent high variance of policy gradients~\cite{pgvar} and the extreme difficulty of optimizing diffusion models directly through BPTT using noisy Q-values as a supervisory signal, which often leads to unstable gradient dynamics~\cite{bengio1994learning, ddpograd}.

% 우리는 또한 각각의 RL baselines가 task마다 필요로 하는 training wall-time을 \todo{Fig. X}에 summarize한다.
% 모든 training cost는 single gpu NVIDIA Quadro RTX A6000의 동일한 조건에서 측정한다.
% {경향성 설명, PPO < SAC < GenPO < DACER < DIPO < QVPO, diffusion incorporating한 setup이 당연하게도 느린 속도를 보였으나, plausible한 속도를 보임 등,,}

\subsection{Experiment 2: Parallelization Capability}
\label{subsec:exp2}

To investigate the scalability and sample efficiency of various DPRL approaches, we evaluate algorithm robustness across a spectrum of parallelization settings. Specifically, we vary the number of parallel environments from 8 to 1024 on the \textit{Ant} and \textit{Unitree Go2} tasks.

As illustrated in Fig.~\ref{fig:num_env} and Tab.~\ref{tab:num_env}, our empirical results reveal a significant disparity in how algorithms respond to environment scaling.
While high-throughput algorithms like PPO and GenPO excel in heavily parallelized settings (e.g., 1024 environments), their performance diminishes sharply as parallelization is restricted.
Specifically, in the \textit{Ant} task, PPO and GenPO experience a catastrophic performance drop of 95.77\% and 95.79\%, respectively, when the environment count is reduced from 1024 to 8.
A similar trend is observed in the \textit{Unitree Go2} task, where GenPO's performance collapses in the 8-environment setup, effectively failing to converge.

In contrast, off-policy-based DPRL methods, most notably DIPO, demonstrate remarkable robustness to limited parallelization.
DIPO maintains a substantial final reward even in the most constrained 8-environment setting, outperforming the high-throughput GenPO by nearly 8.3 times in the \textit{Ant} task.
Interestingly, classical off-policy baselines such as DDPG and SAC also show lower sensitivity to the environment scale compared to on-policy methods, though they often exhibit higher variance or fail to match the peak performance of diffusion-based models in high-parallelization regimes.

These findings suggest a clear trade-off: while GenPO and PPO are optimized for high-throughput scalability, DIPO is uniquely suited for scenarios where extensive parallelization is computationally prohibitive.
This is particularly relevant for training overparameterized models, such as Vision-Language-Action (VLA) architectures~\cite{openvla}, where the memory and computational costs of maintaining thousands of concurrent environments are often infeasible.

Furthermore, our analysis highlights a critical gap in current online DPRL literature, which predominantly reports performance at a single, fixed scale~\cite{qsm, dipoerror, qvpoerror, dpmdsdac, dacer}.
Our results underscore that environment scale is not merely a hyperparameter but a decisive factor in algorithmic ranking.
Consequently, future benchmarking in the DPRL domain should incorporate diverse parallelization scales to ensure a comprehensive assessment of both scalability and robustness.

\begin{table}[t]
\centering
% \caption{Ablation study on the effect of diffusion denoising steps ($K$) for online DPRL baselines. Experiments are conducted on the \textit{Ant} task with a fixed number of environments. \textbf{Best} and \underline{second-best} results are highlighted.}
\caption{Comparison of the average final rewards of online DPRL baselines w.r.t. diffusion denoising steps $K$ on \textit{Ant} task. The mean and standard deviation of the five runs are summarized. \textbf{Best} and \underline{second-best} results are highlighted.}
\resizebox{0.99\columnwidth}{!}{
\label{tab:diffusion_steps}
\small
\begin{tabular}{c|llll}
\toprule
\multirow{2}{*}{\textbf{Steps}} 
% \textbf{Denoising} 
& \multicolumn{4}{c}{\textbf{online DPRL}} \\
\cmidrule{2-5}
% \textbf{Steps} 
& \multicolumn{1}{c}{DIPO~\cite{dipoerror}} & \multicolumn{1}{c}{QVPO~\cite{qvpoerror}} & \multicolumn{1}{c}{DACER~\cite{dacer}} & \multicolumn{1}{c}{GenPO~\cite{genpo}} \\
\midrule

$K=5$  & \underline{60.06 $\pm$ 14.65} & 14.57 $\pm$ 6.22 & 3.74 $\pm$ 1.20 & \textbf{124.09 $\pm$ 17.22} \\
$K=10$ & \underline{67.78 $\pm$ 9.89}  & 19.49 $\pm$ 9.18 & 3.38 $\pm$ 0.46 & \textbf{122.67 $\pm$ 11.71} \\
$K=20$ & \underline{69.90 $\pm$ 14.99} & 27.76 $\pm$ 7.64 & 3.07 $\pm$ 0.80 & \textbf{87.28 $\pm$ 9.92}   \\
$K=50$ & \textbf{71.04 $\pm$ 9.24}  & 24.52 $\pm$ 6.91 & 2.85 $\pm$ 0.78 & \underline{34.65 $\pm$ 5.75} \\

\bottomrule
\end{tabular}}
\end{table}
\begin{figure*}[t]
  \centering
  \includegraphics[width=0.99\linewidth]{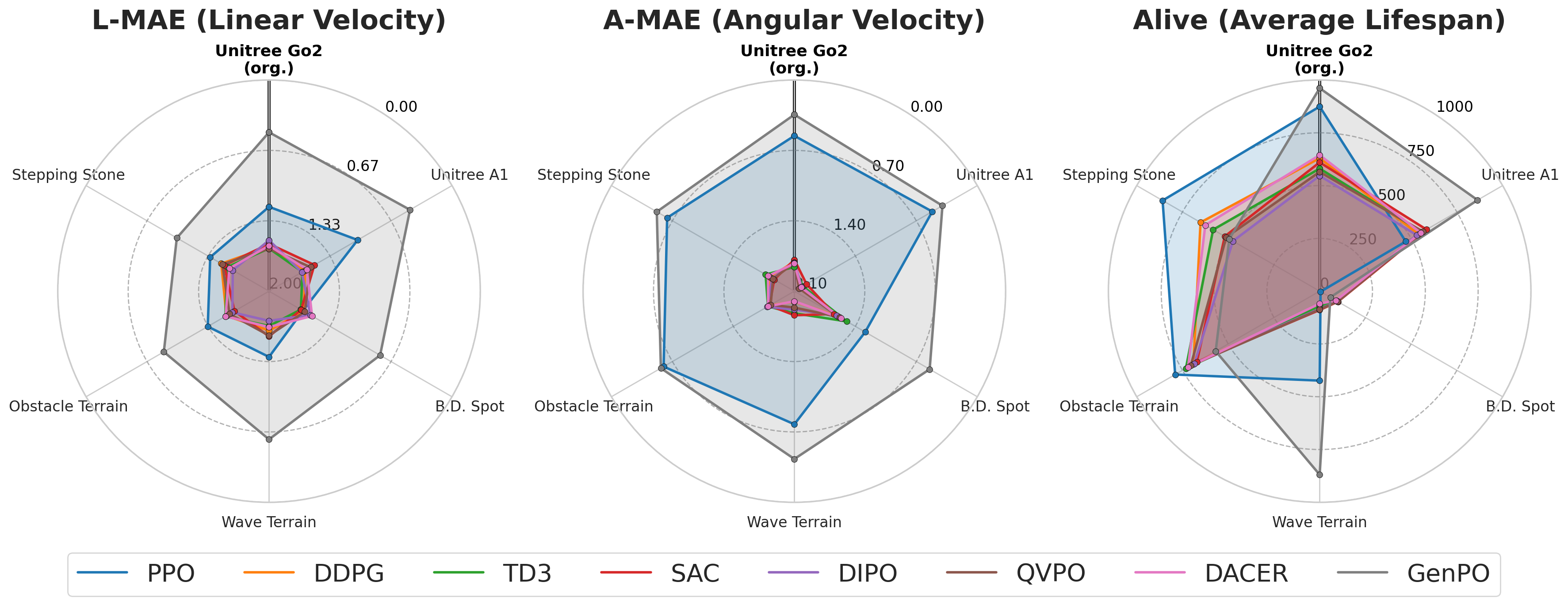}
  \vspace{-5pt}
  \caption{Robustness analysis via radar charts across cross-embodiment and environmental variations. Each radar chart visualizes performance in terms of Linear MAE, Angular MAE, and Average Lifespan. The top vertex represents the in-distribution (trained) scenario, while the remaining axes denote out-of-distribution (OOD) testing conditions, such as cross-embodiments or OOD environments. A larger, more symmetric hexagonal area indicates superior zero-shot transferability and balanced robustness against domain shifts.}
  \label{fig:cross_embodiement}
\end{figure*}
\begin{table*}[t]
\centering
\caption{Zero-shot performance under cross-embodiment and OOD terrain scenarios. (Upper) Transfer results from the source robot (Unitree Go2) to target platforms (Unitree A1, Boston Dynamics Spot). (Lower) Generalization metrics on novel, unseen rough terrains as illustrated in Tab.~\ref{tab:isaac_lab_terrains}. \\ \textbf{Best} and \underline{second-best} results are highlighted.}
\vspace{-5pt}
\label{tab:generalizability}
\footnotesize
\begin{tabular}{l|ccc|ccc|ccc}
\toprule
Metrics & Alive $\uparrow$ & L-MAE $\downarrow$ & A-MAE $\downarrow$ & Alive $\uparrow$ & L-MAE $\downarrow$ & A-MAE $\downarrow$ & Alive $\uparrow$ & L-MAE $\downarrow$ & A-MAE $\downarrow$ \\
\midrule
\textbf{Cross-Embodiment} & \multicolumn{3}{c|}{Go2 (\textit{Original})} & \multicolumn{3}{c|}{A1 Transfer} & \multicolumn{3}{c}{Spot Transfer} \\
\midrule
PPO~\cite{ppo} & \underline{874.32} & \underline{1.201} & \underline{0.555} & 470.40 & \underline{1.031} & \underline{0.517} & 4.92 & 1.630 & \underline{1.287} \\
DDPG~\cite{ddpg} & 629.96 & 1.580 & 1.806 & 539.40 & 1.607 & 2.026 & 96.04 & 1.619 & 1.599 \\
TD3~\cite{td3} & 580.63 & 1.595 & 1.859 & 559.83 & 1.635 & 2.002 & 90.61 & 1.652 & 1.498 \\
SAC~\cite{sac} & 610.05 & 1.552 & 1.787 & \underline{583.63} & 1.504 & 1.960 & \textbf{99.49} & 1.649 & 1.645 \\
DIPO~\cite{dipoerror} & 546.78 & 1.518 & 1.816 & 530.75 & 1.641 & 2.016 & 97.49 & \underline{1.548} & 1.617 \\
QVPO~\cite{qvpoerror} & 564.87 & 1.597 & 1.820 & 564.27 & 1.545 & 2.045 & \underline{98.62} & 1.610 & 1.581 \\
DACER~\cite{dacer} & 644.37 & 1.569 & 1.825 & 552.73 & 1.587 & 2.023 & 88.93 & 1.530 & 1.562 \\
GenPO~\cite{genpo} & \textbf{961.39} & \textbf{0.495} & \textbf{0.344} & \textbf{862.80} & \textbf{0.459} & \textbf{0.400} & 58.30 & \textbf{0.786} & \textbf{0.547} \\
\bottomrule
\toprule
\textbf{OOD environment} & \multicolumn{3}{c|}{Wave Terrain} & \multicolumn{3}{c|}{Obstacle} & \multicolumn{3}{c}{Stepping stone} \\
% Method & Alive & L-MAE & A-MAE & Alive & L-MAE & A-MAE & Alive & L-MAE & A-MAE & Alive & L-MAE & A-MAE \\
\midrule
PPO~\cite{ppo} & \underline{423.68} & \underline{1.377} & \underline{0.776} & \textbf{789.82} & \underline{1.330} & \underline{0.600} & \textbf{858.58} & \underline{1.357} & \underline{0.640} \\
DDPG~\cite{ddpg} & 64.06 & 1.627 & 1.996 & 690.03 & 1.534 & 1.814 & 651.14 & 1.478 & 1.821 \\
TD3~\cite{td3} & 70.54 & 1.670 & 1.880 & 731.15 & 1.522 & 1.797 & 582.41 & 1.543 & 1.770 \\
SAC~\cite{sac} & 89.21 & 1.571 & 1.860 & 671.47 & 1.622 & 1.815 & 518.65 & 1.511 & 1.869 \\
DIPO~\cite{dipoerror} & 83.36 & 1.719 & 1.917 & 685.57 & 1.599 & 1.788 & 474.08 & 1.604 & 1.845 \\
QVPO~\cite{qvpoerror} & 86.36 & 1.580 & 1.937 & 704.72 & 1.568 & 1.828 & 510.62 & 1.490 & 1.855 \\
DACER~\cite{dacer} & 58.93 & 1.660 & 2.000 & \underline{719.18} & 1.525 & 1.799 & \underline{625.11} & 1.567 & 1.798 \\
GenPO~\cite{genpo} & \textbf{869.12} & \textbf{0.596} & \textbf{0.431} & 568.34 & \textbf{0.850} & \textbf{0.568} & 493.95 & \textbf{0.992} & \textbf{0.521} \\
\bottomrule
\end{tabular}
\vspace{-5pt}
\end{table*}

% \begin{table*}[t]
% \centering
% \caption{Environment diversity generalization results across terrain types.}
% \label{tab:env_diversity}
% \small
% \begin{tabular}{l|ccc|ccc|ccc|ccc}
% \toprule
% & \multicolumn{3}{c|}{Original} & \multicolumn{3}{c|}{Wave} & \multicolumn{3}{c|}{Obstacle} & \multicolumn{3}{c}{Stepstone} \\
% Method & Alive & L-MAE & A-MAE & Alive & L-MAE & A-MAE & Alive & L-MAE & A-MAE & Alive & L-MAE & A-MAE \\
% \midrule
% PPO & 874.32 & 1.201 & 0.555 & 423.68 & 1.377 & 0.776 & \textbf{789.82} & 1.33 & 0.6 & \textbf{858.58} & 1.357 & 0.64 \\
% SAC & 610.05 & 1.552 & 1.787 & 89.21 & 1.571 & 1.86 & 671.47 & 1.622 & 1.815 & 518.65 & 1.511 & 1.869 \\
% DIPO & 546.78 & 1.518 & 1.816 & 83.36 & 1.719 & 1.917 & 685.57 & 1.599 & 1.788 & 474.08 & 1.604 & 1.845 \\
% QVPO & 564.87 & 1.597 & 1.82 & 86.36 & 1.58 & 1.937 & \textbf{704.72} & 1.568 & 1.828 & 510.62 & 1.49 & 1.855 \\
% GenPO & \textbf{961.39} & \textbf{0.495} & \textbf{0.344} & \textbf{869.12} & \textbf{0.596} & \textbf{0.431} & 568.34 & \textbf{0.85} & \textbf{0.568} & 493.95 & \textbf{0.992} & \textbf{0.521} \\
% \bottomrule
% \end{tabular}
% \end{table*}
\begin{table}[h]
\centering
\caption{Summary of source and target (OOD) terrains. Source terrains represent the in-distribution environments used for policy optimization, whereas the target terrains consist of structurally diverse, unseen geometries designed for zero-shot transfer performance testing.}
\label{tab:isaac_lab_terrains}
\resizebox{\columnwidth}{!}{
\renewcommand{\arraystretch}{1.6}
\scriptsize
\begin{tabular}{l c m{4.3cm}}

\toprule
 & \textbf{Terrain} & \textbf{Description} \\
\midrule

% --- Row 1 ---
\multirow{4}{*}{\rotatebox[origin=c]{90}{\textbf{Source Terrains~~~~~~~~~~~~~~~~~~~~~~~~~~~}}}
& \makecell{\includegraphics[width=2.1cm]{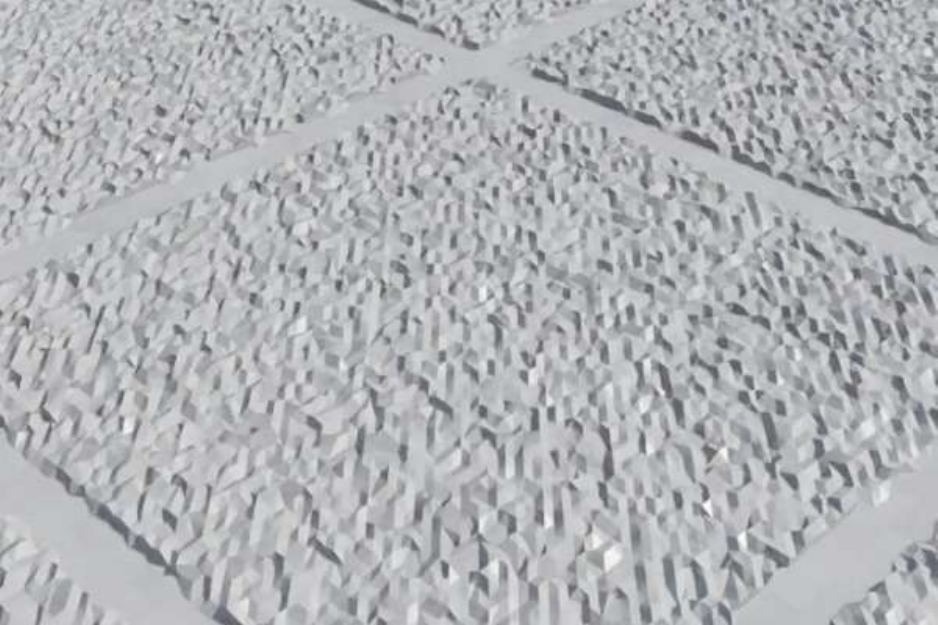} \\ \textbf{Random}} 
% & Terrain with height sampled uniformly from a specified range. \\ \cmidrule{2-3}
& Generated by sampling heights uniformly from a predefined range, demanding rapid reactive balancing to compensate for unpredictable ground irregularities. \\
\cmidrule{2-3}

& \makecell{\includegraphics[width=2.1cm]{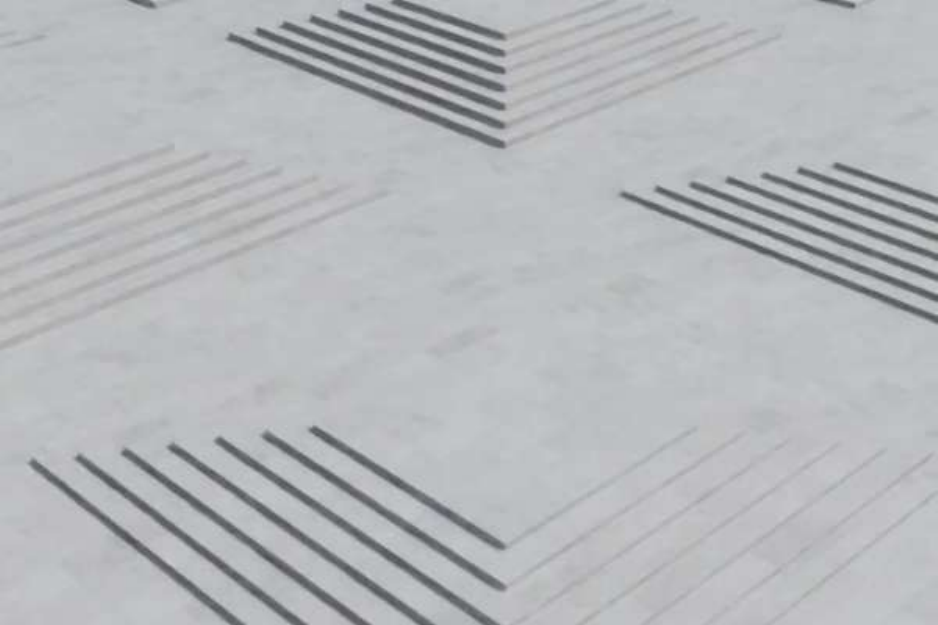} \\ \textbf{Stairs}} 
% & Terrain with a pyramid stair pattern. \\
& Features a structured stair pattern in a pyramidal arrangement. Navigating this terrain requires precise leg lifting and coordination to overcome discrete vertical steps. \\
\cmidrule{2-3}

& \makecell{\includegraphics[width=2.1cm]{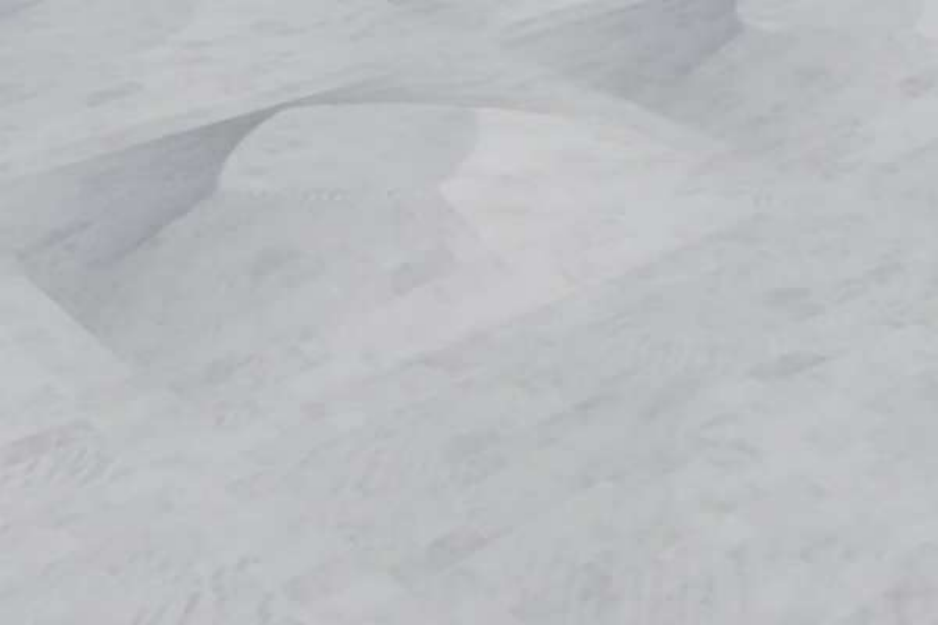} \\ \textbf{Pyramid}} & Comprises a continuous incline leading to a truncated flat platform. This setup necessitates sustained postural stability and torque control to maintain constant velocity on persistent slopes. \\
\cmidrule{2-3}

& \makecell{\includegraphics[width=2.1cm]{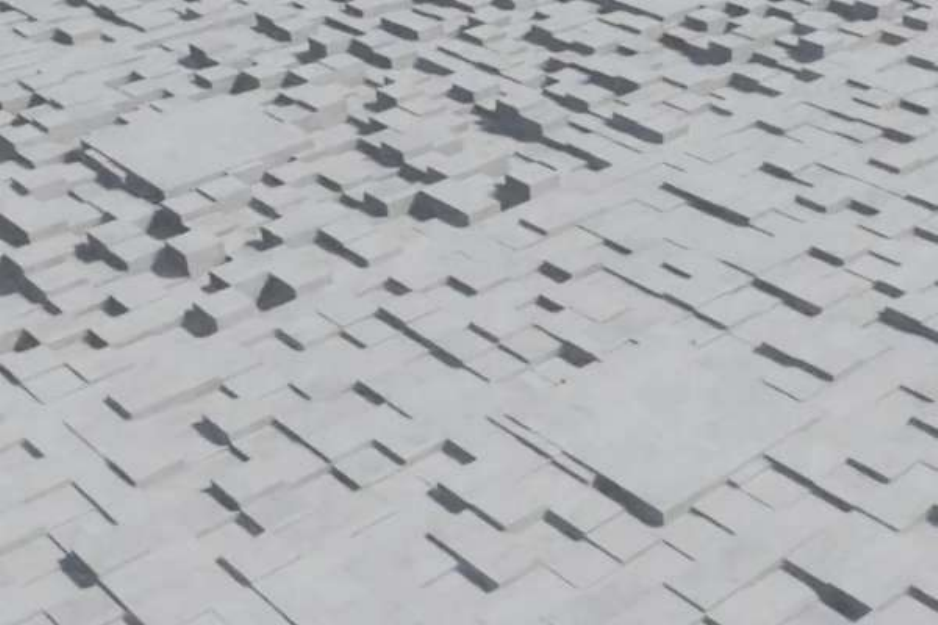} \\ \textbf{Box}} & Constructed with boxes of fixed width and random heights. This discontinuous environment requires agile foot placement to avoid tripping. \\

\midrule

\multirow{3}{*}{\rotatebox[origin=c]{90}{\textbf{Target Terrains~~~~~~~~~~~~~~~}}}
& \makecell{\includegraphics[width=2.1cm]{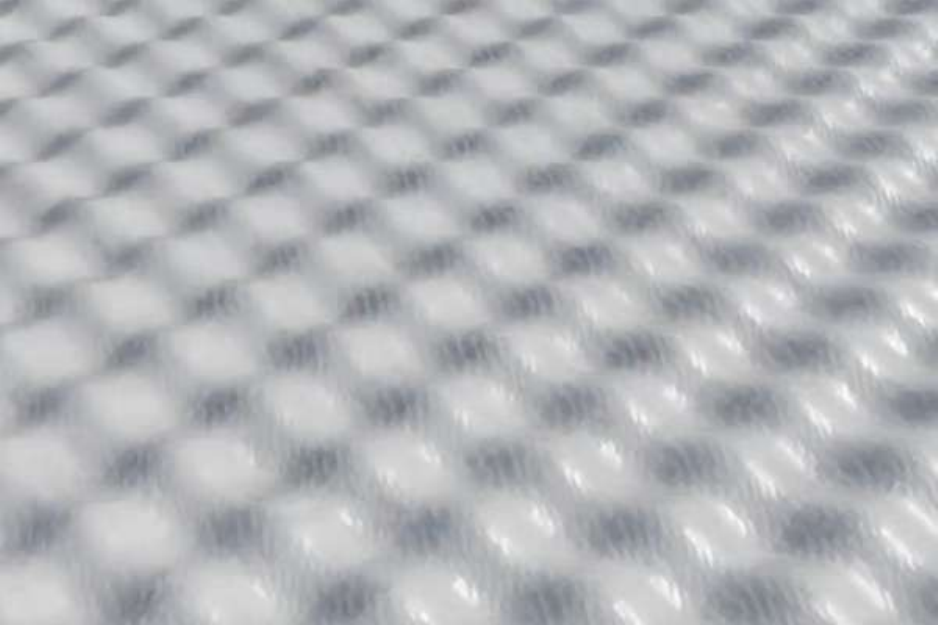} \\ \textbf{Wave}} & 
Formed with a smooth, oscillating sinusoidal height pattern. Following this contour demands continuous gait adaptation to handle non-linear and undulating transitions. \\
% Terrain with a wave pattern. \\ 
\cmidrule{2-3}

& \makecell{\includegraphics[width=2.1cm]{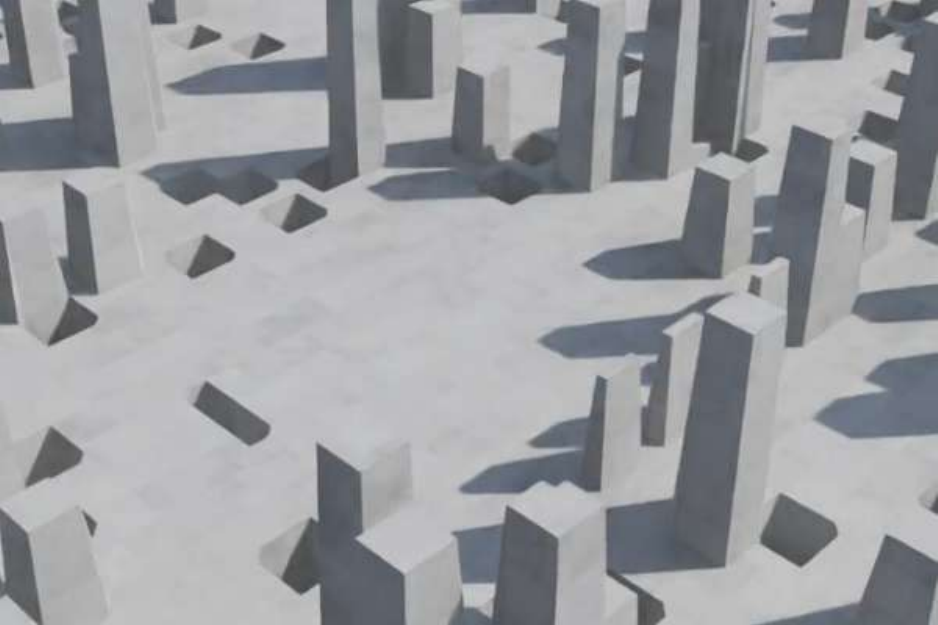} \\ \textbf{Obstacle}} & Contains randomly distributed pillars with varying height offsets. This environment necessitates sophisticated collision avoidance and spatial maneuverability. \\ \cmidrule{2-3}

& \makecell{\includegraphics[width=2.1cm]{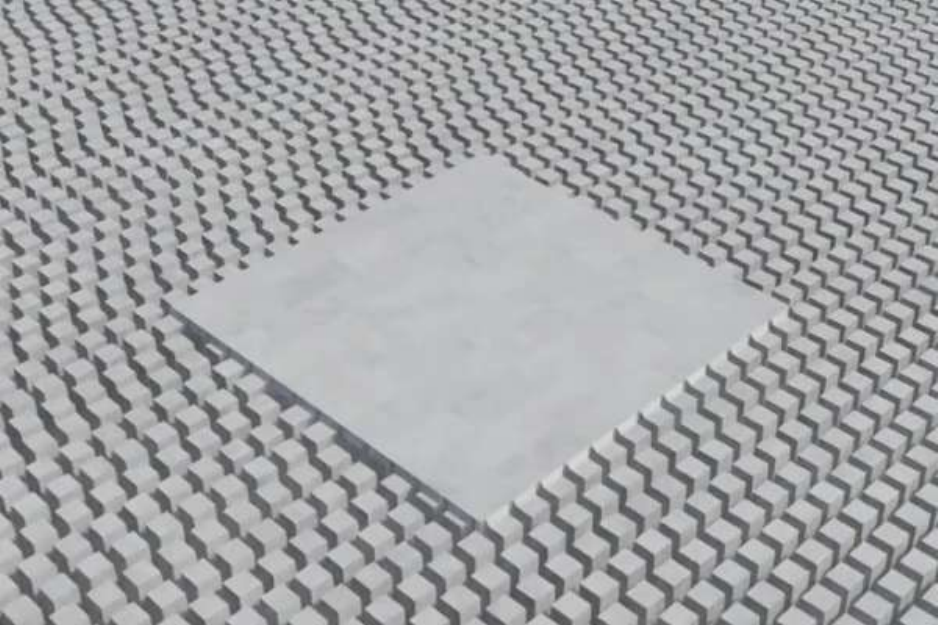} \\ \textbf{Stepstone}} & Consists of isolated stepping stones separated by gaps. This task demands a high-precision landing to navigate across intermittent contact points. \\

\bottomrule

\end{tabular}}
% \vspace{-10pt}
\end{table}

\subsection{Experiment 3: Diffusion Step Scalability}
\label{subsec:exp3}
% We evaluate how performance fluctuates across varying diffusion sampling steps to identify which algorithms remain robust or become vulnerable when pushed toward real-time operational constraints.
% To this end, we conduct experiments by scaling the diffusion sampling steps of online DPRL algorithms in the classic Ant task, as shown in Fig.~\ref{fig:diffusion_step}.

The number of diffusion sampling steps is a critical determinant that governs the balance between the operational speed and the representational quality of the model.
% Diffusion sampling step number is the critical component 모델의 속도와 퀄리티를 결정하는.
We evaluate the sensitivity of performance to varying diffusion sampling steps, $K$, to identify which algorithms remain robust under real-time operational constraints.
To this end, we conduct experiments by scaling $K$ for various online DPRL algorithms in the classic \textit{Ant} task, as summarized in Fig.~\ref{fig:diffusion_step} and Tab.~\ref{tab:diffusion_steps}.

Algorithms utilizing a surrogate diffusion loss—such as DIPO and QVPO—demonstrate a positive correlation between performance and the number of diffusion timesteps.
Specifically, DIPO achieves its peak reward and most stable convergence as $K$ increases toward 50.
In contrast, QVPO tends to exhibit saturated performance beyond a specific threshold (i.e., $K$$=$20).
These trends suggest that for methods where policy optimization is decoupled from the denoising chain during backpropagation, a higher $K$ provides a more refined and expressive policy representation without compromising training stability.
% Algorithms that utilize a surrogate diffusion loss (e.g., DIPO and QVPO) demonstrate a positive correlation between performance and the number of diffusion timesteps.
% Specifically, DIPO achieves its highest reward and most stable convergence when $K$ is increased toward 50, while QVPO tend to demonstrate saturated performance of a specific diffusion sampling step (i.e., K=20).
% This trend suggests that for methods decoupled from the denoising chain during backpropagation, a higher $K$ provides a more refined and expressive policy representation without compromising training stability.
% Algorithms employing surrogate diffusion loss for training (DIPO, QVPO) exhibit performance that scales proportionally with increased diffusion timesteps.

Conversely, DACER and GenPO, which perform backpropagation through all timesteps during training, exhibit a distinct inverse scalability pattern.
While GenPO achieves state-of-the-art performance at lower step counts ($K$$=$5, 10), its performance deteriorates significantly as $K$ increases, eventually suffering a catastrophic drop at $K$$=$50.
We attribute this failure to the numerical instability and gradient vanishing inherent in backpropagating through an excessively long denoising chain~\cite{bengio1994learning, gradclip}.
Indeed, training diffusion models via BPTT is notoriously difficult, as the recursive nature of the denoising process compounds gradient-related issues as the chain length grows~\cite{ddpograd}.
Interestingly, DACER exhibits the most erratic behavior; despite being a BPTT-based method, it fails to achieve meaningful reward levels regardless of the $K$ value.
This suggests that its BPTT objective may fundamentally have a scalability issue with end-to-end diffusion optimization in online DPRL tasks.
% In stark contrast, DACER and GenPO, which perform backpropagation through all timesteps during training, exhibit a distinct inverse scalability pattern.
% While GenPO achieves state-of-the-art performance at lower steps (K=5, K=10), its performance deteriorates significantly as K increases, eventually suffering a catastrophic drop at K=50.
% We attribute this to gradient vanishing during backpropagation through the entire long denoising chain~\cite{bengio1994learning, gradclip}.
% Training diffusion model with BPTT is notoriously difficult and numerically unstable.
% This limitation represents a fundamental scalability bottleneck for BPTT-based methods or requires End-to-End BPTT and constrains their ability to utilize more expressive policy representations.

% \input{tables/environment_diversity}

\subsection{Experiment 4: Cross-embodiment Generalizability}
\label{subsec:exp4}

To evaluate cross-embodiment generalizability, we assess zero-shot transfer capabilities by deploying policies trained on a Unitree Go2 to target platforms with varying degrees of morphological divergence. We categorize the target environments based on the magnitude of the hardware gap: (1) the Unitree A1, which shares the same manufacturer as the source robot and exhibits minor morphological differences; and (2) the Boston Dynamics Spot, which originates from a different manufacturer. The Spot presents significant hardware disparities, testing the algorithm's robustness despite the shared quadrupedal action space.

We select three Key Performance Indicators (KPI) to evaluate the quantitative performance of the policies: Average Lifespan (Alive) before falling, linear mean absolute error (L-MAE), and angular mean absolute error (A-MAE) between commanded and actual movement.
Note that policies are trained on the source robot with proprioception and LiDAR exteroception and deployed directly to target robots without fine-tuning.

Fig.~\ref{fig:cross_embodiement} and Tab.~\ref{tab:generalizability} present the quantitative metrics for cross-embodiment transfer.
As expected, performance degradation scales proportionally with the magnitude of the robot specification gap.
During transfer to the Unitree A1, which possesses only a minor specification gap due to its shared manufacturer lineage, most algorithms---with the exception of PPO---maintain a performance trend consistent with their baseline on the source robot.
Furthermore, GenPO, which leverages the expressive power of Diffusion Models, exhibits a far more robust Average Lifespan (Alive) compared to PPO, which suffers a significant decrease in survival duration.

% Table~\ref{tab:generalizability} and Fig.~\ref{fig:cross_embodiement} present quantitative metrics for cross-embodiment transfer.
% First, as expected, performance degradation scales with robot specification gap magnitude.
% 같은 제조사라서 minor robot specification gap을 가진 Unitree A1으로의 transfer에서는 PPO를 제외한 모든 알고리즘들이 source robot에서의 성능과 비슷한 경향성을 보인다.
% 또한 diffusion model을 leveraging한 GenPO의 경우 Alive metric에서 significant decreasement를 보인 PPO에 비해 robust한 Alive metric 감소율을 보인다.

While algorithms show a consistent trend of performance degradation when deployed to the Unitree A1, the impact is far more pronounced upon transferring to the Boston Dynamics Spot.
Specifically, the significant hardware disparity of the Spot platform triggers a catastrophic failure, leading to a near-total loss of stability as evidenced by the drastic lifespan reductions for both GenPO (961.39 $\rightarrow$ 58.30) and PPO (874.32 $\rightarrow$ 4.92).

Off-policy algorithms demonstrate substantially greater resilience, maintaining functional performance despite suboptimal control accuracy (QVPO alive count: 564.87 $\rightarrow$ 98.62).
Consequently, visual inspection of deployment behaviors reveals distinct failure modes.
On-policy algorithms exhibit rapid instability and early task failure when facing large specification gaps, while off-policy methods maintain functional, albeit suboptimal, performance for extended durations.
This vulnerability of on-policy methods in transfer scenarios questions their reliability for real-world deployment, where robot platform variations are common.

% \begin{figure}[t]
%   \centering
%   \includegraphics[width=0.99\columnwidth]{figures/terrains.png}
%   \caption{DEMO FIGURE: 일반화 성능을 위해 사용된 challenging terrains의 visualization}
%   \label{fig:terrains}
% \end{figure}

\subsection{Experiment 5: OOD Environment Generalizability}
\label{subsec:exp5}

Lastly, we assess the generalization performance of RL methods on OOD environments by introducing environmental perturbations and visual shifts, thereby evaluating the robustness of the learned representations against unforeseen real-world complexities.
Drawing a parallel to cross-embodiment experiments, we conduct zero-shot transfer across various challenging terrains using a Unitree Go2 model as the source policy, which was trained on rough terrain and incorporates LiDAR exteroception.
To evaluate the precision of center-of-mass control, obstacle avoidance, and foot-placement planning, we measure the policy's KPIs (Alive, L-MAE, and A-MAE) across three target terrains as illustrated in Tab.~\ref{tab:isaac_lab_terrains}: wave, obstacle, and stepping stones.
% Similar to cross-embodiement experiments, 우리는 Rough terrain에서 학습되고 LiDAR exteroception를 incorporating한 Unitree Go2 model을 source policy로 다양한 challenging terrains에 대해 zero-shot transfer를 진행한다.
% 우리는 무게중심 컨트롤이나 물체 회피, 발위치 planning의 정밀성을 평가하기 위해서 Figure~\ref{fig:terrains} 와 같은 세가지 terrains including wave, obstacle, stepping stones의 target terrain에서, trained policy의 KPIs (i.e., Alive, L-MAE, and A-MAE)를 evaluate한다.
% Locomotion task의 KPI로는 우리는 넘어지기 전까지의 Average Lifespan time (Alive), command와 실제 이동의 linear mean absolute error (L-MAE), angular mean absolute error (A-MAE) 총 3가지의 metric을 측정한다.

% Table~\ref{tab:env_diversity} and 
 % summarizes quantitative performance across terrain types.
As shown in Fig.~\ref{fig:cross_embodiement} and Tab.~\ref{tab:generalizability}, the overall trends follow similar patterns to the performance in the original setup, with some exceptions in certain cases.
On the wave terrain, which requires center-of-mass control to prevent falling, GenPO demonstrated high robustness, while the other algorithms experienced severe degradation in the Alive metric.
In contrast, in obstacle and stepping stone environments, we observe significant degradation in GenPO's Alive metric.
From qualitative observation of policy rollouts, we observe that GenPO exhibits excessively risky behaviors (e.g., failing to maintain sufficient distance from obstacles, following commands despite dangerous ground in stepping stones) to obtain rewards for command tracking.
We attribute this to overfitting, which is aimed at achieving optimal rewards in the source terrain, thereby emphasizing the necessity of safety-oriented RL research.

% \subsection{Summary of Key Findings}
% \label{subsec:summary}

% 결론적으로, 우리는 이전에 언급했던 실험들을 진행함으로써 총 X가지의 key finding을 summarize한다:
% % Our comprehensive analysis reveals fundamental trade-offs in online DPRL

% \begin{enumerate}
%     \item On-policy diffusion methods (GenPO) excel in highly parallelized multi-environment settings, demonstrating superior asymptotic performance when computational resources enable extensive data collection.
    
%     \item Off-policy algorithms provide critical advantages in sample efficiency under limited parallelization and exhibit substantially greater robustness to cross-embodiment transfer scenarios.
    
%     \item Stochastic loss formulations (DIPO, QVPO) enable stable scaling to increased diffusion timesteps, while proximity-based on-policy methods suffer from gradient vanishing in long denoising chains.
    
%     \item Cross-embodiment transfer represents a critical failure mode for on-policy algorithms, with performance degradation severity correlating with robot specification gap magnitude.
    
%     \item Environmental diversity generalization maintains algorithmic performance rankings from standard settings, suggesting more tractable adaptation compared to embodiment changes.
% \end{enumerate}

% These findings provide actionable insights for algorithm selection based on deployment constraints and highlight critical research directions for improving robustness and scalability of online diffusion policy RL methods.
\section{Discussion}
\label{sec:discussion}

The integration of Diffusion Policies with online Reinforcement Learning (RL) frameworks opens a new frontier in robot learning, yet it introduces unique challenges that distinguish it from traditional RL with Gaussian policies.
In this section, we discuss the inherent limitations of current on-policy and off-policy approaches and the algorithm selection rationale for various robotic training scenarios.
Furthermore, we also outline future research directions for Online DPRL.

\subsection{Topic 1. What are the current challenges of the online DPRL?}
Our empirical analysis in Sec.~\ref{sec:analysis} reveals several fundamental bottlenecks inherent in existing online DPRL paradigms. 
First, DIPO, one of the representative algorithms of action gradient methods, often suffers from suboptimal performance due to the difficulty of obtaining accurate gradients from a potentially biased Q-critic~\cite{dppo, qvpoerror}.
% First, from the perspective of \textit{Action Gradient methods} (e.g., DIPO, QSM), while they aim for precise policy improvement, they often suffer from suboptimal performance due to the difficulty of obtaining accurate gradients from a potentially biased Q-critic.
Specifically, updating actions via deterministic gradients and re-storing them into the replay buffer in DIPO may violate the Markov Decision Process (MDP) assumptions, as the stored transitions no longer reflect the current environment-agent interaction distribution~\cite{DDiffPG}.

Second, QVPO, categorized by $Q$-weighting methods, exhibits vulnerability to reward scale sensitivities, as the $Q$-related reweighting function is directly coupled with the magnitude of manually designed rewards.
For example, the derived Q-weighted variational objective for diffusion policy training cannot handle negative rewards properly~\cite{dpmdsdac}.

Third, while the Proximity-based method, GenPO, benefits from large-scale parallelization in simulation, its performance degrades significantly in resource-constrained scenarios (refer to Sec.~\ref{subsec:exp2}).
Moreover, the intractability of the exact log-likelihood in diffusion models necessitates crude approximations or specialized architectures like invertible diffusion~\cite{edict}, which imposes a bottleneck on both representational flexibility and inference speed.

Lastly, as demonstrated in Sec.~\ref{subsec:exp3}, \textit{BPTT-based methods} such as GenPO and DACER encounter severe scalability issues when the number of diffusion steps is increased.
This is primarily due to gradient instability---including vanishing and exploding gradient---which inherently arises when backpropagating through an extensively long denoising chain~\cite{bengio1994learning, gradclip}
% for complex tasks due to memory constraints and gradient instability.

% To conclude, each paradigm presents distinct trade-offs that remain to be fully addressed.
% We posit that the development of hybrid DPRL algorithms, such as DACERv2---which attempts to synthesize the strengths of action-gradient and BPTT-based methods---represents a promising direction for overcoming these inherent limitations.
In summary, while each online DPRL paradigm offers unique theoretical advantages, they collectively face significant hurdles—ranging from gradient inaccuracy and reward-scale sensitivity to computational intractability and backpropagation-related scalability limits.
Addressing these intertwined challenges is essential for transitioning from specialized simulation environments to reliable, high-performance robotic control in diverse real-world scenarios.
Moreover, as simulation environments become increasingly efficient at supporting large-scale parallelization~\cite{isaaclab, brax}, the demand for robust on-policy algorithms is expected to grow.
Given the current scarcity of literature on on-policy DPRL compared to its off-policy counterparts, we encourage future researchers to explore this relatively uncharted territory to unlock the full potential of high-throughput robotic learning.

\subsection{Topic 2. How do we select the online DPRL algorithms in specific robotic scenarios?} %특정 robotics task에서 어떠한 알고리즘을 선택해야 할까?}
The selection between off-policy and on-policy frameworks involves a strategic trade-off between sample efficiency and gradient stability.
On one hand, off-policy algorithms are advantageous in scenarios with limited parallelization capabilities, such as real-world robotic systems that permit only a single environmental agent instance.
By reusing transitions from a replay buffer, these methods maximize sample efficiency and mitigate the high cost of data acquisition, making them indispensable for resource-constrained settings.

On the other hand, on-policy frameworks are highly effective in environments where massive simulation clusters are available.
In such high-throughput settings, the ability to collect data at scale effectively masks the inherent sample inefficiency of on-policy updates.
This sheer data volume allows the agent to stabilize the high-variance gradients of the diffusion prior, ultimately facilitating more robust and consistent convergence through fresh, on-distribution samples.

From a task complexity standpoint, researchers should first consider the trade-off between policy capacity and sampling latency.
For tasks requiring high-precision planning, increasing diffusion steps is essential.
In such cases, algorithms utilizing diffusion-specific surrogate losses are more scalable than BPTT-based methods.
Notably, for navigation and locomotion tasks requiring continuous adaptation to diverse topographies, the intrinsic brittleness of on-policy exploration must be addressed, perhaps by incorporating off-policy data as a regularizer to prevent premature convergence in unstructured environments.

%-------------------------------------------------------------

\subsection{Topic 3. What is the Future Avenue of Online DPRL?}
While several studies~\cite{dppo, parl, pgd, polygrad} have explored integrating online DPRL with offline-to-online and model-based approaches, the field remains in its early stages with numerous open challenges.
To advance online DPRL toward more general robotic applications, we identify several promising research directions that leverage established RL paradigms, offering guidance for future investigation.

\subsubsection{Action Chunking and Trajectory Planning}
% While this work focuses on single-action predictions, real-world robotics often employs planning-based approaches that generate action chunks~\cite{actionchunking} or multi-step trajectories~\cite{openvla}.
While existing online DPRL methods focus on single-action per-predictions (as policies), real-world robotics often employs planning-based approaches that generate action chunks~\cite{actionchunking} or multi-step trajectories~\cite{openvla} (as planners).
Recent work on Q-Chunking~\cite{qchunking} has demonstrated Q-learning extensions to action sequences for offline-to-online RL.
Integrating temporal action chunking into online model-free DPRL could bridge reactive control and deliberative planning by leveraging diffusion models' multi-step prediction capabilities to generate temporally consistent action sequences aligned with RL objectives.
This evolution from state-to-action policies to state-to-trajectory planners would enhance sample efficiency and robustness in long-horizon tasks.

\subsubsection{Safe Reinforcement Learning Integration}
Our experimental analysis (Sec~\ref{subsec:exp5}) shows that simulation-trained policies can exhibit overly aggressive behaviors from overfitting to source domains when maximizing rewards without safety constraints.
Safe RL approaches—including constraint-based methods (e.g., Constrained Policy Optimization~\cite{cpo}), barrier functions~\cite{cbf}, and robust control~\cite{robustrl}---could naturally integrate with online DPRL to provide formal safety guarantees during exploration while maintaining diffusion-based action expressiveness, facilitating safer deployment in safety-critical applications.

\subsubsection{Multi-Agent Reinforcement Learning}
Multi-agent scenarios present exponentially greater complexity than single-agent settings due to non-stationarity, credit assignment, and coordination requirements.
However, advances in large-scale physics simulators capable of efficiently simulating thousands of agents make online Multi-Agent DPRL (MADPRL) particularly promising.
Initial approaches could integrate established MARL algorithms~\cite{mappo, maddpg, qmix} with online DPRL frameworks, while sophisticated extensions might leverage diffusion models' ability to capture complex joint action distributions for coordination strategies or employ centralized training with decentralized execution paradigms.

\subsubsection{Inverse Reinforcement Learning from Demonstrations}
Inverse Reinforcement Learning (IRL)~\cite{irl, maxentirl, gail} infers reward functions from expert demonstrations, solving the inverse problem of standard RL.
Since most diffusion policies~\cite{diffusionpolicy, mpd, diffuser} research relies on imitation learning from demonstrations, a natural extension reformulates these as reward learning problems rather than direct behavioral cloning.
This ``demonstration-to-reward-to-policy" paradigm offers key advantages: learned rewards generalize beyond demonstrated state-action pairs, provide clear objectives for online fine-tuning, and extract value from suboptimal demonstrations.
Integrating IRL with online DPRL enables a refined learning approach where offline behavioral priors are optimized through interaction, potentially surpassing both standard imitation and pure offline-to-online approaches.

\subsubsection{Hierarchical Reinforcement Learning}
Hierarchical RL (HRL)~\cite{hrl, hiro, latenthrl} decomposes complex decisions into multi-level structures where higher-level policies select among lower-level sub-policies to accomplish long-horizon tasks.
This offers compelling synergy with diffusion policies: while diffusion models excel at generating diverse, multimodal actions, their stochastic nature introduces substantial variance.
A hierarchical framework could employ a simpler higher-level policy to select among trajectory samples from lower-level diffusion sub-policies, mirroring batch action sampling~\cite{diffusionql, qvpoerror} but within a principled HRL structure.
This transforms the exploration-exploitation trade-off from action space to a more manageable sub-policy selection problem, enabling sample-efficient learning of compositional behaviors while preserving diffusion models' expressive capacity.
\section{Conclusion}
\label{sec:conclusion}

% \todo{conclusion rewriting}

This paper presents the first comprehensive review and empirical analysis of Online Diffusion Policy Reinforcement Learning algorithms, addressing the critical challenge of integrating expressive diffusion-based action representations with online reinforcement learning for scalable robotic control.
Through our proposed taxonomy, we systematically categorize existing approaches into four distinct families---\textit{Action-Gradient}, \textit{Q-Weighting}, \textit{Proximity-Based}, and \textit{BPTT-based methods}---based on their policy improvement mechanisms, providing a structured framework for understanding the current research landscape.

Our extensive experiments across 8 RL baselines and 12 diverse robotic tasks in the NVIDIA Isaac Lab benchmark reveal fundamental trade-offs characterizing different algorithmic families.
% Off-policy methods (SAC, DIPO, QVPO, DACER) demonstrate superior sample efficiency and remarkable robustness when environment parallelization is limited, maintaining functional performance even with as few as 8 parallel environments.
% This resilience makes them particularly suitable for real-world robotic training scenarios where computational resources are constrained or high-dimensional observations preclude massive parallelization.
% In contrast, on-policy frameworks (PPO, GenPO) achieve optimal performance in highly parallelized simulation environments but exhibit severe degradation when transferred across embodiments or deployed in resource-limited settings.
Also, our analysis identifies several critical bottlenecks currently impeding practical deployment.
Current online DPRL paradigms encounter distinct bottlenecks depending on their taxonomy: action-gradient methods struggle with gradient bias from critics, Q-weighting approaches exhibit high reward-scale sensitivity, proximity-based models suffer from architectural constraints due to intractable log-likelihoods, and BPTT-based methods face severe scalability and stability issues during backpropagation.
% BPTT-based methods suffer from gradient instability and memory constraints that scale poorly with increased diffusion steps, limiting their ability to leverage more expressive policy representations.
% Q-weighting approaches demonstrate vulnerability to reward scale sensitivity, while action-gradient methods face challenges obtaining accurate gradients from potentially biased critics.
% Proximity-based methods require crude approximations of intractable log-likelihoods, imposing architectural constraints limiting both representational flexibility and inference speed.

Based on these empirical findings, we provide concrete guidelines for algorithm selection: off-policy diffusion methods are recommended for edge computing, real-world training, or scenarios with expensive data acquisition, while on-policy frameworks are preferable when massive simulation clusters enable high-throughput data collection.
For tasks requiring high-precision planning over long horizons, algorithms utilizing diffusion-specific surrogate losses demonstrate superior scalability compared to BPTT-based approaches.

Looking forward, we identify five promising research directions that could substantially advance the field: (1) integrating temporal action chunking to bridge reactive control and deliberative planning, (2) incorporating safe RL principles to prevent overly aggressive behaviors observed in our transfer experiments, (3) extending to multi-agent scenarios leveraging large-scale simulation capabilities, (4) reformulating demonstrations as reward learning problems via inverse RL, and (5) employing hierarchical decomposition to manage high diffusion stochasticity while preserving expressive capacity.

As simulation platforms continue advancing in both fidelity and parallelization efficiency, and as demand for robust generalist robotic agents grows, we anticipate hybrid approaches---combining strengths from multiple algorithmic families---will play increasingly important roles.
The systematic analysis presented in this work provides both a rigorous performance profile of current methods and a roadmap for future architectural innovations in diffusion-based reinforcement learning for robotics.

% Can use something like this to put references on a page
% by themselves when using endfloat and the captionsoff option.
\ifCLASSOPTIONcaptionsoff
  \newpage
\fi

% trigger a \newpage just before the given reference
% number - used to balance the columns on the last page
% adjust value as needed - may need to be readjusted if
% the document is modified later
%\IEEEtriggeratref{8}
% The "triggered" command can be changed if desired:
%\IEEEtriggercmd{\enlargethispage{-5in}}

% references section
% can use a bibliography generated by BibTeX as a .bbl file
% BibTeX documentation can be easily obtained at:
% http://mirror.ctan.org/biblio/bibtex/contrib/doc/
% The IEEEtran BibTeX style support page is at:
% http://www.michaelshell.org/tex/ieeetran/bibtex/
\bibliographystyle{IEEEtran}
% argument is your BibTeX string definitions and bibliography database(s)
\bibliography{bibtex/bib/IEEEabrv,bibtex/bib/IEEEexample}

% BIOGRAPHY
% \input{scripts/7_biography}

\clearpage
% if have a single appendix:
%\appendix[Proof of the Zonklar Equations]
% or
%\appendix  % for no appendix heading
% do not use \section anymore after \appendix, only \section*
% is possibly needed

% use appendices with more than one appendix
% then use \section to start each appendix
% you must declare a \section before using any
% \subsection or using \label (\appendices by itself
% starts a section numbered zero.)
%

\clearpage
\appendices

\section{Hyperparameter setup}

% 본문에서 언급했듯이 우리는 hyperparameter tuning에 의한 implicit bias를 막기위해서 algorithm-specific한 hyperparameter외의 general hyperparameters를 최대한 통일하여 공정하고 절제된 실험을 진행한다.
% Tab 1에는 Sec.~\ref{subsec:exp1}의 모든 실험에 사용된 algorithm에 상관없이 공유하는 general hyperparameter를 task마다 정리해둠
% Tab 2에는 on-policy algorithms인 PPO와 GenPO가 공유하는 hyperparameters을 task마다 정리해둠
% Tab 3에는 off-policy algorithms인 DDPG, TD3, SAC, DIPO, QVPO, DACER가 공유하는 hyperparameters를 task마다 정리해둠
% Finally, Tab 4에는 online DPRL들에 사용되는 task-specific parameters를 정리해둔다. Note that strict하게 task의 종류에 상관없이 강건하게 작동하는지 검증하기 위해 모든 task에 동일한 task-specific parameters를 사용함.

As discussed in the main paper, to mitigate implicit bias arising from algorithm-specific hyperparameter tuning, we standardized all general hyperparameters across different algorithms to ensure a fair and controlled experimental environment.
The overall hyperparameter configurations are organized as follows:
First, the specific parameters dedicated to online DPRL methods are provided in Table~\ref{tab:hp4}.
Tab~\ref{tab:hp1} enumerates the general hyperparameters shared by all algorithms to establish a consistent baseline, while Tab~\ref{tab:hp2} and Tab~\ref{tab:hp3} respectively detail the parameters specific to the on-policy group (PPO and GenPO) and the off-policy group (DDPG, TD3, SAC, DIPO, QVPO, and DACER).
% Our setup is organized into the following tables: Tab.~\ref{tab:hp1} presents the general hyperparameters shared across all algorithms for each task, ensuring a consistent baseline for comparison;
% Tab.~\ref{tab:hp4} specifies the task-specific parameters employed for online DPRL methods.
% Tab.~\ref{tab:hp2} summarizes the hyperparameters shared by on-policy algorithms (PPO and GenPO) across the evaluated tasks;
% Tab.~\ref{tab:hp3} details the hyperparameters common to off-policy algorithms, including DDPG, TD3, SAC, DIPO, QVPO, and DACER;
Note that to rigorously verify whether our proposed approaches function robustly regardless of the environment, we maintained identical task-specific parameters across all tasks.
This restrictive setup underscores the inherent capability and stability of the algorithms under a unified configuration.

\begin{table}[h]
\centering
\caption{Algorithm-specific hyperparameters of online DPRL methods (DIPO, QVPO, DACER, GenPO). All task-specific hyperparameters are shared across all environments.}
% \resizebox{0.99\linewidth}{!}{
\label{tab:hp4}
\small
\begin{tabular}{ll}
\toprule
\textbf{Hyperparameters} & \textbf{DIPO} \\
\midrule
Diffusion sampling steps & 100 \\
Action gradient steps & 20 \\
Target policy update freq. & 1 \\
Action learning rate & $3.0 \times 10^{-2}$ \\
\midrule
\textbf{Hyperparameters} & \textbf{QVPO} \\
\midrule
Diffusion sampling steps & 20 \\
Target policy update freq. & 1 \\
Running q mean & 1 \\
Running q std. & 0 \\
Alpha mean & 0.001 \\
Alpha std. & 0.001 \\
Entropy weight & 0.02 \\
diffusion sampling number & 64 \\
batch sampling number & 10 \\
\midrule
\textbf{Hyperparameters} & \textbf{DACER} \\
\midrule
Diffusion sampling steps & 20 \\
Target policy update freq. & 1 \\
Entropy learning rate & 0.03 \\
Initial entropy value & 0.02 \\
Target entropy & -dim(action) \\
Alpha update frequency & total timesteps / 10 \\
\midrule
\textbf{Hyperparameters} & \textbf{GenPO} \\
\midrule
Diffusion sampling steps & 5 \\
Target policy update freq. & 1 \\
Compress coefficient & 0.01 \\
Mixing coefficient & 0.9 \\
\bottomrule

\end{tabular}%}
\end{table}

\begin{table*}[h]
\centering
\caption{Common hyperparameters shared across all algorithms.}
\resizebox{0.99\linewidth}{!}{
\label{tab:hp1}
\small
\begin{tabular}{lllllll}
\toprule
\textbf{Environments} & \textbf{Cartpole} & \textbf{Ant} & \textbf{Humanoid} & \textbf{Franka Lift} & \textbf{Franka Open Drawer} & \textbf{Allegro Cube} \\
\midrule
\# of environment & 1024 & 1024 & 1024 & 1024 & 1024 & 1024 \\
Learning rate (Actor) & $3.0 \times 10^{-4}$ & $3.0 \times 10^{-4}$ & $5.0 \times 10^{-4}$ & $1.0 \times 10^{-4}$ & $5.0 \times 10^{-4}$ & $5.0 \times 10^{-4}$ \\
Learning rate (Critic) & $3.0 \times 10^{-4}$ & $3.0 \times 10^{-4}$ & $5.0 \times 10^{-4}$ & $1.0 \times 10^{-4}$ & $5.0 \times 10^{-4}$ & $5.0 \times 10^{-4}$ \\
Hidden layers (Actor) & [32, 32] & [256, 128, 64] & [400, 200, 100] & [256, 128, 64] & [256, 128, 64] & [512, 256, 128] \\
Hidden layers (Critic) & [32, 32] & [256, 128, 64] & [400, 200, 100] & [256, 128, 64] & [256, 128, 64] & [512, 256, 128] \\
Discount factor & 0.99 & 0.99 & 0.99 & 0.99 & 0.99 & 0.998 \\
GradNorm clip & 1.0 & 1.0 & 1.0 & 1.0 & 1.0 & 1.0 \\

\bottomrule
\toprule
\textbf{Environments} & \textbf{Anymal-D} & \textbf{B.D. Spot} & \textbf{Unitree Go2} & \textbf{Unitree G1} & \textbf{Unitree H1} & \textbf{Cassie} \\
\midrule
\# of environment & 1024 & 1024 & 1024 & 1024 & 1024 & 1024 \\
Learning rate (Actor) & $1.0 \times 10^{-3}$ & $1.0 \times 10^{-3}$ & $1.0 \times 10^{-3}$ & $1.0 \times 10^{-3}$ & $1.0 \times 10^{-3}$ & $1.0 \times 10^{-3}$ \\
Learning rate (Critic) & $1.0 \times 10^{-3}$ & $1.0 \times 10^{-3}$ & $1.0 \times 10^{-3}$ & $1.0 \times 10^{-3}$ & $1.0 \times 10^{-3}$ & $1.0 \times 10^{-3}$ \\
Hidden layers (Actor) & [128, 128, 128] & [512, 256, 128] & [512, 256, 128] & [512, 256, 128] & [512, 256, 128] & [512, 256, 128] \\
Hidden layers (Critic) & [128, 128, 128] & [512, 256, 128] & [512, 256, 128] & [512, 256, 128] & [512, 256, 128] & [512, 256, 128] \\
Discount factor & 0.99 & 0.99 & 0.99 & 0.99 & 0.995 & 0.99 \\
GradNorm clip & 1.0 & 1.0 & 1.0 & 1.0  & 1.0 & 1.0 \\
\bottomrule
% \multicolumn{2}{l}{\small *QVPO and DACER use [256, 256, 256].} \\
% \multicolumn{2}{l}{\small **PPO uses ELU; DACER uses ReLU.}
\end{tabular}}
\end{table*}

%-----------------------------------------------------------------------------

\begin{table*}[h]
\centering
\caption{Common hyperparameters shared across on-policy Algorithms (PPO, GenPO).}
\resizebox{0.99\linewidth}{!}{
\label{tab:hp2}
\small
\begin{tabular}{lllllll}
\toprule
\textbf{Environments} & \textbf{Cartpole} & \textbf{Ant} & \textbf{Humanoid} & \textbf{Franka Lift} & \textbf{Franka Open Drawer} & \textbf{Allegro Cube} \\
\midrule
Rollouts & 16 & 16 & 32 & 24 & 96 & 24 \\
Learning epochs & 8 & 4 & 5 & 8 & 5 & 5 \\
\# of mini-batches & 8 & 2 & 4 & 4 & 96 & 12 \\
GAE Smoothing Param. & 0.95 & 0.95 & 0.95 & 0.95 & 0.95 & 0.95 \\
Desired KL & 0.008 & 0.008 & 0.01 & 0.01 & 0.008 & 0.016 \\
Surrogate clip & 0.2 & 0.2 & 0.2 & 0.2 & 0.2 & 0.2 \\
Value clip & 0.2 & 0.2 & 0.2 & 0.2 & 0.2 & 0.2 \\
Entropy coefficient & 0.0 & 0.0 & 0.0 & 0.001 & 0.001 & 0.002 \\
Value loss coefficient & 2.0 & 1.0 & 2.0 & 2.0 & 2.0 & 2.0 \\

\bottomrule
\toprule
\textbf{Environments} & \textbf{Anymal-D} & \textbf{B.D. Spot} & \textbf{Unitree Go2} & \textbf{Unitree G1} & \textbf{Unitree H1} & \textbf{Cassie} \\
\midrule
Rollouts & 24 & 24 & 24 & 24 & 24 & 24 \\
Learning epochs & 5 & 5 & 5 & 5 & 5 & 5 \\ 
\# of mini-batches & 4 & 4 & 4 & 4 & 4 & 4 \\
GAE Smoothing Param. & 0.95 & 0.95 & 0.95 & 0.95 & 0.95 & 0.95 \\
Desired KL & 0.01 & 0.01 & 0.01 & 0.01 & 0.01 & 0.01 \\
Surrogate clip & 0.2 & 0.2 & 0.2 & 0.2 & 0.2 & 0.2 \\
Value clip & 0.2 & 0.2 & 0.2 & 0.2 & 0.2 & 0.2 \\
Entropy coefficient & 0.005 & 0.0025 & 0.01 & 0.008 & 0.01 & 0.01 \\
Value loss coefficient & 1.0 & 0.5 & 1.0  & 1.0  & 1.0  & 1.0  \\
\bottomrule

\end{tabular}}
\end{table*}

%-----------------------------------------------------------------------------

\begin{table*}[h]
\centering
\caption{Common hyperparameters shared across off-policy Algorithms (DDPG, TD3, SAC, DIPO, QVPO, DACER).}
\resizebox{0.99\linewidth}{!}{
\label{tab:hp3}
\small
\begin{tabular}{lllllll}
\toprule
\textbf{Environments} & \textbf{Cartpole} & \textbf{Ant} & \textbf{Humanoid} & \textbf{Franka Lift} & \textbf{Franka Open Drawer} & \textbf{Allegro Cube} \\
\midrule
Batch size & 4096 & 4096 & 4096 & 4096 & 4096 & 4096 \\
Replay buffer size & 4096 & 4096 & 4096 & 4096 & 4096 & 4096 \\
Soft update parameter & 0.005 & 0.005 & 0.005 & 0.005 & 0.005 & 0.005 \\
Random action steps & 4 & 4 & 4 & 4 & 4 & 4 \\

\bottomrule
\toprule
\textbf{Environments} & \textbf{Anymal-D} & \textbf{B.D. Spot} & \textbf{Unitree Go2} & \textbf{Unitree G1} & \textbf{Unitree H1} & \textbf{Cassie} \\
\midrule
Batch size & 4096 & 4096 & 4096 & 4096 & 4096 & 4096 \\
Replay buffer size & 4096 & 4096 & 4096 & 4096 & 4096 & 4096 \\
Soft update parameter & 0.005 & 0.005 & 0.005 & 0.005 & 0.005 & 0.005 \\
Random action steps & 4 & 4 & 4 & 4 & 4 & 4 \\
\bottomrule

\end{tabular}}
\end{table*}

% % use section* for acknowledgment
% \section*{Acknowledgment}

\end{document}